\documentclass[11pt,a4paper]{article}
\usepackage{arxiv}
\usepackage[utf8]{inputenc}
\usepackage[T1]{fontenc}
\usepackage{amsmath}
\usepackage{bm}
\usepackage{algorithm}
\usepackage{algpseudocode}
\usepackage{amssymb}
\usepackage[bottom]{footmisc}
\usepackage{graphicx}
\graphicspath{{./images/}}
\usepackage[font=small,labelfont=bf]{caption}
\usepackage{subfig}
\usepackage{hyperref}

\usepackage[backend=biber,style=nature,url=false,isbn=false]{biblatex}
\usepackage{xcolor}
\definecolor{reviewblue}{RGB}{0,50,180}

\usepackage[inkscapearea=page,inkscapelatex=false]{svg}
\usepackage[export]{adjustbox}
\usepackage{sidecap}
\usepackage[english]{babel}
\usepackage[autostyle, english = american]{csquotes}
\MakeOuterQuote{"}
\usepackage{multicol}
\usepackage{float}

\newenvironment{supplementary}{
  \setcounter{figure}{0}
  \setcounter{equation}{0}

}{}

\newcommand{\unit}[1]{\,\text{#1}} 

\DeclareMathOperator{\E}{\mathbb{E}}

\DeclareMathOperator{\R}{\mathbb{R}}

\newcommand{\norm}[1]{\left\lVert#1\right\rVert}

\title{Robust and efficient enriched representation with intrinsic neuronal heterogeneity} 

\author{
    \href{https://orcid.org/0000-0001-6063-7992}{\includegraphics[scale=0.06]{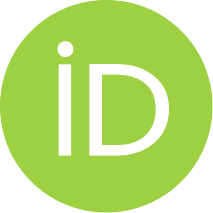}\hspace{1mm}Arash ~Golmohammadi \textsuperscript{1}}
    \\
    \texttt{arash.golmohammadi@med.uni-goettingen.de} \\
	\And
	\href{https://orcid.org/0000-0002-1959-5644}{\includegraphics[scale=0.06]{orcid.pdf}\hspace{1mm}Jannik ~Luboeinski \textsuperscript{1, 2}} \\
	\texttt{jannik.luboeinski@med.uni-goettingen.de} \\
	\And
	\href{https://orcid.org/0000-0003-1901-6232}{\includegraphics[scale=0.06]{orcid.pdf}\hspace{1mm}Christian ~Tetzlaff \textsuperscript{1, 2, *}} \\
	\texttt{christian.tetzlaff@med.uni-goettingen.de} \\
}
\date{%
    \vspace{-1em}
	\textsuperscript{1} \small{Group of Computational Synaptic Physiology, Department for Neuro- and Sensory Physiology,\\ University Medical Center Göttingen, Humboldtallee 23, 37073 Göttingen, Germany}\\%
    \vspace{.5em}
	\textsuperscript{2} \small{Campus Institute Data Science (CIDAS), Goldschmidtstraße 1, 37077 Göttingen, Germany}\\
    \vspace{.5em}
	\small{* Corresponding author}
}

\begin{document}

\begin{refsection}

\maketitle

\begin{abstract}
    Heterogeneity is a ubiquitous property of many biological systems with profound computational implications. Integrating this heterogeneity into artificial computing systems has recently gained attention. While optimizing neuronal and synaptic heterogeneity for a specific task can achieve state‑of‑the‑art performance, such top‑down optimization is biologically implausible, prone to catastrophic forgetting, and both data‑ and energy‑intensive. In contrast, biological organisms possess inherent, task‑unspecific, stable heterogeneity and can perform many tasks with minimal metabolic cost. Inspired by this, we study task‑agnostic heterogeneity in recurrent neural networks solving hundreds of diverse temporal tasks. Heterogeneity is introduced as intrinsic disorder in neuronal membrane time constants. Our results show that intrinsic heterogeneity markedly improves overall performance and robustness in both rate‑ and spike‑coded networks. Small heterogeneous networks, even without recurrence, match the performance of large, highly recurrent, homogeneous networks at a fraction of the cost and complexity. We suggest that, despite being task‑agnostic, heterogeneity enhances representational capacity by diversifying neuronal responses, demonstrated on two machine‑learning and one neuroscience benchmark. Our work highlights the utility of intrinsic task‑agnostic heterogeneity for versatile, energy‑efficient computing—a highly relevant property for the neuromorphic community, where device‑to‑device variability could be exploited to offset limited training budgets and architectural constraints.
\end{abstract}

\section*{Plain Language Summary}
\textbf{Aim and background}:  Living brains are composed of diverse cells. This cellular diversity plays an important role in enabling organisms to perform numerous metabolic and cognitive tasks efficiently. Therefore, incorporating this diversity into artificial systems can make them more efficient. However, tuning such diversity is expensive.

\textbf{Method}: We create networks of neurons with either identical or diverse responses. As in biological brains, the diversity is random and not tuned. We test the networks on hundreds of different tasks.

\textbf{Main finding}: Networks that contain this intrinsic diversity perform better and are more robust than uniform networks. Even small networks with added variety can match the performance of much larger, highly connected uniform networks while using far less energy and complexity. The variety helps the networks represent information more richly, as shown on several standard benchmarks.

\textbf{Significance}: The results suggest that embracing natural, task‑independent variability can make artificial computing systems cheaper, faster, and more energy‑efficient. Such variability is also useful in hardware design for maintaining efficiency and reducing costs.

\newpage

\section{Introduction}

The brain is a complex recurrent neural network (RNN) whose dynamics--the temporal activity of its constituent neurons--underlie thoughts, behaviors, and actions. Neuronal activity depends strongly on the electrophysiological properties of neurons and synapses. These properties enable the brain to execute a plethora of cognitive tasks with an estimated power consumption of only 20 W \cite{mink1981}. Therefore, distilling the structure of these properties is necessary for understanding the brain's mechanism of function and designing intelligent systems.

Unlike artificial systems, in which computational units are often identical, biological neurons and synapses exhibit considerable electrophysiological heterogeneity \cite{chen2009, zeng2017, seeman2018, cembrowski2019, hasse2019, chen2021, yao2021, campagnola2022, yao2023, langlieb2023}, sometimes exceeding two orders of magnitude within the same neuronal type. The computational benefit of such diversity for a given function or brain module is highlighted in numerous theoretical \cite{golomb1993, shamir2006, ostojic2009, perez2010, luccioli2010, tripathy2013, lengler2013, mejias2014, angulo-garcia2017, beiran2018, luccioli2019, divolo2021, destexhe2022, gast2024} and experimental studies \cite{padmanabhan2010, holmstrom2010, angelo2012, kennedy2014, dezeeuw2023}. Owing to its prominence and function, it is unlikely that parameter heterogeneity has arisen purely from randomness; rather, it likely reflects a structure that the brain exploits for the organism's survival.

Neuronal heterogeneity can be organized either top‑down, in a task‑informed manner, or bottom‑up, intrinsically and task‑agnostically. In the former case, all biological properties (such as synaptic conductances, ion concentrations, molecular expressions, etc.) are treated as degrees of freedom that can be simultaneously adjusted to optimize some high‑level objective function. This approach, which is the mainstream paradigm in machine learning, has proven remarkably effective. In particular, gradient‑based learning when combined with neuronal heterogeneity has led to performance levels equal to or better than the state of the art on several common benchmarks \cite{matsubara2017, fang2021, perez-nieves2021, winston2023, zheng2024}.

Despite this success, the gradient‑backpropagation rule is criticized for its biological implausibility \cite{lillicrap2020, ororbia2024} and task‑specificity, which results in the catastrophic‑forgetting phenomenon \cite{french1999, aleixo2023}. The former is particularly true for non‑synaptic parameters. Although synaptic update rules that backpropagate only a local estimate of the gradient exist \cite{aleixo2023, ororbia2024}, a biologically plausible equivalent for non‑synaptic parameters (e.g., neuronal) is lacking. Furthermore a top‑down optimization of neuronal heterogeneity would require drastic activity‑dependent changes in either the neurons’ passive biophysical properties or their morphology. Yet, both of these properties are known to remain largely stable throughout adulthood in mammals \cite{chen2010, pannese2011}. Additionally, such end‑to‑end training involving both synaptic and neuronal parameters, even for a single task, is energetically costly \cite{mehonic2022} and data‑intensive \cite{hestness2017, liu2024}.

In contrast to such top‑down heterogeneity organization, bottom‑up parameter heterogeneity does not suffer from the issues of task‑specificity (since it is task‑agnostic) and high computational cost (as it involves no activity‑dependent training). Hence, it is reasonable to assume that biological properties are organized intrinsically. However, it is not immediately clear whether such a task‑agnostic parameter organization can support the broad range of vital and cognitive computations required by living organisms. While it is conceivable that such task‑agnostic structures facilitate the execution of one task while hindering another, their holistic effect on a collection of tasks is unknown. The goal of this study is to address this question.

To this end, we use the reservoir computing paradigm \cite{jaeger2002, maass2002, lukosevicius2009} and construct recurrent neural networks (RNNs) that differ only in their level of intrinsic neuronal heterogeneity. Each network is then assessed on hundreds of dissimilar tasks that span a wide spectrum of complexities. Thus, by vastly expanding the task space, we mitigate the task‑specificity bias, a crucial shortcoming of studies that focus on one or few benchmarks. 
Furthermore, to ensure our results are not model‑dependent, we consider both spiking and rate‑based RNNs. On the one hand, the rate‑based implementation serves as a prototype for common machine‑learning models, allowing us to predict the effect of task‑unspecific heterogeneity on mainstream artificial systems. On the other hand, the spiking implementation illustrates how crucial intrinsic heterogeneity is in the biologically more realistic, event‑based coding, and that useful efficiency gains may also be achieved on neuromorphic hardware systems. 

Building on our previous work \cite{golmohammadi2024}, our results indicate that a specific form of intrinsic heterogeneity significantly enhances both overall performance and robustness, irrespective of the neural code. In particular, we observe that small, heterogeneous RNNs perform comparably to, and sometimes better than larger homogeneous networks, yielding simpler systems with higher energy efficiency. We show that the computational principle underlying this performance boost is \textit{resource diversification}, which facilitates the execution of diverse tasks across various computing systems. We demonstrate the manifestation of this diversification across three common benchmarks and conclude by discussing the implications of our findings for the neuromorphic community.

\begin{figure}[ht]
	\centering
    \includegraphics[width=.85\linewidth]{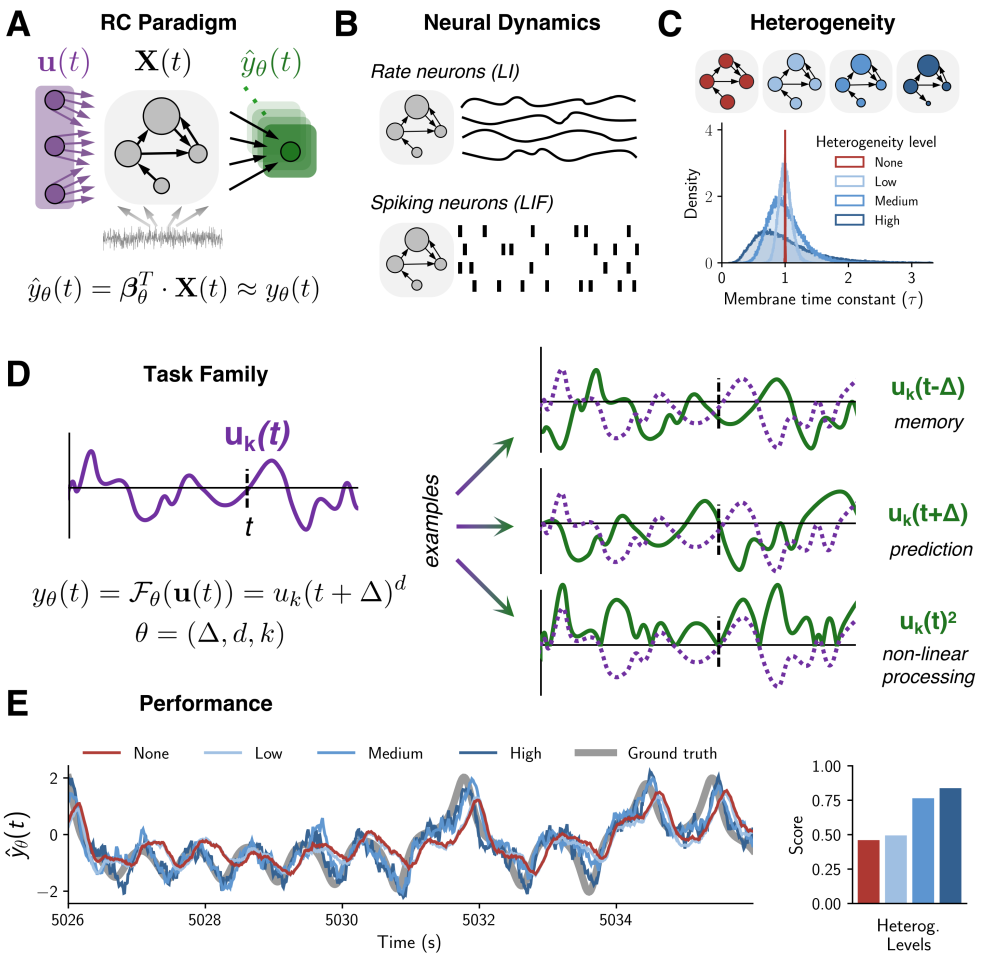}
	\caption{
    \textbf{Method summary}. 
    \textbf{(A)} The reservoir computing paradigm. The multi‑dimensional input $\bm u(t)$ drives a recurrent network, giving rise to the neural state $\bm X(t)$. This state is then read out linearly to approximate arbitrary target functions or tasks $y_{\theta}(t)$. $\theta$ indexes each task in a collection of tasks. Note that all target functions use the same state $\bm X(t)$ but different readouts $\bm{\beta}_{\theta}$. 
    \textbf{(B)} A schematic representation of \textit{(top)} rate‑based (leaky integrator, LI) and \textit{(bottom)} spiking (leaky integrate‑and‑fire, LIF) neuronal codes. The state of neurons in the LI implementation is a continuous time series, whereas neurons in the LIF implementation spike discretely.  
    \textbf{(C)} Neuronal heterogeneity is introduced by assigning different membrane time constants to the neurons, drawn from a distribution with a given mean and variance. The variance determines the heterogeneity level.  
    \textbf{(D)} The task family targeted by our approach includes tasks such as memory recall, input prediction, input processing, and their combinations.  
    \textbf{(E)} \textit{(Left)} The output $\hat y_{\theta}(t)$ of networks with various heterogeneity levels for an exemplary task ($\theta=(61, 1, 1)$). \textit{(Right)} Performance of each network measured using the coefficient of determination, a scalar score that aggregates the normalized mismatch between the ground truth and the output over an unobserved test interval. A value of 1 indicates a perfect match, whereas 0 corresponds to a chance‑level prediction.}
	\label{fig:method_summary}
\end{figure}

\section{Results}
\subsection{Network and task setup}\label{sec:setup}
The network topology is random and sparse, with loosely balanced excitatory and inhibitory populations. The feed‑forward weights are dense and drawn from a normal distribution. Networks are driven by a time‑dependent multidimensional input $\bm u(t)$ and neuron‑specific white noise. These inputs result in a time‑dependent neuronal state $\bm X(t)$, which is used to linearly approximate the target tasks $y_{\theta}(t)$.  (cf. Fig.~\ref{fig:method_summary}A). Neurons evolve according to either rate‑based (leaky integrator, LI) or spike‑based (leaky integrate‑and‑fire, LIF) dynamics (Fig.~\ref{fig:method_summary}B). Heterogeneity is introduced by allowing the membrane time constants, (defined as the ratio between capacitance and leak conductance), to differ across neurons, while keeping their mean fixed (Fig.~\ref{fig:method_summary}C). Unless otherwise stated, membrane time constants are drawn from a log‑normal distribution with mean $\mathbb{E}[\tau]=1$ and variance $\operatorname{Var}[\tau]=h\,\mathbb{E}[\tau]^2$, where $h\in\{0,\,0.1,\,1,\,10\}$ defines the heterogeneity level. 

Apart from the time constant distribution, all networks share the same connectivity matrix, feed‑forward and recurrent weights, and the same input and noise realizations. We use a network size of $N=250$, recurrent sparsity $p=0.1$, and an excitatory‑to‑inhibitory ratio of 4:1. The magnitude of excitatory and inhibitory recurrent weights is controlled by a global recurrent gain $J=1$ and a fixed variance of $\sigma_{0}=1$. The gains of the deterministic feed‑forward input and the random noise are $J_{u}=1$ and $J_{n}=0.1$, respectively (corresponding to a signal‑to‑noise ratio of $10$). These hyperparameters are varied in Section~\ref{sec:robustness} to assess the robustness of our findings to these choices.

Performance of a given network--whether heterogeneous or homogeneous--depends on the complexity of the task. An easy task can be performed perfectly by a simple network, whereas even large‑scale, modular, and complex networks may fail on challenging tasks. Thus, a high‑quality benchmark that gauges a network’s computational capacity must include (a) dissimilar tasks and (b) tasks spanning a wide range of complexities.

We curate such a task array by focusing on a particular family of cognitive tasks, where the output $y(t)$ is a pure function of the stimulus $\bm u(t)$, and independent of the network’s internal state \(\bm X(t)\). Such family, which encompasses several working‑memory like functions, can be formally cast as the mapping $y(t) = \mathcal{F}\big (\bm u(t) \big)$, where the functional form of \(\mathcal{F}\) determines the specifics of the task. For convenience we employ a parameterized mapping. Specifically our task family is defined by $\mathcal{F}_{\theta}\bigl(\bm u(t)\bigr)=u_{k}(t+\Delta)^{d}$, where \(\theta=(\Delta,d,k)\) are the mapping parameters. Changing \(\Delta\) shifts the input in time, emulating memory recall ($\Delta<0$) or stimulus forecasting ($\Delta<0$). The exponent \(d\in\mathbb{N}\) controls the degree of nonlinearity applied to the time‑shifted stimulus mirroring nonlinear input processing. Finally the index \(k\in\{1,\dots,K\}\) selects the input component that is to be extracted or attended (cf. Fig.~\ref{fig:method_summary}D). For any input $\bm u(t) \in \R ^K$, we uniformly sampled parameters from the range $d\in\{1,2,\dots,6\}$ and $k \in \{1,\dots,K\}$ and logarithmically from $\Delta \in [-2,2]$, resulting a task array of size $294 \times K$. Note that we first temporally scale $\bm u(t)$ so that the stimulus has an effective timescale of one (cf. Section \ref{sec:stim} for details).

Once the task family is constructed, we empirically quantify the dissimilarity among tasks and the range of task complexities.  
Task dissimilarity is measured by computing the distribution of temporal correlations (cosine similarities) between all pairs of tasks. Task complexity, is estimated with the cosine distance between the target \(y_{\theta}(t)\) and its generating input component \(u_{k}(t)\); A complexity value of 0 corresponds to the simplest task (a pure relay), whereas a value of 1 denotes maximal complexity, where the target is orthogonal to the generating input (see Section~\ref{sec:complexity}).

To quantify the task-specific performance of each network, we first determine the linear readout $\bm \beta_{\theta}$ that best approximates that task $y_{\theta}(t)$ over a training interval. We then compute $s_{\theta}$, the coefficient of determination between network's prediction $\hat y_{\theta}(t) = \beta_{\theta}^T \cdot \bm X(t)$ and the task on an unobserved test interval (see Sec.~\ref{sec:fit} for details). $s_{\theta} \le 1$ is the normalized mismatch between the two traces; A value of 1 indicates a perfect match, 0 corresponds to chance‑level prediction, and negative values signal performance worse than chance (cf. Fig.~\ref{fig:method_summary}E).  Finally, the collection of task‑specific scores yields the overall performance profile of each network, which is broken down by heterogeneity level.

\subsection{Heterogeneity increases performance score across tasks, inputs, and dynamics}\label{sec:processing} 
\begin{figure}[ht]
	\centering
	\includegraphics[width=\linewidth]{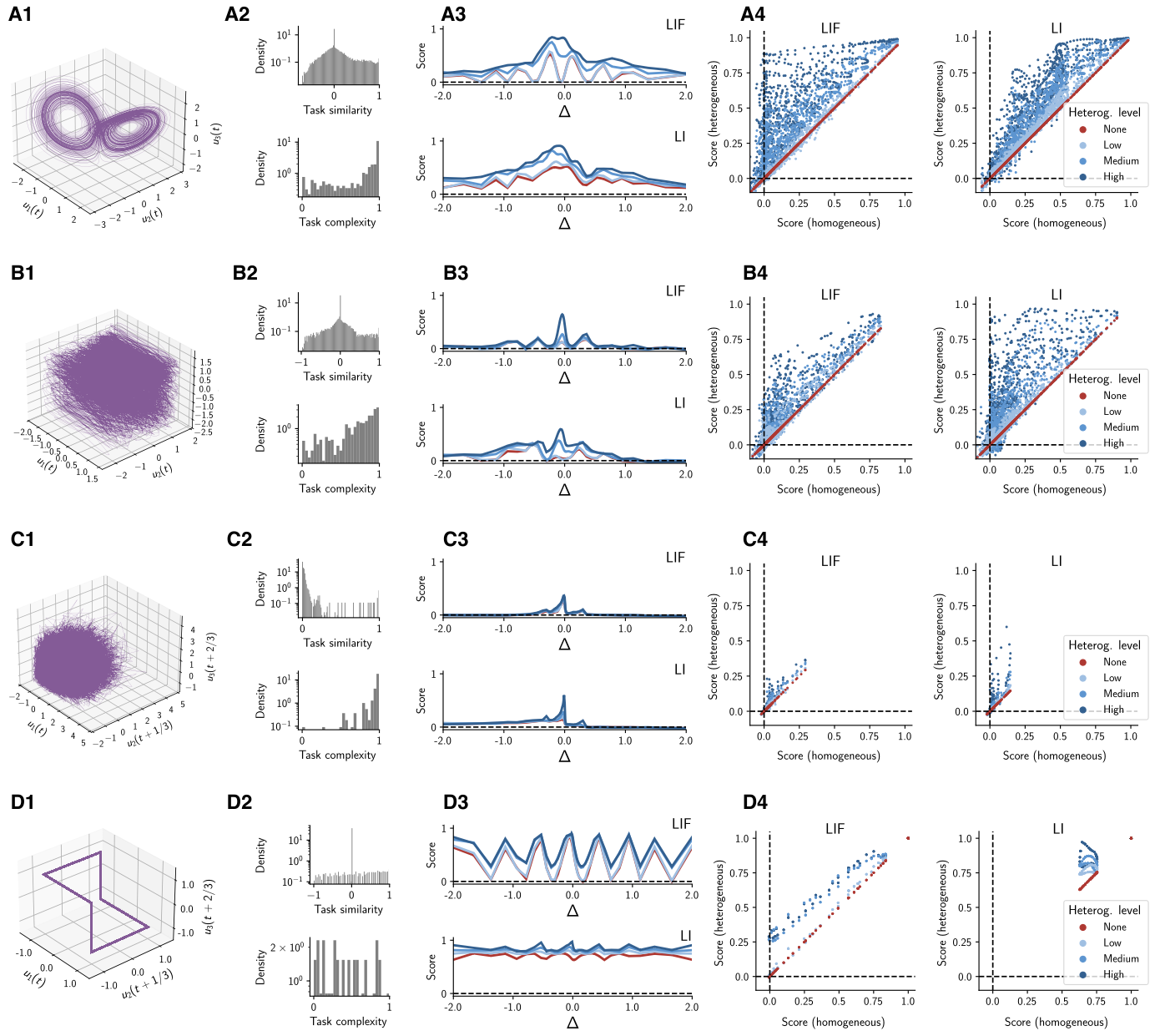}
    \caption{\textbf{Intrinsically heterogeneous networks process time series more accurately.} 
    \textbf{(A1)} The chaotic trajectory of the Lorenz system is used as a three-dimensional input to create a collection of 882 tasks. 
    \textbf{(A2)} \textit{(top)} Mutual similarity and \textit{(bottom)} complexity distribution of the tasks, showing they are mostly dissimilar and span a broad spectrum of complexities. 
    \textbf{(A3)} Cross‑sections of the performance profile for $d=2$, $k=1$ and various $\Delta$ values are shown for the \textit{(top)} spiking and \textit{(bottom)} rate‑based neurons. 
    \textbf{(A4)} Direct score comparison between heterogeneous (y‑axis) and corresponding homogeneous (x‑axis) networks for each task (each dot). Colors indicate the degree of heterogeneity. The diagonal dotted line marks equal performance, while the vertical and horizontal dashed lines denote the chance‑level score. The right‑hand panel corresponds to the spiking implementation and the left‑hand panel to the rate‑based implementation. Panels (A3) and (A4) share the same color code. 
    The other rows show similar experiments with different inputs: 
    \textbf{(B)} Mackey‑Glass input (three‑dimensional; two components chaotic, one periodic), 
    \textbf{(C)} Order‑30 NARMA input (one‑dimensional, chaotic and non‑smooth), and 
    \textbf{(D)} Absolute value of a sinusoid as input (one‑dimensional, smooth and predictable).
    For 1D inputs, the first column shows the 3D Poincare sections, and 294 tasks are generated.}
    \label{fig:perf}
\end{figure}

To emulate the multi-modal, partially predictable nature of sensory inputs in the brain, we first used a multi-dimensional chaotic time series as $\bm u(t)$. Specifically, we employed the trajectory of the Lorenz system \cite{lorenz1963} (see Sec.~\ref{sec:stim}), a well‑known three‑dimensional chaotic time series (Fig.~\ref{fig:perf}A1), which serves as a common benchmark for time‑series modeling \cite{gilpin2023a}. As Fig.~\ref{fig:perf}A2 shows, the majority of the tasks are independent of one another. Moreover, they exhibit a broad spectrum of complexities, which is due to the unpredictability of the stimulus. Taken together with the sheer number of tasks (882), this task family constitutes a high‑quality benchmark for quantifying the performance profiles of homogeneous and heterogeneous networks.

Panels in Fig.~\ref{fig:perf}A3 show the performance profile for a representative subset of tasks that probe the network’s (quadratic) memory or prediction capabilities ($k=1$, $d=2$) across a range of temporal‑shift values $\Delta$. In both rate‑ and spike‑based implementations, increasing heterogeneity yields higher performance scores over almost the entire $\Delta$ range. The small standard deviation of the scores – still modest after a ten‑fold enlargement – indicates that this improvement is statistically significant (see Section~\ref{sec:robustness}). The oscillatory pattern of performance stems from the revolution between manifold lobes. A similar effect is expected from any (partially) periodic target function $y_{\theta}$. The occasional performance peaks of homogeneous networks indicate that they are suitably initialized for those subsets of tasks (specific $\Delta$ values) but not for others. In contrast, heterogeneous networks are insensitive to task selection, highlighting their versatility.

Importantly, the benefit is not limited to this particular subset. As the panels in Fig.~\ref{fig:perf}A4 illustrate, for every task the scores of heterogeneous networks (plotted on the $y$‑axis) are systematically greater than those of the corresponding homogeneous networks (plotted on the $x$‑axis) in both implementations.

Although this performance profile demonstrates the superiority of heterogeneous networks across more than 800 tasks, the curated task family is influenced by the topological structure of its generating time series, specifically, the stable attractor of the Lorenz system. To test whether the observed performance boost depends on the stimulus structure, we replaced the Lorenz components with three independent one‑dimensional Mackey–Glass time series \cite{mackey1977}, two of which are chaotic. As Fig.~\ref{fig:perf}B1 shows, this concatenation eliminates the topological structure in the input. Nevertheless, the pronounced superiority of heterogeneous networks is largely preserved (cf. Fig.~\ref{fig:perf}B3–B4). Also note that Fig.~\ref{fig:perf}B3 indicates that the resulting task family remains of high quality.

As Figs.~\ref{fig:perf}A and B show, the temporal characteristics of $\bm u(t)$ have a large impact on the networks’ performance profiles. To test whether the superiority of heterogeneous networks persists when these characteristics change, we considered two extreme cases: the NARMA‑30 series and a purely periodic signal. NARMA‑30 is an order 30 autoregressive 1‑D process that generates a highly irregular, non‑smooth trajectory \cite{atiya2000}. In contrast, an input consisting of a single periodic component is maximally smooth and predictable. Panels Fig.~\ref{fig:perf}C1–C4 illustrates that, for the NARMA‑30 input, the irregularity produces challenging tasks and consequently markedly lower performance across all tasks and neuronal implementations. Conversely, the performance profile for the periodic input is relatively high and structured, as shown in Fig.~\ref{fig:perf}D1–D4. Importantly, despite these differences in overall performance, heterogeneous networks exhibit a superior performance profile in both cases, indicating that the benefit of heterogeneity is robust to the temporal characteristics of the input.  

In the remainder of the paper we use the Lorenz stimulus and investigate the sensitivity of these results to variations in the hyperparameters.

\subsection{Heterogeneous networks are more robust to hyperparameters variation} \label{sec:robustness}
To assess whether the performance gains induced by heterogeneity depend on the RNN hyperparameters, we systematically varied several key settings and recomputed the performance profiles of the networks as described above. For brevity, we report only the changes in mean performance here; the full distributions of scores across individual tasks, together with their uncertainties, are provided in Sections E and F, respectively.

\begin{figure}[!b]
	\centering
	\includegraphics[width=\linewidth]{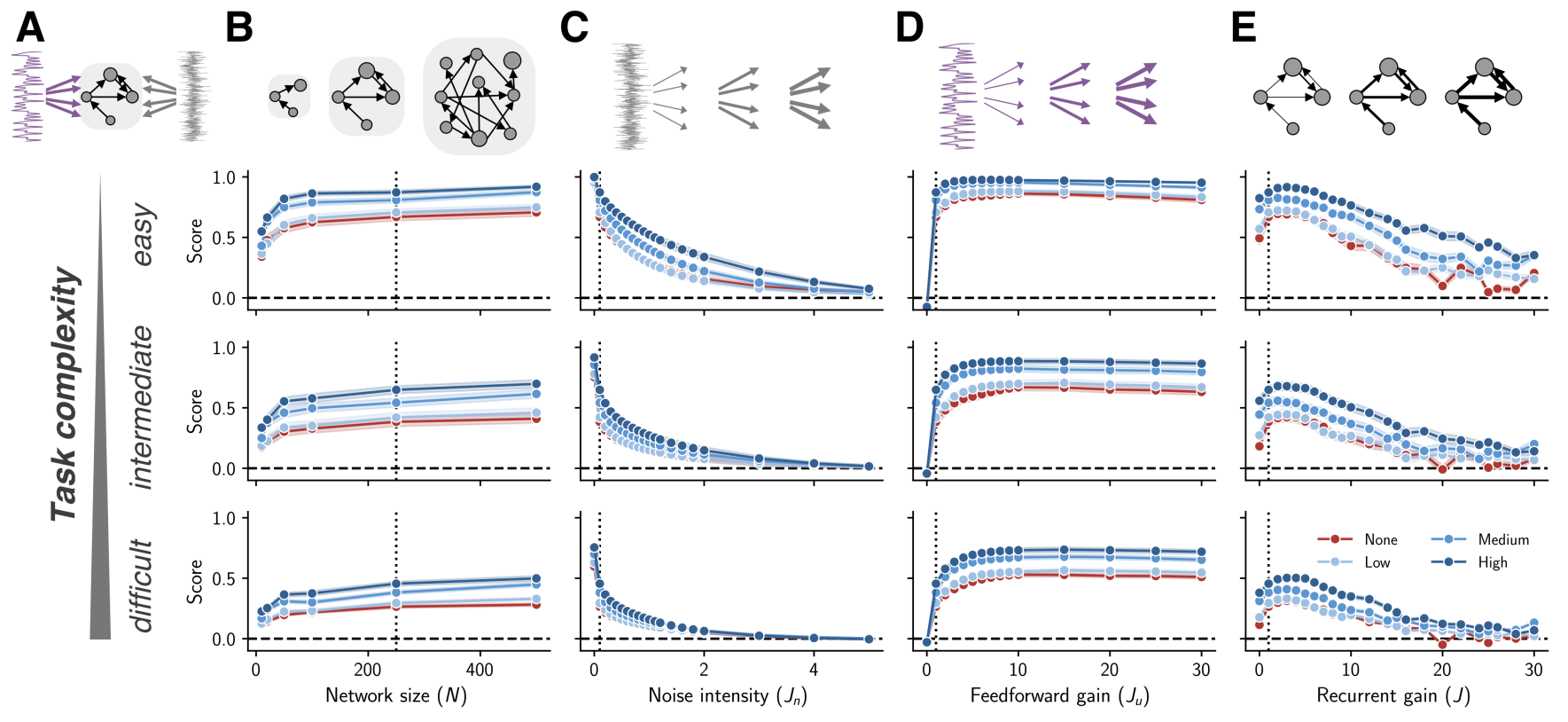}
	\caption{
    \textbf{Neuronal heterogeneity renders networks robust.}  
    \textbf{(A)} \textit{(Top)} Schematic of the network and its inputs. Purple and green traces represent the deterministic input and stochastic noise that are projected onto the network at its center. \textit{(Bottom)} Tasks are divided into three complexity tiers (indicated by the gradient), and performance is averaged within each tier for clarity.  
    \textbf{(B–E)} \textit{(Top)} Schematic of the varied hyperparameters; see panel (A) for reference. \textit{(Bottom)} Impact of varying network size (B), noise intensity (C), feedforward gain (D), and recurrent gain (E) on the performance of rate‑based RNNs. All panels use the same color code. The plateaus in panel (D) arise from saturation of neuronal activity. The uniform decline of the score at large recurrent‑gain values in panel (E) reflects a transition to chaos. Analogous results for spiking networks are shown in Fig. S3. In all panels dots and shaded area represent the mean and 95\% confidence interval, calculated over tasks (of a fixed complexity tier) and three non-overlapping training sets (see Sec.~\ref{sec:uq} for details). 
    }
	\label{fig:robustness_li}
\end{figure}

We first examined how network size \(N\) influences the performance profile. Because reservoir computers are universal approximators \cite{maass2002, grigoryeva2018, gonon2021}, larger networks generally achieve higher performance. In the limit \(N\!\to\!\infty\), the random recurrent coupling among neurons is guaranteed to generate an arbitrarily rich dynamical landscape, sufficient to approximate any function regardless of the network’s heterogeneity level. Real networks, however, are finite (\(N<\infty\)) and therefore perform less optimally. Consequently, a key question is whether heterogeneity can rescue the performance of such finite networks that do not satisfy the assumptions of the universality theorem. As Fig.~\ref{fig:robustness_li}B (and Fig. S3B) shows, the average performance of heterogeneous rate‑based (respectively spiking) networks is superior to that of their homogeneous counterparts across all sizes and complexity tiers. Complexity tiers are defined by dividing the full complexity value range (0-1) into three equal-width intervals. Task with complexity values in the lower, middle, and upper thirds are labeled respectively as easy, medium, and hard. Remarkably, networks as small as \(N=50\) neurons, when highly heterogeneous, outperform a homogeneous network ten times larger in every task tier. This indicates that time‑constant heterogeneity is an effective strategy for recreating a rich dynamical repertoire and rescuing performance in the small‑size regime.

Next, we asked whether heterogeneous networks, like biological organisms, are robust to strong stochastic stimuli. To answer this question we varied the noise intensity $J_{n}$ (the standard deviation of the fluctuations) from 0 (no noise) to 5 (corresponding to a signal‑to‑noise ratio of $0.2$) and repeated the experiments. As shown in Fig.~\ref{fig:robustness_li}C (and Fig. S3C), heterogeneous rate‑based and spiking networks outperform their homogeneous counterparts at all noise levels. Thus, although they are not immune to extreme noise, they exhibit markedly greater resilience.

We then examined the effects of the feed‑forward gain $J_{u}$ (from the input to the network) and the recurrent gain $J$ (within the network), which together determine the dynamical regime of the RNN. When $J_{u}\gg J$, the network is effectively input‑driven. Moreover, if $J$ is sufficiently small, the network exhibits a fading‑memory condition: whether or not input is present, any memory of the initial state gradually fades away. In this input‑driven regime, the fading‑memory condition \cite{boyd1985, maass2002} (also known as the echo‑state property \cite{jaeger2001}, contractive dynamics \cite{bullo2024}, or dissipative dynamics \cite{strogatz2015}) is particularly desirable for working‑memory‑like input‑output associations (as defined in Section \ref{sec:setup}), because such mappings can be learned without interference from the internal state.

Conversely, when $J_{u}\ll J$, the network operates in a largely state‑driven (autonomous) regime, in which the next state of the network is dictated primarily by its previous state rather than by the external input. At the extreme, very strong recurrence gives rise to chaotic activity \cite{sompolinsky1988}, which is undesirable for systems that must implement working‑memory‑like input-output associations with dynamic states. In practice, a moderate level of recurrence is necessary to ensure the richness of the dynamical repertoire, supporting memory maintenance, and nonlinear processing required by the task.

As Fig.~\ref{fig:robustness_li}D shows for the rate‑based networks (and Fig. S3D for the spiking RNNs), increasing the feed‑forward gain $J_{u}$ improves performance across all complexity tiers. A larger $J_{u}$ pushes the network toward an input‑driven regime, making the input‑output association easier. Nevertheless, even at very high $J_{u}$ values, the maximal performance of homogeneous networks remains lower than that of highly heterogeneous networks for almost all $J_{u}$ levels.

Fig.~\ref{fig:robustness_li}E and Fig. S3E summarize the effect of varying the recurrent gain $J$ in the rate‑based and spiking RNNs, respectively. As before, heterogeneous networks uniformly outperform their homogeneous analogs in both implementations. Moreover, as expected, very strong recurrence prevents any meaningful association between input and output. For the rate‑based networks, the abrupt performance drop at $J_{c}\approx 18$ marks the critical recurrence value at which all networks--regardless of heterogeneity level--transition into an autonomous chaotic state. This critical value, however, is considerably higher for spiking RNNs (cf. Fig. S20G).
Yet, apart from the uniform superiority of heterogeneous networks and the abrupt performance drop associated with the transition to chaos, Fig.~\ref{fig:robustness_li}E and Fig. S3E reveal another striking pattern. As discussed earlier, for tasks requiring memory and non-linear processing, a non-zero $J$ is necessary. Consistent with this view, for the collection of tasks defined by our benchmark, the average performance of the rate‑based networks peaks at an intermediate value $J<J_{c}$. However, even at this optimal $J$, homogeneous networks still lag far behind heterogeneous systems that have no recurrent communication at all ($J=0$). This suggests that enriching the dynamical repertoire by tuning the recurrent gain $J$ is far less effective than the enrichment provided by heterogeneous neurons.

Further, we investigated how the structure of the recurrent connectivity influences performance, by varying the network’s sparsity $p$, the excitatory‑to‑inhibitory population ratio $f$, and the recurrent‑weight dispersion $\sigma_{0}$. For a fair comparison, each of these factors is swept while preserving the overall excitatory-to-inhibitory balance of the network. Note that this balance between inhibitory and excitatory populations is controlled by the mean of the population‑specific weights and is independent of their variance (dispersion). The variance, however, determines how strictly Dale’s law is obeyed. Low weight dispersion means that synapses originating from a given population have similar polarity (all inhibitory or all excitatory), as in the classical formulation of Dale’s law \cite{yang2020}. In contrast, high weight dispersion allows a population to project synapses of mixed polarity, even though the average weight conforms to the population’s identity--effectively implementing a “soft” Dale’s law. Figures  S4B–D and S5B–D summarize the effects of changing these recurrent‑related hyperparameters. Interestingly, neither sparsity nor weight dispersion affect performance and heterogeneous networks outperform their homogeneous version.

Note that in general network performance depends on the sampled data, resulting in a level of uncertainty in the measured performance. To capture this uncertainty, we trained three independent readouts for each network using non‑overlapping time intervals (trials), evaluated their performance on the same test set, and used the variance of the performance scores as a proxy for uncertainty. We repeated this procedure for all network configurations and tasks, and we summarize the results in Fig. S13 and Fig. S14 for the rate‑based and spiking systems, respectively. In both figures, heterogeneous networks exhibit performance variance that is comparable to or lower than that of homogeneous networks; however, in all cases the uncertainty is significantly lower than the measured performance.

\subsection{Heterogeneity saves energy on various computing systems}\label{sec:energy}
In the previous section we showed that heterogeneity enables even small, weakly coupled networks to perform well. Consequently, using such simple heterogeneous networks--instead of large, densely coupled homogeneous ones--can potentially reduce energetic cost without compromising performance. Energy minimization is a key objective for artificial, biological, and emerging neuromorphic systems. Therefore, we defined and estimated hardware‑dependent energy costs for our rate‑ and spiking‑based implementations, as well as for two neuromorphic chips (Loihi 2 and SpiNNaker 2). The energy calculations are described in Section A, and the emulation on the neuromorphic hardware is described in section \ref{sec:neuromorph_replication}.

\begin{figure}[ht]
	\centering
	\includegraphics[width=0.9\linewidth]{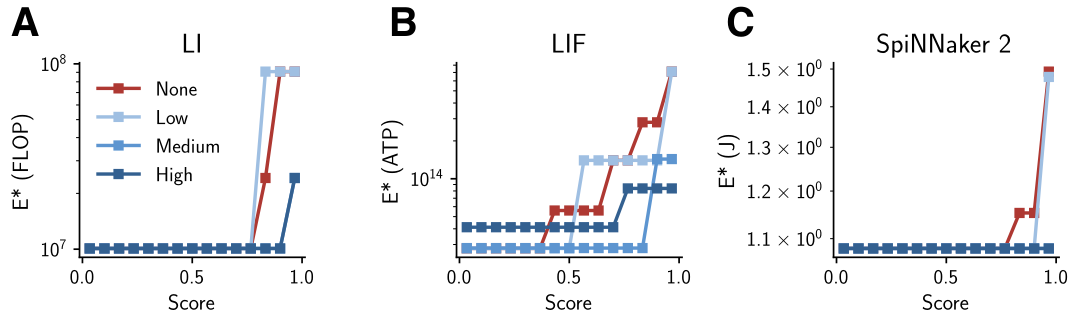}
	\caption{
    \textbf{Heterogeneity lowers the cost at each performance level.} 
    At every performance bin, the network with minimal energy consumption ($E^{*}$) is the most efficient. In all systems, heterogeneity substantially reduces the total minimal cost (note the logarithmic scale of the y‑axis). This cost saving primarily results from using small but performant networks. The definition of cost for each hardware system reflects its specific characteristics. We used \textbf{(A)} the total number of floating‑point operations (FLOPs) and the memory footprint for rate‑based networks, and \textbf{(B)} the total expenditure of adenosine triphosphate (ATP) molecules in biological spiking networks. \textbf{(C)} For the neuromorphic hardware system SpiNNaker 2, the total energy consumption (in joules) was estimated based on direct measurements of their power usage. A network with an equivalent firing rate was simulated for 100 steps, and the energy expenditure was linearly extrapolated to match the number of time steps used in the CPU simulations (cf. Section A). Note that the traces associated with lower heterogeneity are partially occluded by the one of high heterogeneity level.
    }
	\label{fig:energy}
\end{figure}

Since our setup is designed to gauge the effect of task‑agnostic heterogeneity across a task family, a fair metric of cost saving also needs to aggregate the cost incurred by all tasks within that family. 
Hence, for each heterogeneity level we first pooled the performances of networks of various sizes simulated for a fixed horizon. We then identified the performance range obtained by those networks, and finally found the network with minimum cost \(E^{*}\) within each score bin. In this way, one can clearly separate the cost profiles of networks with different heterogeneity levels across a range of tasks. The results for each heterogeneity level are summarized in Figs.~\ref{fig:energy}.

As Fig.~\ref{fig:energy} shows, the remarkable performance of small heterogeneous networks enables them to achieve high scores at only a fraction of the cost incurred by homogeneous networks--a result that holds across all considered computing systems. We found this also to be the case for networks that do not follow Dale's law (see Supplementary Fig.~\ref{fig:energy_nodale}, which additionally provides results for Loihi 2). Although higher performance levels typically require greater energy consumption (see also \cite{sengupta2013b, niven2016} for biological evidence), heterogeneity mitigates this increase by orders of magnitude.

To quantify this milder energy increase, we plotted the score of networks of various sizes and heterogeneity levels against their energy consumption for LI and LIF dynamics (cf. Figs.~\ref{fig:set_li}-\ref{fig:set_lif}). Since the energy improvement depends on task complexity, we plotted score-performance pairs for a single task from each complexity tier, as well as the average score-performance pair within the tier. In both cases, we observe that, on average, heterogeneous networks not only exhibit a higher score offset but also a smaller energy-to-score slope, particularly in LIF dynamics (cf. Fig.~\ref{fig:set_lif}), indicating a more graceful increase in energy consumption with performance.

\subsection{Task versatility through parameter diversification}\label{sec:mechanism}
\begin{figure}[ht]
	\centering
	\includegraphics[width=.9\linewidth]{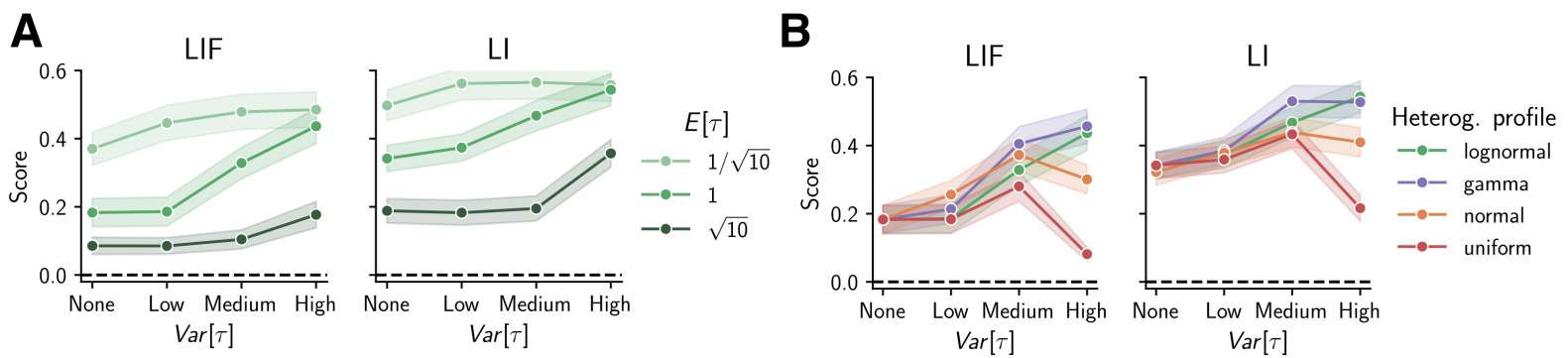}
	\caption{
    \textbf{The heterogeneity profile substantially affects task versatility in networks.}  
    \textbf{(A)} At any given value of $\E[\tau]$, heterogeneity improves performance. By spreading neuronal time constants over a broader interval, heterogeneity increases the likelihood of covering the regions required by the tasks. 
    \textbf{(B)} The performance depends on the heterogeneity profile. Skewed distributions (log‑normal and gamma) result in higher performance levels. The empirical mean and variance were identical for all profiles. The shaded area in (A) and (B) correspond to indicate 95\% confidence interval across all tasks.}
	\label{fig:ndistro}
\end{figure}

In our simulations so far, heterogeneity was imposed by controlling the dispersion of the time‑constant profiles. However, the dispersion foci—$\E[\tau]$ and the heterogeneity profile—were kept intact. Could choosing a different $\E[\tau]$ or profile modulate the benefit conferred by heterogeneity? We tested this possibility by repeating our experiments for various $\E[\tau]$ values and profiles, and summarize the results in Fig.~\ref{fig:ndistro}. This figure confirms the existence of such a modulation, but in a fairly nuanced manner, as detailed below.

As Fig.~\ref{fig:ndistro}A shows, both spike‑ and rate‑based RNNs achieve higher overall accuracy when \(\E[\tau] < 1\). Importantly, for a suitably chosen \(\E[\tau]\), even homogeneous RNNs perform well. Nevertheless, this does not diminish the advantage of heterogeneity, for two reasons. First, any given collection of tasks possesses an optimal range of time constant values; performance improves as the network’s parameters approach this range. Second, locating that optimal range requires a costly search over \(\E[\tau]\) and the heterogeneity profile. Heterogeneity sidesteps this search by providing a broad distribution of time constants, ensuring that some neurons are close to the optimal values without prior knowledge. It is precisely in this context that task‑agnostic parameter heterogeneity proves beneficial. As Fig.~\ref{fig:ndistro}A shows, only by diversifying neuronal resources across a wide range of values, without prior knowledge about the location of the optimal $\E[\tau]$ or fine-tuning for specific tasks, performance is increased in all conditions.

Nevertheless, the extent of the diversification benefit crucially depends on the type of tasks and the shape of the heterogeneity profile. As illustrated in Fig.~\ref{fig:ndistro}B, not all heterogeneity distributions have equal effects on the average performance of our task collection. While heterogeneity with long‑tailed profiles (gamma and lognormal) yields a performance improvement, this monotonic trend is absent for localized distributions (uniform and normal). Thus, beyond the presence of heterogeneity, its profile markedly influences the versatility achieved across a given set of tasks.

\subsection{Heterogeneity improves representation capacity}
\begin{figure}[ht]
	\centering
	\includegraphics[width=\linewidth]{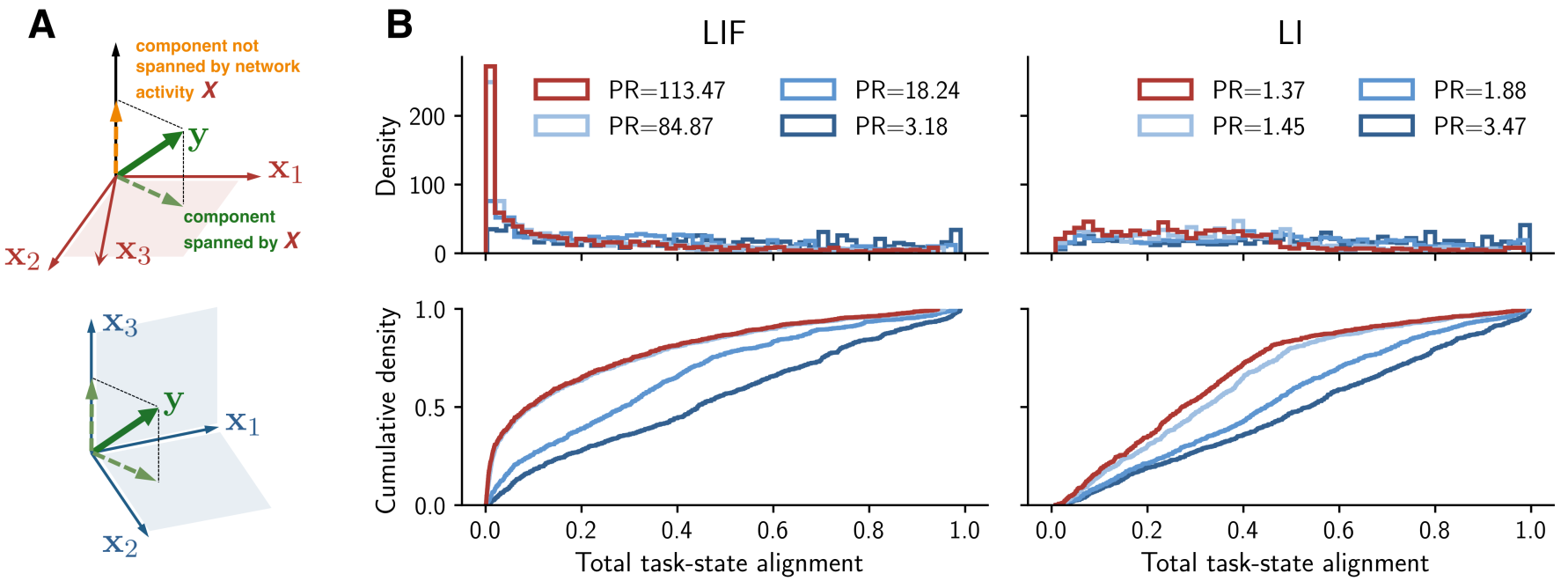}
    \caption{
    \textbf{Heterogeneity increases temporal alignment between tasks and the network’s state}  \textbf{(A)} Any temporal task $y_{\theta}(t)$ can be regarded as a high‑dimensional vector $\mathbf{y}$ in an appropriate vector space. The network states $\bm X(t)$ form a basis for that space. The extent to which the basis can span $\mathbf{y}$ determines the ultimate performance of a network for that task. Performance is perfect only if the task lies in the subspace spanned by the state (full state‑task alignment). A partial overlap results in suboptimal performance (partial state‑task alignment). \textit{(top)} A low‑dimensional (2D) span; \textit{(bottom)} a high‑dimensional (3D) span. Only the latter can fully represent the task $\mathbf{y}$.
    \textbf{(C)} Distribution of task‑state alignment for networks with different degrees of heterogeneity for \textit{(left)} LIF and \textit{(right)} LI models. \textit{(Top)} Total overlap histogram. \textit{(Bottom)} Cumulative overlap density. Higher scores in heterogeneous networks are corroborated by a systematic shift toward higher task‑state overlap across tasks. However, note that higher alignment does not necessarily mean a higher‑dimensional space spanned by the network. Direct assessment of the participation ratio (PR) indicates that while the state’s dimensionality increases with heterogeneity in the LI model, it decreases in LIF networks. Neural states are decorrelated before computing alignment. Colors indicate different heterogeneity levels, as in Fig.~\ref{fig:perf}.
    }
	\label{fig:overlap}
\end{figure}

Parameter diversification renders neurons' responses to the same stimulus less similar. The consequence of such dissimilar responses is a higher representation capacity in the neural space: more temporal functions can be approximated by the same neuronal substrate. Although it is not possible to derive an analytical expression for this representation capacity, it is nevertheless straightforward to quantify its effect empirically. Higher capacity means a higher temporal overlap between then neuronal states $\bm X(t)$ and the tasks $\{y_{\theta}(t)\}_{\theta}$ of our task family \cite{dambre2012, kubota2021} (cf. Fig.~\ref{fig:overlap}A).
Therefore, we measured this temporal alignment between all network–task pairs (using cosine similarity, see Sec.~\ref{sec:overlap}) and plotted the alignment distribution—as a measure of representation capacity—for networks with various heterogeneity levels in Fig.~\ref{fig:overlap}B. The systematic shift of the distributions toward higher values indicates that larger heterogeneity is associated with stronger task‑state alignment, and hence higher representation capacity.

It is also worth noting that, while a higher task‑state alignment is synonymous with a higher average performance, no direct link between performance and the dimensionality of the neural state exists (see Section C for a discussion). As illustrated in the legend of the same panel, the effective state dimensionality (quantified via the participation ratio) in spiking networks is much higher for homogeneous networks than for heterogeneous networks, whereas rate‑based networks show the opposite trend. In either case, however, as demonstrated earlier, heterogeneous networks outperform their homogeneous counterparts. This indicates that, in rate networks, higher performance is accompanied by a dimensionality expansion, whereas in spiking networks it is accompanied by a reduction of dimensionality.

\subsection{Demonstration on three real-world datasets}\label{sec:demo}

If neuronal heterogeneity renders computing systems versatile across a broad range of tasks, we should be able to observe its beneficial effect on specific tasks of high interest as well. Therefore, we sought this effect in three benchmark problems: the speech command (SC) classification task \cite{warden2018}, permuted sequential MNIST (psMNIST) \cite{goodfellow2015}, and contextual decision‑making (CDM) problem \cite{mante2013}. The SC benchmark contains densely sampled audio recordings of different speakers pronouncing 35 words under natural environmental conditions, posing a challenging task for all networks. However, the SC audio segments are short, making them unsuitable for measuring long‑range dependencies. This limitation is addressed by psMNIST, which introduces long‑term interactions through a random permutation. Finally, the CDM task is highly relevant to cognitive and systems neuroscience because it involves sensory integration, gating, memory, and competition between choices. Further details are provided in Sec.~\ref{sec:method_demo}.

\begin{figure}[ht]
	\centering
	\includegraphics[width=\linewidth]{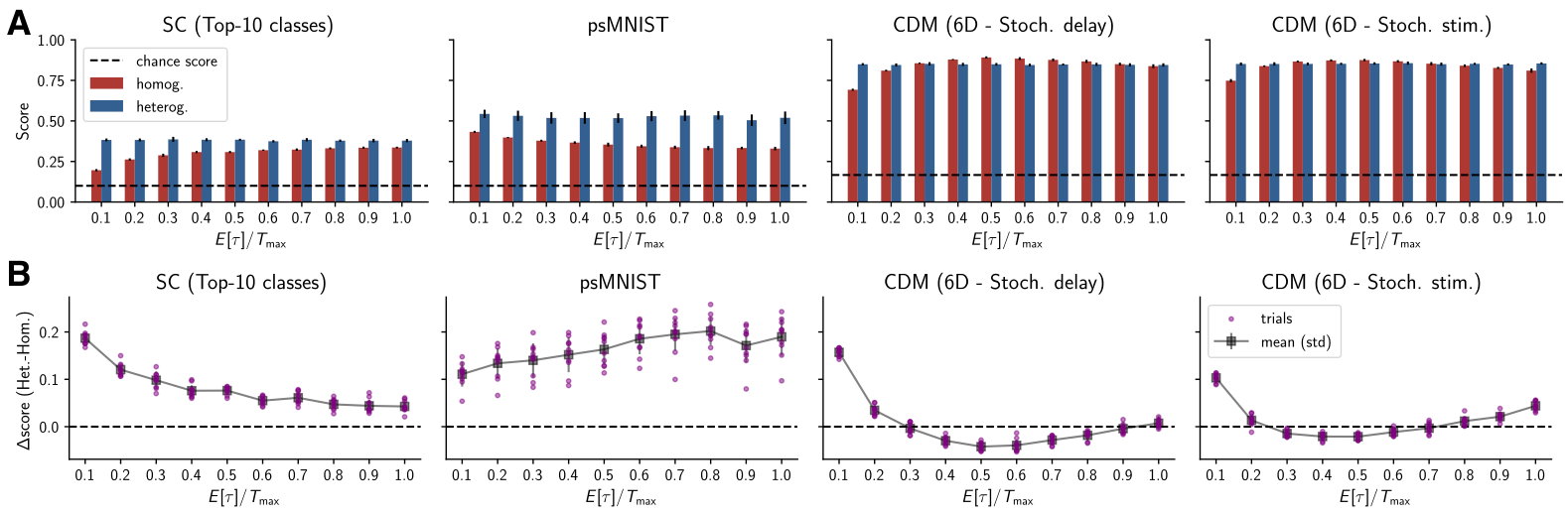}
	\caption{
    \textbf{Heterogeneity improves versatility in real-world problems.} Each system is composed only of $N=50$ fully decoupled rate neurons with $\tanh$ non-linearity to isolate manifestation of heterogeneous time constants. Heterogeneous time constants are distributed uniformly from 1 timestep to the maximal length of all trials $T_{\text{max}}$. The last 10 timesteps of neuronal trajectory are used a feature vectors of a multinomial logistic regression, with minimal regularization. \textbf{(A)} Mean scores and their standard deviations (error bars) across realizations. Bars heights and the error bars indicate respectively the mean and standard deviation of 10 independent realizations of feed-forward weights and time constants. \textbf{(B)} Score differences between the heterogeneous and homogeneous networks across 10 trials. Each dot represents on trial. The gray line, shows  average and standard deviations across trials. \textit{Columns}: from left to right: Speech commands (limited to samples with top 10 most frequent labels); Permuted sequential MNIST; Contextual decision-making task (with 6 choice) with stochastic delay and stochastic stimulation period}. 
	\label{fig:demo}
\end{figure}

Indeed, due to the universal approximation theorem for RNNs \cite{grigoryeva2018}, very large random networks, even if homogeneous, can solve these tasks. Thus, to highlight the effect of intrinsic neuronal heterogeneity, we considered a fairly extreme case: small networks ($N=50$) with no recurrent coupling ($J=0$). For simplicity, here we only compare a homogeneous rate network with a maximally heterogeneous one. Yet, as demonstrated in Sec.~\ref{sec:mechanism}, the choice of $\E[\tau]$ substantially affects the performance of homogeneous networks. Therefore initialized networks with various mean time constant values ranging from 10\% to 100\% of the maximum trial length $T_{\max}$. For heterogeneous networks, on the other hand, we allowed time constants to take any value from one timestep up to $T_{\max}$, further emphasizing our lack of knowledge about the optimal timescale in each task (cf. Sec.~\ref{sec:mechanism}). For each pair of network 10 different network realizations (feed-forward weights, and time constants) are generated. Classification is performed using the last ten timesteps of each trial, and the aggregated performance (mean and standard deviation) is summarized in Fig.~\ref{fig:demo}.

While in all cases the performance levels are far from perfect, such modest systems perform better than chance in every task. More importantly, the heterogeneous network consistently outperforms the homogeneous one on the SC and psMNIST tasks, which respectively highlight its greater capability to capture short‑ and long‑range interactions in the stimuli. This picture is not altered by other choices of temporal‑feature aggregation function or regularization (cf. Figs. S25 and S26).

For the CDM task, which requires distinct forms of computation (as opposed to the single memory form in the other two tasks), a more nuanced trend emerges. The peak homogeneous performance at low values of $\E[\tau]$ (i.e., fast input integration) indicates the suitability of smaller time constants for the CDM. Consequently, heterogeneous systems whose time constants are diversified \textit{outside} this optimal range underperform those homogeneous systems whose time constants falls within that interval. This is reminiscent of Fig.~\ref{fig:ndistro}B, where unsuitable heterogeneity profiles hamper performance. Nevertheless, when compared to homogeneous systems with large $\E[\tau]$ (i.e., low stimulus‑integration rate), heterogeneous networks perform better because some of their diversified time constants fall \textit{inside} the optimal range for the CDM task. Note that in all these tasks, the relatively decent performance of heterogeneous systems is achieved without fine-tuning, end-to-end optimization, or exhaustive search.

\section{Discussion}
Heterogeneity is a well‑documented feature of biological neural systems \cite{chen2009, zeng2017, seeman2018, hasse2019, chen2021, yao2021, campagnola2022, yao2023, langlieb2023}. Due to its significant computational role \cite{golomb1993, luccioli2010, shamir2006, tripathy2013, beiran2018, ostojic2009, perez2010, mejias2014, divolo2021, destexhe2022}, it seems conceivable that integrating heterogeneity into artificial systems yields similar computational advantages.
Such integration, however, can be achieved with two distinct approaches. The first, which has attracted some attention in recent years, introduces and tunes neural heterogeneity in a task‑specific manner \cite{matsubara2017, fang2021, perez-nieves2021, winston2023, habashy2024a, zheng2024}. While being highly effective for the specific tasks for which it is tuned, it is unclear whether such a task‑informed heterogeneity profile improves versatility across tasks. The heterogeneity observed in biological organisms is likely not tuned to a single task. Consequently, we treat heterogeneity as an intrinsic property of the neural substrate, as it is the case in biological organisms. The aim of this work is to investigate whether such a task‑agnostic heterogeneity structure can benefit or harm the holistic efficacy of neural systems across broad range of tasks.

To test whether intrinsic heterogeneity improves network versatility across tasks, we used the reservoir computing paradigm \cite{maass2000, jaeger2001}, in which the internal substrate is left intact during training, and compared the performance of homogeneous and heterogeneous networks across a battery of dissimilar tasks. By coupling chaotic stimuli with a parameterized nonlinear mapping, we curated a comprehensive benchmark of low‑bias and high discriminative power, consisting of hundreds of dissimilar tasks spanning a wide range of complexities (Fig.~\ref{fig:perf}). This benchmark allowed us to assess the role of neuronal heterogeneity in a task‑independent manner, addressing a key limitation of previous studies \cite{matsubara2017, fang2021, perez-nieves2021, winston2023, habashy2024a, zheng2024} that examined only a handful of tasks.

Our findings reveal that, in both rate‑ and spike‑coded networks, heterogeneous RNNs--despite possessing a task‑agnostic parameter organization--significantly outperform their homogeneous counterparts (cf. Fig.~\ref{fig:perf}), exhibit lower performance variability (Fig. S13- S18), and are more robust to a variety of internal and external perturbations (Fig.~\ref{fig:robustness_li},  S3, S4, and S5). 

The benefit of intrinsic heterogeneity stems from enriching the dynamical repertoire of neural activity, thereby increasing its representation capacity by forming a more expressive temporal filter bank. Although intrinsic heterogeneity also manifests in synaptic dynamics and connectivity, the most prominent factor in functional specialization is the diversity of neural responses \cite{duarte2019}.
. This principle, although long known to engineers (e.g., \cite{crochiere1983}, Ch.~7), is less appreciated by the connectionist approach, where representation capacity is increased by introducing recurrent connections and finely controlling their strength, e.g., to maintain memory \cite{hopfield1982}, to map inputs to outputs \cite{krizhevsky2017}, to embed a dynamical response \cite{laje2013}, etc. Nevertheless, neuronal heterogeneity alone, i.e., without recurrence, can exhibit a high representation capacity, as evidenced by the superiority of a decoupled heterogeneous system over homogeneous yet coupled networks in Fig.~\ref{fig:robustness_li}E.

Representation enrichment through neural heterogeneity is particularly relevant from an evolutionary perspective and by analogy to neuromorphic systems. By virtue of neuronal diversity, simple organisms, despite their simple nervous systems, can gain high versatility and robustness across a wide range of cognitive tasks, thereby increasing their survival chance (cf. Figures~\ref{fig:robustness_li}B and  S3B). Moreover, a task‑agnostic neuronal parameter distribution emancipates them from costly training, thereby reducing their metabolic use. Similarly, neuromorphic systems, which are typically limited by architectural complexity and intended to achieve low energy consumption, can benefit from intrinsic neuronal diversity to reduce network size and total energy usage (cf. Fig.~\ref{fig:energy}). 
Here, we used extrapolation to estimate the energy consumption of heterogeneous networks running on digital neuromorphic chips (cf. Section \ref{sec:energy}), indicating that heterogeneous networks may be beneficial in enabling more energy-efficient solutions implemented on these systems. Moreover, in certain neuromorphic systems, heterogeneity is intrinsic to the hardware and does not need to be explicitly programmed. Instead, one may capitalize on device‑to‑device variability caused by manufacturing imperfections \cite{wang2019b, covi2021, billaudelle2022}, rather than suppressing it, to construct robust systems. Furthermore, as Figures~\ref{fig:robustness_li}E and  S3E suggest, the relatively high representation capacity of even fully decoupled systems give neuromorphic devices with very simple architectures—and thus lower manufacturing and quality‑assessment costs— the potential to perform as effectively as chips with highly integrated units.

It is worth stressing that the benefit of intrinsic time‑constant heterogeneity should be understood across tasks, not per task. Although relatively rare, there are multiple occurrences (e.g., in Figs.~\ref{fig:perf} and \ref{fig:demo}) of homogeneous systems outperforming the heterogeneous ones. This is expected and aligns with the No Free Lunch theorem \cite{wolpert1997}. When a specific task is of utmost interest (as opposed to many) and sufficient training resources are available, an end‑to‑end optimization is almost certainly a better solution (e.g., compare our task-unspecific scores in \ref{fig:demo}A-B with the scores reported in \cite{yin2021a}). Thus, the applicability of task‑agnostic heterogeneous time constants ultimately boils down to the trade‑off between the scope of tasks a system is expected to perform and its training budget. For a system with a narrow set of goals, it might be best to optimize the heterogeneity (as well as synaptic parameters) for those particular tasks. On the other hand, for multipurpose systems with a limited training budget, a task‑agnostic parameter initialization (with no further training) might be a more economical solution.

Nevertheless, even for single tasks, heterogeneous parameters can serve as a good initialization for the subsequent optimization process, as previously evidenced in \cite{perez-nieves2021}. This is particularly true for scenarios in which the optimal range of timescales is unknown or when many distinct timescales are involved in the task. Indeed, for tasks with both bounded (cf. Fig.~\ref{fig:demo}) and unbounded (cf. Fig.~\ref{fig:perf}) timescales, performance achieved by our heterogeneous initializations was often better than (or at least comparable to) a finely tuned homogeneous initialization. However, the optimal form (i.e., its profile, mean, and higher‑order statistics) depends on the specifics of the task under consideration (cf. Fig.~\ref{fig:ndistro}).

In this study we exclusively used standard reservoirs as our underlying RNN. A highly important factor in the functionality of reservoirs is the scaling of recurrent and feedforward weights \cite{lukosevicius2009, cucchi2022}. This is the case because neither projection is optimized in a standard reservoir; thus, choosing them incorrectly might shrink the dynamical range of neurons, pushing them toward full saturation or complete silence. While the scaling of recurrent weights can be readily controlled by the spectral radius of the connectivity (to induce fading memory), the correct scaling of the feedforward projection is less trivial. This is particularly true for inputs, where the relative amplitude of different channels is of semantic or functional importance. A prime example is audio processing, where different channels indicate the power content within different frequency bands, or our CDM task, where the amplitude of different channels suggests the level of coherence in the stimulus (see Sec.~\ref{sec:method_demo} for details). In that case, preprocessing of the input is as crucial as scaling the feedforward weights and controlling heterogeneity, and shall not be overlooked.

Moreover, standard reservoirs, being fading‑memory machines, are by construction unsuitable for certain classes of tasks such as precise autonomous signal generation (e.g., \cite{sussillo2009, laje2013}), indefinite‑memory tasks (e.g., peak‑hold \cite{boyd1985}, associative memory \cite{hopfield1982}), or transitions across fixed points (e.g., \cite{sussillo2012, nachstedt2017a}). Reservoirs approximate the mapping between an external input and the target via their transient dynamics. Hence, whenever a task can be decomposed into transient modes, a reservoir‑based framework—and thus heterogeneity—can potentially be applicable. For example, the psMNIST task requires a transient response with an extremely long memory trace, SC needs a quick response to distinct frequency bands, and CDM necessitates the existence of nonlinear units that integrate over time. Since all of these tasks encapsulate some form of computation through dynamics, they can be cast into the reservoir computing framework and thus benefit from time‑constant heterogeneity, as shown in Fig.~\ref{fig:demo}.

In general, using neural models or connectivity structures rather the one of standard reservoirs, unlocks new computational possiblities \cite{lukosevicius2009}. The heterogeneous-time constant reservoirs presented here is one such instantiations. Yet, synaptic time constants \cite{quax2020, kozachkov2022}, multiplexing \cite{appeltant2011,appeltant2014}, structured \cite{pellegrino2023, kawai2023} or hierarchical connecitvity \cite{pandarinath2018,maes2020}, or feedback pathways \cite{sussillo2009}, when conjugated with a learning algorithms, demonstrate superior performance compared to the standard network configurations. Therefore, it remains to be answered whether diversifying these non-cannonical parameters and architecture can result in superior performance without end-to-end training.

It is important to emphasize that the definition of the time constant used in this study is based on the cell’s electrophysiological properties (namely its capacitance and leak conductance). This definition should not be confused with the other common notion of a time constant that is defined operationally, based on the decay rate of the autocorrelation function of the cell’s activity \cite{cavanagh2020, zeraati2023, stern2023}. This choice was inspired by the experimental protocol used in a recent public physiological dataset \cite{aisynphys2019} that quantified the intrinsic properties of a cell using long, sustained stimuli \cite{aisynphys_method}. Indeed, as shown previously in \cite{perez-nieves2021} and verified by us, such activity‑independent neuronal time constants follow a long‑tailed distribution, which incentivized us to use a log‑normal profile for the time constant for the majority of this work. In contrast, the distribution of time constants defined via the latter operational definition is influenced by the statistics of the external input, circuit structure, and synaptic dynamics \cite{vlachos2012, dubreuil2022, stern2023}. Although, under certain conditions, a similar skewed time‑constant profile can arise \cite{stern2023}, the computational benefit of such activity‑dependent heterogeneity is yet to be investigated. The specialization of neural responses may also arise from imposing constraints on the neural code. For instance, by minimizing stimulus encoding error, \cite{zeldenrust2021} demonstrated the emergence of two neuron types (integrator and resonator). It is therefore plausible that different variations of predictive coding (or similar constraints) give rise to distinct optimal heterogeneity profiles.

While heterogeneity in time constants enhances the robustness and accuracy of multi-tasking, it is important to emphasize that heterogeneity alone does not eliminate the need for domain adaptation. As highlighted in Section~\ref{sec:robustness} and \ref{sec:demo}, the base time scale and input gains must still be adapted to match the specifics of a given task (e.g., psMNIST, SC, or CDM). Therefore, heterogeneity should not be viewed as a one-size-fits-all solution.

While here we focused on heterogeneity of the membrane time constant, diversity in other temporal parameters—such as synaptic delays, synaptic response kinetics, postsynaptic‑receptor turnover, vesicle‑recycling rates, and so on—may produce similar effects. For example, \cite{sun2023a, sun2025} demonstrated successful execution of temporal tasks by optimizing axonal delays, \cite{winston2023} investigated the same problem by tuning the after‑spike current, and \cite{habashy2024a} showed that both delays and neuronal time constants can be equivalently tuned to achieve optimal performance. Moreover, the benefit of heterogeneity is not limited to the temporal domain. A classic example is the heterogeneity in orientation selectivity of cells in visual cortex, enabling animals to detect edges of all possible orientations \cite{hubel1962, hubel1968}. Similar principles have been proposed for the olfactory system \cite{rospars2014, ualiyeva2024}, auditory system \cite{kuo2009, kanold2014, vasquez-lopez2017, tao2017}, and somatosensory system \cite{gatto2019, han2023}, as well as for spatial encoding via grid cells \cite{redman2025}. Additionally, the role of cellular and subcellular heterogeneity is not limited to high‑level cognitive functions. For instance, a recent study shows that perturbing the regulators of cell‑type diversity leads to developmental deficiencies \cite{khouri-farah2025}. These observations, together with the results presented here, further support the hypothesis of intrinsic heterogeneity serving as a vital biological feature—rather than an undesirable imperfection—endowing organisms with versatility and robustness across a wide variety of tasks and conditions.

A limitation of our work is its reliance on a single task family. Although this family can produce a variety of working‑memory‑like computations, it is constrained to target functions that attend to single time point only. Consequently, it cannot capture temporal interactions, gating mechanisms, or integrative processes. While we partially mitigated this shortcoming by demonstrating our approach on three independent benchmarks selected specifically to address these additional computational aspects (cf. Sec.~\ref{sec:demo}), a systematic replication of the study using a wider array of task families would be highly informative.

In conclusion, this work elucidates the functional role of neuronal heterogeneity at the network level, with substantial implications for neuroscience, machine learning, and neuromorphic computing. This pronounced effect underscores the importance of incorporating parameter variations when studying the mind and when designing brain‑inspired devices and algorithms.

\section{Methods}\label{sec:method}
\subsection{Networks configuration}\label{sec:reservoir}
Each network consists of \(N\) neurons, each belonging to either an excitatory (E) or inhibitory (I) subpopulation. A fraction \(f\) of the neurons are excitatory. Neurons are connected randomly and sparsely with connection probability \(p\). We do not fix the in‑ or out‑degree of any neuron; consequently, the number of synapses fluctuates around the expected value \(Np\). Autapses are prohibited. The \(K\)-dimensional feedforward input is projected onto the network via fixed, dense, random connections \(w^{u}_{ik}\), whose values are drawn from a standard normal distribution. 

Within the network, the synaptic weight from neuron \(j\) to neuron \(i\) is denoted by \(w_{ij}\) and remains constant. Its value is drawn from a population‑specific Gaussian distribution with mean \(\mu_{x}\) and variance \(\sigma_{x}^{2}\) (\(x\in\{E,I\}\)); the sign of the weight determines its polarity (excitatory or inhibitory). The means of these Gaussian distributions are chosen so that the network is in balance \cite{vreeswijk1996, brunel2000}, i.e.:
\begin{equation}\label{eq:balance}
f\mu_E + \left(1-f\right)\mu_I = 0.
\end{equation}

Note that, unlike \cite{rajan2006}, our networks are in a loose balance, which aligns better with measurements in the cerebral cortex \cite{ahmadian2021} (see also \cite{deneve2016}). For simplicity, the synaptic‑weight variance $\sigma^{2}_{x}$ is kept the same for both populations:
\begin{equation}\label{eq:dispersion}
\sigma^2_I = \sigma^2_E = \sigma^2_0.
\end{equation}
We refer to this as weight dispersion. As in \cite{vreeswijk1998, brunel2000}, to ensure that input fluctuations are independent of network size, recurrent weights are scaled by \(J/\sqrt{Np}\), where \(J\) is a global recurrent gain that modulates the strength of internal connections. Similarly, feedforward weights are scaled by \(J_{u}/\sqrt{K}\), which makes the variance of the external input independent of its dimensionality while allowing the overall input strength to be adjusted via \(J_{u}\).  Finally, each neuron receives independent and identically distributed (i.i.d.) white noise, whose intensity is controlled by \(J_{n}\).

\subsection{Stimulus synthesis}\label{sec:stim}
For the majority of our simulations, we used the solution of the Lorenz system \cite{lorenz1963}, described by
\begin{align}\label{eq:lorenz}
	\dot x &= \sigma (y-x), \\
	\dot y &= x(\rho - z) - y, \\
	\dot z &= xy - \beta z.
\end{align}

The Lorenz system is a well‑known chaotic system; with parameters $\rho=28$, $\sigma=10$, and $\beta=8/3$, its trajectory lies on a chaotic attractor. To minimize transient behavior, we chose the initial conditions $(x_0, y_0, z_0)=(-1.96582031,\,-1.08886719,\,2.17578125)$, a point close to the stable manifold.

A second stimulus was generated by concatenating three one‑dimensional time series obtained from Mackey–Glass system \cite{mackey1977}, a delayed differential equation given by:
\begin{equation}\label{eq:mg}
	\dot x = \frac{ax(t-\delta)}{1+x(t-\delta)^n} - bx(t).
\end{equation}
For the commonly used parameter values $a=0.2$ and $b=0.1$, delay parameters $\delta\lesssim16.5$ produce periodic orbits, whereas $\delta\gtrsim17$ generate chaotic trajectories \cite{wernecke2019}. The full stimulus was constructed using three delay values, $\delta=10$, $50$, and $80$. In all cases we set the initial condition $x_{0}=1.2$ and sampled a random history uniformly from $[1.1,1.3]$ for the preceding time steps up to $-\delta$.

The trajectories of both the Lorenz and Mackey–Glass systems were generated by numerically integrating their respective equations using the Runge–Kutta 4/5 (RK45) method. 

We also employed two additional stimuli with high and very low predictability, respectively: a periodic input \(u(t)=|\sin t|\) and the nonlinear autoregressive moving‑average (NARMA) model \cite{atiya2000}. The NARMA system is a discrete‑time dynamical system described by
\begin{equation}\label{eq:narma}
	u[t+1] = a_1u[t] + a_2\sum_{i=0}^{n-1} u[t] u[t-i] + b\xi[t-(n-1)] \xi[t] + c
\end{equation}
with $\xi$ being a uniform random variable in $[0,\,0.5]$ and $n$ indicating the order of the model. We used parameters $a_{1}=0.2$, $a_{2}=0.04$, $b=1.5$, $c=0.001$, zero initial conditions, and a model order of $n=30$. This discrete signal is then interpreted as samples of a continuous stimulus with time step $\Delta t$, i.e., $u[i]=u(i\Delta t)$. For all stimuli, values at arbitrary times were obtained by linear interpolation.

The choice of parameters in the dynamical systems above affects the amplitude and temporal characteristics of the resulting trajectories. Thus, for a fair comparison each component was first standardized and then temporally rescaled so that the overall timescale of the compound stimulus equals $1$, i.e., the average membrane time constant of our networks. Concretely, we computed the power spectrum of each component independently, identified the frequency with maximal power, and defined a compound frequency as the geometric mean of these peak frequencies. Time was then uniformly scaled for all components such that the compound frequency becomes $1\,$Hz. With this scaling we ensure that the network and the input, on average, operate on similar time scales, although either system can still exhibit much shorter or much longer temporal dependencies.

\subsection{Neuronal dynamics}\label{sec:dyn}
We implemented two neuronal dynamics: the rate‑based leaky integrator (LI) and the spiking leaky integrate‑and‑fire (LIF) models.

For the LI dynamics, the membrane voltage of the \(i\)‑th neuron, \(v_i\), evolves according to
\begin{align}\label{eq:dyn-li}
	\tau_i \frac{dv_i(t)}{dt} &= -v_i(t) + \frac{J}{\sqrt{Np}}\sum_{j=1}^N w_{ij} r(v_j(t)) + \frac{J_u}{\sqrt{K}}\sum_{k=1}^{K} w^u_{ik} {u_k}(t) + J_n \xi_i(t), \\
    r(x) &= [1 + \exp(-x)]^{-1}. \label{eq:af}
\end{align}
Here $\tau_i$ is the neuron‑specific membrane time constant, $r(\cdot)$ is a static sigmoidal nonlinearity (activation function), $u_k(t)$ is the $k$‑th component of the external stimulus, and $\xi_i(t)\sim\mathcal N(0,1)$ is neuron‑specific white noise. Throughout this work we use a strictly positive sigmoidal activation function, although other nonlinearities are possible. The function $r(\cdot)$ is interpreted as the (max‑)normalized firing rate (hence unitless), whereas the recurrent and feedforward weights $w_{ij}$ and $w_{ik}^{u}$ have the same units as the membrane potential $v_i$. Importantly, the gains $J$ and $J_{n}$ scale the recurrent input and noise amplitude, respectively.

For the LIF model the membrane voltage of the $i$‑th neuron evolves according to  
\begin{align*}
	C_i \frac{dv_i(t)}{dt} &= g_L (E-v_i(t)) + I_\text{bg} + \frac{g_\text{rec} }{\sqrt{Np}}\sum_{j=1}^N w_{ij} s_j(t) + \frac{g_\text{ff}}{\sqrt{K}}\sum_{k=1}^{K} w^u_{ik} {u_k}(t) + g_L J_n^2\xi_i(t)
\end{align*}
$C_i$ denotes the neuron‑specific membrane capacitance, while the reversal potential $E$, the leak conductance $g_{L}$, and the recurrent ($g_{\text{rec}}$) and feedforward ($g_{\text{ff}}$) conductances are assumed to be identical across neurons. A constant background current $I_{\text{bg}}$ is introduced to make the LIF dynamics comparable to the LI dynamics (see below). The term $s_{j}(t)$ represents the post‑synaptic potential triggered by presynaptic spikes.  A LIF neuron emits a spike when its membrane voltage reaches the threshold $v_{\text{thr}}$, after which the voltage is reset to $v_{\text{reset}}$. To cap the maximal possible activity--analogous to the saturating activation function in the rate model--the membrane potential is held at $v_{\text{reset}}$ for an absolute refractory period $\tau_{\text{ref}}$.  These dynamics can be summarized by the following set of equations:
\begin{align}\label{eq:dyn-lif}
	\tau_i \frac{dv_i(t)}{dt} &= -v_i(t) + (v_\text{bg} - E) + \frac{J}{\sqrt{Np}}\sum_{j=1}^N w_{ij} s_j(t) + \frac{J_u}{\sqrt{K}}\sum_{k=1}^{K} w^u_{ik} {u_k}(t) + J_n \xi_i(t), \\
    s_j(t) &= \tau_\text{ref} \sum_{f} \delta(t-t_j^f), \label{eq:synaptic_model}\\
    t_j^f  &= \arg v_j(t) = v_{\text{thr}}, \label{eq:crossing}\\
	v_j(t) &= v_\text{reset},  \hspace{2cm} t\in (t_j^f, t_j^f+ \tau_\text{ref}], \label{eq:reset}
\end{align}
where the neuron‑specific membrane time constant is \(\tau_i = C_i/g_L\), and the global recurrent and feedforward gains, \(J = g_{\text{rec}}/g_L\) and \(J_u = g_{\text{ff}}/g_L\), correspond to their rate‑based analogs. For computational efficiency we model synapses as instantaneous elements (Eq.~\ref{eq:synaptic_model}); however, temporally extended synaptic kernels (e.g., \(s_j(t)=\kappa(t)*\sum_f \delta(t-t_j^{f})\) with standard kernel functions) are equally valid.
The factor $\tau_{\text{ref}}$ in Eq.~\ref{eq:synaptic_model} ensures that the voltage increment produced by a single spike during the refractory period $\tau_{\text{ref}}$ matches the increment generated by the rate model with a saturating activation function over the time interval.

The dynamics of LI and LIF neurons can be matched using several strategies, each requiring coordinated tuning of different parameters. Here we focus on aligning the dynamical range of the two neuron models, as described shortly. For alternative approaches, see Section B.

The voltage of a LIF neuron is bounded above by the threshold $v_{\text{thr}}$ (i.e., $v_i\le v_{\text{thr}}$), whereas the voltage of an LI neuron is unbounded. In both cases, however, only a limited voltage interval carries the relevant coding information. For a LIF neuron this interval is $[v_{\text{reset}},\,v_{\text{thr}}]$, and for an LI neuron with the sigmoidal activation function of Eq.~\ref{eq:af} it is $[-0.5,\,0.5]$. Any pair of values that differ by one unit (e.g., a 1‑mV difference) between $v_{\text{reset}}$ and $v_{\text{thr}}$ satisfies this condition. We arbitrarily chose  $v_{\text{reset}} = E = -70\ \text{mV}$ and $v_{\text{thr}} = -69\ \text{mV}$ and set the absolute refractory period to $\tau_{\text{ref}} = 2\ \text{ms}$. The background input $v_{\text{bg}}$ is tuned so that, at rest, the LIF neuron fires at a baseline rate $\nu_0 = 5\ \text{Hz}$. This input is neuron‑specific and can be computed directly from Eq.~\ref{eq:nu2v}.
\begin{align}\label{eq:nu2v}
    v_{\text{bg}, i} (\nu)&= \frac{(v_\text{reset}-E) z_i - (v_\text{thr} - E)}{z_i-1}, \\
    z_i &= \exp\Big( \frac{1}{\nu \tau_i} -\frac{\tau_\text{ref}}{\tau_i} \Big)
\end{align}
which is found by solving the generic form of $\nu-I$ relation (presented in Eq. S9) for the background current. 

\subsection{Integration details}\label{sec:integration}

All networks were initialized at zero and integrated with the explicit Euler method using a time step of  $\Delta t = \frac{\E[\tau]}{100}=0.01$. This step size balances numerical accuracy (it is sufficiently small relative to the neuronal time constants) with computational efficiency (it is large enough to make long simulations feasible). Only for a few networks did this step size produce inaccurate integration, which was diagnosed by an anomalous performance profile across all tasks. In those cases the system was reintegrated with a Runge–Kutta integrator, and the dense solution was sampled at intervals of $\Delta t$ to maintain the same sampling rate across all networks.

All networks are integrated for $L = L_{\text{train}} + L_{\text{test}}$ time steps, and the results are saved to disk for later training and evaluation. For LIF networks, we saved the spike trains. For LI networks, the neuronal voltages were stored with 16‑bit floating‑point precision to reduce the storage load (compared with the 64‑bit precision used during the simulations). Although some information is lost during this conversion, we anticipate that its effect on performance will be minimal, as demonstrated in artificial systems \cite{gernigon2023, wang2018c} and computational studies \cite{benna2016}.

\subsection{Datasets, optimization, and performance quantification}\label{sec:fit}
The lengths of the training ($L_{\text{train}}$) and test ($L_{\text{test}}$) subsets must be sufficiently long to provide enough samples for reliable read‑out optimization and an unbiased assessment of within‑distribution generalization. Assuming that the tasks evolve on an average timescale $\tau_y$, we require the training and test intervals to span at 20 and 10 \textit{oscillations} of that timescale, respectively. Moreover, because larger networks contain more learnable parameters, we scale the training duration proportionally to the number of neurons. With these criteria, the numbers of time steps in the training and test subsets are, respectively, 
\begin{align}\label{eq:subset_N}
	N_{\text{train}} &= (N+1)\,\frac{20\tau_y}{\Delta t},\\
    N_{\text{test}}  &= \frac{10\tau_y}{\Delta t}.
\end{align}
However, because of the structure of our tasks, the training and test subsets must be longer. We evaluate the networks on memory‑recall and forecasting tasks up to a maximal temporal horizon of $2\tau_{u}$. Consequently, for every time point in the training and test sets, the target function must be available $2\tau_{u}/\Delta t$ steps both in the past and in the future. Hence, the total lengths of the subsets are

\begin{align}\label{eq:subset_L}
	L_{\text{train}} &= N_{\text{train}} + \frac{4\tau_u}{\Delta t} = (N+1)\frac{20\tau_y}{\Delta t} + \frac{4\tau_u}{\Delta t},\\
    L_{\text{test}}  &= N_{\text{test}} + \frac{4\tau_u}{\Delta t} = \frac{10\tau_y}{\Delta t} + \frac{4\tau_u}{\Delta t}.
\end{align}
Since all tasks are generated from a stimulus whose overall timescale is $\tau_u = 1$ (see Section~\ref{sec:stim}), we assume that the temporal variation of the target inherits this same timescale. Consequently, we set $\tau_y = 1$.

In Section~\ref{sec:energy}, to ensure that energy consumption arises solely from differences in network size, all simulations were run with a fixed training set length of 100 (in units of the base time scale). For the test set, as before, a fixed length of 10 was used for all networks.

The training and test datasets are constructed by pairing the network states with the corresponding target functions. For rate‑based networks, the state \(\bm X(t)\) is obtained by passing each neuron's membrane voltage through the activation function \eqref{eq:af}. For spiking networks, the spike trains are first convolved with an exponential kernel (see Eq.~\ref{eq:synaptic_model}) 

\begin{equation*}\label{eq:syn_conv}
    \phi(t)=\exp\!\bigl(-t/\tau_{\phi}\bigr),\qquad \tau_{\phi}=10\Delta t,
\end{equation*}
to produce a continuous‑time representation of the state. We varied the kernel’s time constant (over \(\tau_{\phi}\in[5,100]\,\Delta t\)) and kernel shapes (alpha and Gaussian). In none of these cases did we observe any qualitative difference in performance profiles. 

All 882 time‑aligned target functions are stacked into the matrix $\bm Y(t)$, yielding a dataset $\mathcal{D}=\{\bm X(t),\bm Y(t)\}$ for each network. The dataset is then split into training and testing subsets of lengths $L_{\text{train}}$ and $L_{\text{test}}$, respectively. From each subset we extract the central $N_{\text{train}}$ and $N_{\text{test}}$ samples and temporally shuffle them to remove transient effects, producing the state–target pairs $\mathcal{D}_{\text{train}}=\{\bm X_{\text{train}}, \bm Y_{\text{train}}\}$ and $\mathcal{D}_{\text{test}}=\{\bm X_{\text{test}},\bm Y_{\text{test}}\}$.

The optimal linear read‑out \(\bm\beta_{\theta}\in\mathbb{R}^{N+1}\) for the state matrix \(\mathbf X_{\text{train}}\in\mathbb{R}^{(N+1)\times N_{\text{train}}}\) (including the bias) that approximates a task with parameters \(\theta\) (i.e., \(\mathbf y_{\theta,\text{train}}\in\mathbb{R}^{N_{\text{train}}}\)) is obtained by minimizing the objective function
\begin{equation}\label{eq:ridge}
	\mathcal L_{\theta}(\bm \beta) = ||\textbf y_{\theta, \text{train}} - \bm \beta^T \textbf X_{\text{train}} ||^2 + \lambda ||\bm \beta||^2,
\end{equation}
where the regularizer \(\lambda = 10^{-6}\) is introduced solely to avoid singular matrices. We also fit an intercept to accommodate targets with a non‑zero mean. The analytical solution of problem \ref{eq:ridge} is 
\begin{equation}\label{eq:sol}
	{\bm \beta}_{\theta} = \arg\min_{\bm \beta} \mathcal L_{\theta}(\bm \beta) = (\textbf X_{\text{train}}\textbf X_{\text{train}}^T  + \lambda \textbf I)^{-1} \textbf X_{\text{train}} \textbf y_{\theta, \text{train}}.
\end{equation}

The performance of a network on a given task is quantified by the coefficient of determination of the read‑out \(\bm{\beta}_{\theta}\) on the test set:
\begin{equation}\label{eq:coeff-det}
	s_{\theta} = 1 - \frac{||\bm y_{\theta,\text{test}} - \hat {\bm y}_{\theta,\text{test}}||^2}{||\bm y_{\theta,\text{test}} - \bar {\bm y}_{\theta,\text{test}}||^2},
\end{equation}
where $\hat{\mathbf y}_{\theta}= \bm\beta_{\theta}\,\mathbf X_{\text{test}}$ and $\bar{\mathbf y}_{\theta,\text{test}}$ denote the network’s prediction and the target mean, respectively. The scalar $s_{\theta}$ therefore quantifies the normalized mismatch between the prediction and the ground truth. A perfect match yields $s_{\theta}=1$, $s_{\theta}=0$ corresponds to chance‑level performance (i.e., predicting only the target mean), and negative values indicate performance worse than chance (approaching $-\infty$ for infinitely poor predictions).

For the classification problems in Sec.~\ref{sec:demo}, we used multinomial logistic regression. Specifically, for each sample $i \in \{1, \dots, n\}$, a linear combination of neural activity $\mathbf{X}_{\text{train}}$ (features) is computed to produce a probability predictor $\eta_{ic}$ for each class $c \in \{1, \dots, C\}$, and the logistic link function maps this predictor to class probabilities (cf. \cite{james2023}, Chapter 4, or \cite{bishop2006}, Sections 4.3.2–4.3.4):

\begin{align}
  \eta_{ic} &= \boldsymbol{\beta}_c \cdot \boldsymbol{x}_i + \beta_{c,0} \label{eq:predictor} \\
  \hat p_{ic} &= \frac{\exp(\eta_{ic})}{\sum_{k=1}^{C} \exp(\eta_{ik})} = \frac{\exp(\boldsymbol{\beta}_c \cdot \boldsymbol{x}_i + \beta_{c,0})}{\sum_{k=1}^{C} \exp(\boldsymbol{\beta}_k \cdot \boldsymbol{x}_i + \beta_{k,0})} \label{eq:pred_prob}
\end{align}

The regression coefficients are then obtained by minimizing the negative log-likelihood (assuming independence of samples), plus a regularization term:
\begin{equation}\label{eq:ll}
  -\log \mathcal{L}(\boldsymbol{\beta}) = -\frac{1}{n} \sum_{i=1}^n \sum_{c=1}^C \mathbf{1}[y_i = c] \log(\hat p_{ic}) + \frac{1}{2\lambda} \|\boldsymbol{\beta}\|_F^2
\end{equation}
Here, $\mathbf{1}[\cdot]$ denotes the indicator function, $n$ is the number of training samples, and $\lambda$ is the regularization strength. Unlike in linear regression, the coefficient matrix $\boldsymbol{\beta} \in \mathbb{R}^{N \times C}$ has dimensions corresponding to $N$ features and $C$ classes. We used \texttt{scikit-learn}'s implementation of multinomial logistic regression, which fits an overparameterized model by estimating coefficients for all $C$ classes rather than only $C-1$ independent ones. This overparameterization is effectively managed through regularization.

\subsection{Uncertainty quantification}\label{sec:uq}
There are two main sources of uncertainty in the performance $s_{\theta}$: variations in the training data and variations in the network realization (connectivity, weights, and time constants). To estimate the former, we expanded the training set threefold and assessed the performance of three independent readouts for each task (cf. Figs. S13–S14). For the latter, we created 10 network realizations, with fully independent connectivity, feedforward and recurrent weights, and time constants (cf. Figs. S15–S18). Score variation due to either source was extremely small, even for small networks of size 50.

Note that score variation across different tasks (cf. Fig.~\ref{eq:lorenz}) can be viewed as a third source of uncertainty. This deterministic source of variability, however, depends on the task definition as well as the input, and should not be conflated with the sources of uncertainty discussed above.

Scores reported in the main text are the averages over these three independent readouts for a single realization.

\subsection{Complexity and similarity definition}\label{sec:complexity}

Task complexity is a network‑independent measure that quantifies the non‑triviality of the mapping from the input $\bm{u}(t)$ to the target output $y_{\theta}(t)$. Several absolute metrics could be used, e.g., mutual information between input and output, the Wasserstein distance, Jeffreys divergence, the degree of non‑linearity (expressed via the multi‑index of multivariate polynomials), or-- for the analytical mappings considered here-- the number of elementary operations required to transform $\bm{u}(t)$ into $y_{\theta}(t)$. For simplicity, however, we adopt a relative measure by comparing each task to the simplest task in our set. A natural choice for this reference task is the identity function

\begin{equation}\label{eq:complexity_ref}
    y_{0}(t)=u_{k}(t),
\end{equation}
which simply relays the $k$‑th component of the input without any modification. For a generic task $y_{\theta}(t)=u_{k}(t+\Delta)^{d}$, we define its complexity as the cosine dissimilarity to the reference:

\begin{equation}\label{eq:complexity}
	C = 1 - \Big|\frac{\bm y_{\theta}^T \cdot \bm y_0}{\norm{\bm y_{\theta}} \norm{\bm y_0} }\Big|.
\end{equation}
A perfect alignment with the reference yields $C=0$, whereas tasks orthogonal to the reference achieve the maximal complexity $C=1$.

Cosine similarity is also used to quantify the similarity between pairs of tasks (Fig.~\ref{fig:perf}, second column). Specifically, the similarity between tasks $\theta$ and $\theta'$ is defined as the cosine of the angle between their centered vectors:
\begin{equation}\label{eq:similarity}
	S_{\theta \theta'} = \frac{(\bm y_{\theta} - \bar{\bm y}_{\theta})^T \cdot (\bm y_{\theta'} - \bar{\bm y}_{\theta'})}{ \norm{\bm y_{\theta} - \bar{\bm y}_{\theta}} \norm{\bm y_{\theta'} - \bar{\bm y}_{\theta'}} }.
\end{equation}
Subtracting the mean eliminates static offsets, ensuring that tasks that are dynamically independent but share the same average are not mistakenly identified as similar.

\subsection{Overlap and dimensionality definition}\label{sec:overlap}

The temporal overlap between a network’s state $\bm X(t)$ and a task $y_{\theta}(t)$ is computed in a similar way. First, we orthogonalize the state by applying principal‑component analysis (PCA) to obtain $\hat{\bm X}(t)$, which captures $99.9\%$ of the variance. Then, we evaluate the cosine similarity between the mean‑subtracted task $y_{\theta}(t)-\bar{y}_{\theta}$ and each orthogonalized neuronal component $\hat{x}_i(t)$:
\begin{equation}\label{eq:suboverlap}
	\cos \alpha_{\theta, i} = \frac{(\bm y_{\theta} - \bar{\bm y}_{\theta})^T \cdot \hat{\bm x}_i} { \norm{\bm y_{\theta} - \bar{\bm y}_{\theta}} \norm{\hat{\bm x}_i} }.
\end{equation}

Because the states are orthogonalized, the total overlap between task $\theta$ and the state $\bm X(t)$ is the sum of the squared partial overlaps:
\begin{equation}\label{eq:overlap}
	\text{Overlap}\big(y_{\theta}(t), \bm X(t)\big) = \sum_{i=1}^{N} \cos^2 \alpha_{\theta, i}.
\end{equation}

Note that this definition of overlap is closely related to the notion of \textit{information processing capacity} (IPC) introduced in \cite{dambre2012}. In short, IPC decomposes the state into orthonormal polynomials (e.g., Hermite, Legendre, Laguerre). The $i$‑th decomposition coefficient--equivalent to the cosine similarity between the normalized state and the $i$‑th basis function--determines the capacity of the state to perform the processing type specified by basis $i$. Because the polynomial bases are orthogonal by construction, IPC provides a complete quantification of a system’s ability to carry out independent types of processing. While \cite{dambre2012} proved that the IPC of any dynamical system is bounded by the number of its dynamical units, \cite{kubota2021} showed that, when such systems are used to reconstruct a task, networks perform best when the IPC of the system and that of the task are maximally similar. In other words, reconstruction error is minimized when the network’s state contains exactly the same processing required by the task.  These studies rely on the existence of well‑defined Hilbert spaces in which both the task and all neuronal states can be represented \cite{lemaitre2010}. There are many ways to span a Hilbert space; IPC adopts the convenient basis that disentangles different processing types from one another. However, because performance in reservoir computers depends only on the quality of task reconstruction, it suffices to measure the extent to which the (incomplete) basis set formed by the neural states spans a given task. Forming a complete basis set--whether the IPC basis or any other--is unnecessary.  Also note that, since we always add a constant term to our state $\bm X(t)$ during regression, the mean value of every task can always be captured. Consequently, the non‑trivial overlap must be computed between the states and the mean‑subtracted values of the task.

For dimensionality we use the participation ratio as a proxy because it is computationally simple and robust \cite{gao2017, recanatesi2020}. Let the centered state be  $\bm X_{c}(t)=\bm X(t)-\mathbb{E}\bigl[\bm X(t)\bigr]$, and denote its singular values by \(\{\sigma_i\}\). The participation ratio is then defined as
\begin{equation}\label{eq:pr}
	PR = \frac{\big(\sum_{i}{\sigma_i^2})^2}{\sum_{i}{\sigma_i^4}},
\end{equation}
where $\sigma_i$ denotes the $i$‑th singular value. For a system of $N$ neurons recorded over $T > N$ time steps, $PR$ is bounded between $1$ (all variance concentrated in a single component) and $N$ (all units contribute equally to the variance) \cite{gao2017, stringer2019}. A higher participation ratio reflects less regular, more “jagged” neural trajectories. However, this irregularity is constrained by the regularities of the driving input \cite{stringer2019} and by the dynamics of the system \cite{gao2017}. The low‑dimensional neural states in Fig.~\ref{fig:overlap}B are therefore likely due to the Lorenzian input, whose fractal dimension is very close to $2$ \cite{kuznetsov2020}.

\subsection{Emulation of network dynamics on neuromorphic hardware}\label{sec:neuromorph_replication}
We investigated energy efficiency using two state‑of‑the‑art neuromorphic platforms, Loihi 2 and SpiNNaker 2. Loihi 2, developed by Intel Labs, implements networks of spiking neurons directly in hardware, enabling rapid processing of complex data while consuming very little energy \cite{davies2021, shrestha2024, pierro2025}. In contrast, SpiNNaker 2, a neuromorphic computing platform developed by TU Dresden and the University of Manchester, achieves high efficiency and flexibility by running neural networks on asynchronously communicating general‑purpose ARM processors \cite{hoppner2022, nazeer2024, chen2025}. Both systems exhibit exceptional energy efficiency--particularly for time‑series processing tasks \cite{yan2021, davies2021, vogginger2024, shrestha2024, lux2024, pierro2025}--and are therefore poised to contribute strongly to more sustainable high‑performance computing as well as to edge‑device computing.

To emulate the dynamics of our simulated networks on neuromorphic hardware, we first computed the mean firing rate of each neuron in the CPU‑based simulations. Then, on both SpiNNaker 2 and Loihi 2, we instantiated networks with the same connectivity as in the corresponding CPU implementation and imposed the calculated mean firing rates \(\nu_{\text{set}}\) on each neuron by applying a neuron‑specific background input (cf. Eq.~\ref{eq:dyn-lif} and \cite{dayan2001a}, p.~164):

\begin{equation}\label{eq:neuromorphic_emulation_voltage}
v_\text{bg} = \frac{v_{\text{thr}}}{1 - \exp(- 1 / (\nu_\text{set} \cdot \tau_\text{m}))},
\end{equation} 
where $v_{\text{thr}}$ is an arbitrarily chosen firing threshold of $0.01\unit{V}$ and $\tau_{\text{m}}$ is a membrane time constant of $0.01\unit{s}$. In addition, all synaptic weights-- including those of existing connections-- are set to zero. This common approach ensures that the full computational load associated with relaying spikes is accounted for \cite{luboeinski2025}, while spikes do not affect the imposed firing rates $\nu_{\text{set}}$. Consequently, because neurons and synapses in the recurrent network perform exactly the same number of operations as they would in a complete simulation, we obtain fair estimates of runtime and energy consumption. Note that the specific choice of membrane time constants is irrelevant here, since the neuron‑specific inputs already incorporate the effects of time‑constant heterogeneity. On both Loihi 2 and SpiNNaker 2, we emulated the full range of network sizes $N$ considered in the study.

\subsection{Cost estimation}\label{sec:cost_estimation}
Since a network’s performance depends on its state $\bm X(t)$, we estimate the energy cost of generating those states. This cost, however, is hardware‑dependent, so a distinct definition of cost is used for each implementation. Moreover, we separate the cost into a static (activity‑independent) component and a dynamic (activity‑specific) component. The total energy cost of the state is therefore the sum of its static and dynamic components. The definitions of these components for the different implementations are described in Supplementary Section A.

\subsection{Details of the real-world tasks}\label{sec:method_demo}

In each task, two sets—one training and one testing—were generated. Each dataset contained a tuple consisting of the trial‑concatenated input $\bm u(t)$, the vector of correct label for all trials $\bm y$, and the trial lengths. The input $\bm u(t)$ is streamed to the network. At the beginning of each trial, the network state is reset to zero using the “fixation” cue embedded in the input, either as an additional channel (e.g., in SC, CDM) or as a distinct value within the same input channel (psMNIST). Fixation lasts for one timestep only.

The inputs from each benchmark undergo a benchmark‑specific and a benchmark‑independent preprocessing procedure. The former is detailed in Sec. H, while the latter step is described here. To ensure that all input trials perturb the network’s state similarly and that each input channel has a comparable effect on a given neuron, inputs are trial‑normalized and the feedforward weights are scaled. Specifically, irrespective of the length of each trial, the trial‑specific segment of the input $\bm u(t)$ was normalized to have a total norm of 1. Furthermore, to equalize the effect of different channels on a given neuron, the magnitude of the feedforward weights was row‑normalized by the standard deviation of the input channels over the training period. The input gain $J_u$ was then adjusted for each benchmark to ensure the neuron's state goes well beyond the linear regime of the activation function (defined customarily as activations larger than 0.5). Gain values between 1 and 100 were used.

Networks were integrated using the forward Euler method, producing a voltage trajectory that was then passed through a $\tanh$ activation function. These activations were pooled over the time domain to create a feature space for each trial. We considered the following pooling strategies:
\begin{itemize}
    \item \textbf{Go}: The last 10 timesteps of the activation traces are concatenated. This is analogous to receiving a “Go” cue at the end of each trial in a cognitive or psychological experiment, although we did not feed such an stimulus to the network explicitly. 
    \item \textbf{Average}: The average activation over the entire course of the trial is used as the feature vector for that trial.
    \item \textbf{Snapshot}: Ten equally spaced snapshots of the activation trajectory, taken from 10\% to 100\% of the trial length, are concatenated to form the trial features. 
\end{itemize}

For the psMNIST and SC datasets, samples were loaded randomly using a seeded random number generator (\texttt{TorchGenerator} object). The fixed permutation of the MNIST dataset was performed using an independent seeded random number generator (Numpy's \texttt{BitGenerator}), which was called only once. The same seed (1234) was used for both generators. In both datasets, we used 3200 samples for training and 3200 samples for testing. While no restrictions were imposed on the labels in the MNIST dataset, the Speech Commands dataset was limited to the 10 most frequent labels.

Classification is performed using multinomial logistic regression, i.e., class probabilities are regressed jointly against one another. We used the \texttt{scikit-learn} library with the Newton conjugate-gradient solver. Regularization is employed to (i) remove the null subspace of the Hessian matrix, rendering the optimization problem strictly convex; (ii) resolve the softmax gauge‑freedom ambiguity that arises from the symmetric implementation of logistic regression in \texttt{scikit-learn}; and (iii) guarantee existence and uniqueness of the weights when perfect class separation occurs (otherwise the weights diverge to infinity). As in the regression problem of Sec.~\ref{sec:fit}, the regularization effect must be relatively small; otherwise the solver only minimizes a small weight norm. For a fair comparison, both heterogeneous and homogeneous networks are read out using the same regularization. The effect of pooling and regularization choices is shown in Sec J. Label‑prediction accuracy was used as the performance measure, and is defined as 
\begin{equation}
    \text{Accuracy} = \frac{\text{num. correctly predicted labels}}{\text{num. samples}}.
\end{equation}

The error in the difference in Fig.~\ref{fig:demo}B is estimated using the error propagation formula:
\begin{align}
    z &= s_{het} - s_{hom} \\
    (\Delta z)^2 &= (\Delta s_{het})^2 + (\Delta s_{hom})^2.
\end{align}
which is valid due to the independence of our realizations.

\section{Other software resources}
Throughout this work, we employed free and open‑source libraries (apart from the proprietary components required for the neuromorphic hardware frameworks). The rate‑based networks were implemented in Python using \verb|NumPy| \cite{harris2020} and \verb|SciPy| \cite{virtanen2020}, and parallelized with \verb|Joblib| \cite{thejoblibdevelopers2025}. The \verb|Brian| simulator \cite{stimberg2019} was used for the CPU implementation of spiking networks. For visualization, we used \verb|Matplotlib| \cite{hunter2007} and \verb|Seaborn| \cite{waskom2021} (and, optionally, \verb|SciencePlots| \cite{garrett2023}).

To measure the energy and computational efficiency of the spiking model on neuromorphic hardware, we employed the following software packages: for SpiNNaker 2, \verb|py-spinnaker2| (forked from commit 12c3828), \verb|s2-sim2lab-app| (commit 4cd3a06), and \verb|spinnaker2-stm| (commit b449cca); for Loihi 2, the \verb|Lava‑Loihi| package (release version 0.7.0).

For the demonstrations in Sec.~\ref{sec:demo}, we used \texttt{torchaudio} (version 2.11.0) \cite{yang2022,hwang2023}  and 
\texttt{torchvision} (version 0.27.0) \cite{torchvision2016}  packages from the \texttt{pytorch} (version 2.12.0) ecosystem \cite{paszke2017}. Only CPU releases where used. Furthermore, \texttt{neurogym} package (version 2.2.0) \cite{crocioni2026} was used to create an environment for the CDM task.

\section{Code availability}

All scripts used for CPU simulations are publicly accessible at \cite{golmohammadi2026} or at \url{https://gitlab.gwdg.de/golmohammadi/hetero}, with version tag v1.2 corresponding to the current revision. Detailed documentation for environment setup and usage examples is available on a dedicated documentation page at \url{https://hetero-a61646.pages.gwdg.de/}.

\section{Data availability}
We downloaded the MNIST dataset from \url{http://yann.lecun.com/exdb/mnist/} using the TorchVision library. The Speech Commands dataset (version 0.02) was downloaded from \url{http://download.tensorflow.org/data/speech_commands_v0.02.tar.gz} using the TorchAudio library. For both datasets, the original train/test splits were used. The validation set was omitted for Speech Commands. All other datasets were generated synthetically.

\section{Acknowledgments}
We would like to thank Dr. Sebastian Schmitt, Dr. Andrew Lehr, Dr. Michael Fauth, Ms. Shirin Shafiee, and Ms. Nahid Safari for fruitful discussions.

\section{Funding}
This research was funded by the German Research Foundation via project no. 536022519 and SFB1286, projects C01 and Z01, as well as by the German Federal Ministry of Education and Research (BMBF), grant no. 01IS22093A "KISSKI".

\section{Authors contribution}
AG and CT conceptualized the study. AG designed the simulations and analysis pipeline. JL performed simulation and energy measurements on neuromorphic chips. CT acquired resources and funding. AG wrote the original draft. All authors wrote, reviewed, and edited the manuscript.

\section{Competing Interests}
The authors declare no competing interests.


\printbibliography
\end{refsection}

\clearpage
\begin{refsection}
\setcounter{page}{1}
\begin{supplementary}
\appendix

\section{Cost estimation details}\label{sec:method_energy}
\subsection{Rate implementation}
In the rate‑based networks, neuronal and synaptic activity is represented by numbers with a given floating‑point precision (e.g., 32‑ or 64‑bit). Consequently, generating the network state requires manipulating and storing these numbers throughout the simulation. We therefore define the total number of floating‑point operations (FLOP) and the memory footprint (MEM) as the primary contributors to the overall cost. These quantities correspond to the dynamic and static costs, respectively.

Following the notation introduced in Section~\ref{sec:reservoir}, a network with $N$ neurons, $N^{2}p$ recurrent synapses, and $KN$ feed‑forward synapses that receives a $K$‑dimensional stimulus has costs that scale as follows:
\begin{align}\label{eq:cost_rate}
	E_\text{mem} &= \underbrace{\mathcal{O}(N)}_\text{\shortstack{state}} + \underbrace{\mathcal{O}(K)}_\text{\shortstack{input}} + \underbrace{\mathcal{O}(N^2 p)}_\text{\shortstack{rec. weights}} + \underbrace{\mathcal{O}(NK)}_\text{\shortstack{ff. weights}}+ \underbrace{\mathcal{O}(1)}_\text{\shortstack{shared time constants}} &\quad \text{(homogeneous)}\\
	E_\text{mem} &= \underbrace{\mathcal{O}(N)}_\text{\shortstack{state}} + \underbrace{\mathcal{O}(K)}_{\text{\shortstack{input}}} + \underbrace{\mathcal{O}(N^2 p)}_{\text{\shortstack{rec. weights}}} + \underbrace{\mathcal{O}(NK)}_{\text{\shortstack{ff. weights}}}+ \underbrace{\mathcal{O}(N)}_{\text{\shortstack{distinct time constants}}} &\quad \text{(heterogeneous)}\\
	E_\text{flop} &= \underbrace{L}_{\text{\shortstack{timesteps}}} \Big( 
    \underbrace{\mathcal{O}(N)}_{\text{\shortstack{volt. update}}} + \underbrace{\mathcal{O}(N^2p)}_{\text{\shortstack{sum. rec. input}}} + \underbrace{\mathcal{O}(NK)}_{\text{\shortstack{sum. ff. input}}} 
    \Big) 
    &\quad \text{(homog. or heterog.)}
\end{align}
Note that, although heterogeneous networks incur a slightly larger memory cost, the FLOP cost is identical for homogeneous and heterogeneous networks. Moreover, while $E_{\text{flop}}$ scales with the number of timesteps $L$, the memory footprint $E_{\text{mem}}$ does not depend on $L$, because the allocated memory remains constant throughout the simulation. The total number of timesteps is set to $L = L_{\text{train}} + L_{\text{test}}$, as defined in Section~\ref{sec:fit}. Since $L$ is scaled with the network size $N$, smaller networks are additionally less costly due to their shorter simulation time.

\subsection{Spiking implementation}
For our spiking networks, we estimated the metabolic cost of an equivalent biological network in terms of the total consumption of adenosine triphosphate (ATP) molecules. This enables us to quantify the benefit of heterogeneity in real neural tissue. Our estimates follow the framework of \cite{attwell2001, howarth2012}, which partitions the total neuronal energy budget into signaling and housekeeping components. The signaling cost itself is further decomposed into five categories:

\begin{enumerate}
    \item \textbf{Resting.} Even at rest neurons consume energy to maintain their membrane reversal potentials. This is achieved by the Na\textsuperscript{+}/K\textsuperscript{+} pump, which continuously extrudes three Na\textsuperscript{+} ions and imports two K\textsuperscript{+} ions; each exchange consumes one ATP molecule. Using the Na\textsuperscript{+} and K\textsuperscript{+} currents at the resting potential together with the ionic charge, the total rate of ion exchange can be estimated. ATP is consumed by both neuronal and glial cells. The estimated consumption rates are  
    \begin{itemize}
        \item Neuron: $102\,\text{M ATP}\,\text{s}^{-1}\,\text{cell}^{-1}$
        \item Glia: $342\,\text{M ATP}\,\text{s}^{-1}\,\text{cell}^{-1}$
    \end{itemize}
    
    \item \textbf{Action‑potential generation and propagation.} To generate an action potential (AP) the membrane voltage must increase by roughly $50\text{–}100\,$mV (depending on the neuronal compartment). Given the specific membrane capacitance and the volume of the different compartments (axon, dendrite, and soma), one can compute the net positive charge that must be injected to produce such a voltage change. However, at any instant both positive and negative currents flow across the membrane. Consequently, the actual number of ions exchanged depends on the kinetics of the channels, specifically the overlap between Na\textsuperscript{+} and K\textsuperscript{+} currents. In the cerebral cortex, this overlap increases the total ion exchange by about $24\%$ (according to \cite{howarth2012}, which updated the previously over‑estimated value used in \cite{attwell2001}) compared with an idealized AP with no overlap \cite{sengupta2010}. The resulting ATP usage per compartment is  
    \begin{itemize}
        \item Unmyelinated segments: $97.5\,\text{M ATP}\,\text{AP}^{-1}$
        \item Cell body: $5.1\,\text{M ATP}\,\text{AP}^{-1}$
        \item Dendrites: $16.2\,\text{M ATP}\,\text{AP}^{-1}$
    \end{itemize}
    
    \item \textbf{Presynaptic.} A presynaptic AP raises intracellular Ca\textsuperscript{2+} concentration, which must be restored by Na\textsuperscript{+} exchange. The elevated Ca\textsuperscript{2+} also triggers vesicle processing and recycling (docking, fusion, exo‑ and endocytosis), which consumes ATP, albeit far less than the processes above. The estimated ATP cost per vesicle is  
    \begin{itemize}
        \item Ca\textsuperscript{2+} influx resetting: $12\,\text{k ATP}\,\text{vesicle}^{-1}$
        \item Presynaptic vesicle recycling: $400\,\text{ATP}\,\text{AP}^{-1}$
    \end{itemize}
    
    \item \textbf{Postsynaptic.} Neurotransmitter release opens postsynaptic receptors for a finite time, allowing excess Na\textsuperscript{+} ions to enter the postsynaptic neuron; these ions must be removed by the Na\textsuperscript{+}/K\textsuperscript{+} pump. Different receptor types admit different amounts of Na\textsuperscript{+} because of their distinct abundance and electrophysiological properties. The best available estimates, which take the relative abundance of receptors into account, give the following costs  
    \begin{itemize}
        \item non‑NMDA: $67\,\text{k ATP}\,\text{vesicle}^{-1}$ (average)
        \item NMDA: $70\,\text{k ATP}\,\text{vesicle}^{-1}$
        \item mGlu: $3\,\text{k ATP}\,\text{vesicle}^{-1}$
    \end{itemize}
    
    \item \textbf{Glutamate recycling.} Glutamate molecules that are not taken up by postsynaptic receptors diffuse in the extracellular matrix and are recycled by glial cells. The ATP cost for glutamate uptake (by glia), intracellular processing, export to the neuron, and repackaging into a vesicle is approximately $1.1\times10^{4}$ ATP per glutamate molecule. 
\end{enumerate}

\cite{attwell2001, howarth2012} estimate that, on average, each vesicle contains about 4000 glutamate molecules. Combining the contributions above, the total ATP‑consumption rate of a network with $N$ neurons, $N_{\text{glia}}$ glial cells, and $N^{2}p$ synapses that fire at an average rate $\nu$ with neurotransmitter‑release probability $r$ is
\begin{align}\label{eq:cost_spike}
	\dot E(N, p, \nu, J, N_{glia}, r) &= [(102 \times 10^6 \ \frac{\mathrm{ATP}/s}{\mathrm{glia}}) N_{\mathrm{glia}} + (342 \times 10^6 \ \frac{\mathrm{ATP}/s}{\mathrm{neuron}}) N] & (\text{resting}) \\ 
						 &+ [(120\times 10^6 \ \frac{\mathrm{ATP}}{\mathrm{spike}}) N\nu] & (\text{AP gen. and prop.})\\
						 &+  [(12.4\times 10^3 \ \frac{\mathrm{ATP}}{\mathrm{release}}) N^2p \nu r]\sqrt{J} & (\text{presynaptic}) \\
						 &+  [(140\times 10^3\ \frac{\mathrm{ATP}}{\mathrm{release}}) N^2 p \nu r]\sqrt{J} & (\text{postsynapic}) \\
						 &+  [(11\times 10^3\ \frac{\mathrm{ATP}}{\mathrm{release}}) N^2 p \nu r]\sqrt{J} & (\text{glutamate recycling}). \label{eq:glut} 
\end{align}
In the formula above we assumed that the synaptic strength \(J\) modulates the pre‑ and post‑synaptic terms equally. In our calculations we set \(N_{\text{glia}} = N\) (cf. \cite{vonbartheld2016}) and \(r = 1\) because all synapses in the simulations were deterministic, although \cite{attwell2001} use the more realistic fusion probability \(r = 0.25\) (2000 out of 8000 boutons release neurotransmitter). The firing rate \(\nu\) is replaced by the temporal average measured in the simulation. Finally, the expression is multiplied by the total simulation time \(T = L\Delta t\) to obtain the signaling energy.

In addition to the signaling energy, the housekeeping cost--comprising the subcellular machinery and cellular homeostasis--contributes to the total ATP expenditure. It is estimated that housekeeping accounts for an additional one‑third of the total signaling‑related energy \cite{attwell2001, howarth2012}.

Since the housekeeping energy scales with activity, the total dynamic energy cost of the biological networks is taken as \(\tfrac{4}{3}\) of the sum of Equations \ref{eq:cost_spike}–\ref{eq:glut}, whereas the activity‑independent energy expenditure corresponds only to Equation \ref{eq:cost_spike}.

\subsection{Neuromorphic implementation}
We measured both the total execution time and total energy consumption of the workload (cf.~\ref{sec:neuromorph_replication}) on each neuromorphic board--containing either a single SpiNNaker 2 or a single Loihi 2 chip--and repeated the measurements over ten trials.  To separate static (constant‑overhead) from dynamic (workload‑related) energy costs, we first measured the total energy consumption of idle networks by simulating stimulus‑free networks with the same neurons and synapses that never fire. This baseline energy reflects the hardware‑related consumption and was taken as the static cost. Next, we emulated our spiking networks using the background input described in Section~\ref{sec:neuromorph_replication} for the same duration, measured the total energy again, and subtracted the idle energy for each network to estimate the activity‑dependent dynamic power consumption.

Note that, because spike readout and logging power measurements significantly slowed down the simulations on both chips, networks were simulated for only 100 timesteps. To enable a consistent comparison with our training and inference implementations, both static and dynamic energy values were upscaled by a factor of $L/100$, where $L$ is the intended number of timesteps for each network. Please note that these extrapolated values constitute an upper limit on the energy consumption of longer runtimes, as the initialization of the simulation, which is necessarily included in the measurement, has a relatively larger impact on the energy consumption of short runtimes such as 100 timesteps compared to longer runtimes. It has not been possible to simulate for the same number of timesteps as in the CPU simulations, since the memory of neither of the two neuromorphic chips is sufficient for the energy tracking and spike recording across such long runtimes (cf. \cite{loihi2techbrief,hoppner2022}) -- as recording the spikes is necessary for an eventual full implementation of the model, it may not be missing in our emulation. Therefore, the provided numbers must be considered estimates that can serve as a guideline for neuromorphic implementations in the future. In this regard, our results indicate that using heterogeneous networks would bring further benefits for the energy efficiency on the neuromorphic chips Loihi 2 and SpiNNaker 2, which have already been demonstrated to enable highly energy-efficient execution of spiking neural networks.

Also note that we did not aim to develop an architecture‑independent energy model; rather, we employed an energy‑estimation strategy tailored to our tasks and network configurations. For a more thorough investigation of neuromorphic energy consumption, platform‑wide benchmarks such as those in \cite{ostrau2022,yik2025} can be employed.  

For power analysis of the SpiNNaker 2 and Loihi 2 chips, we used the SpiNNaker2 STM Python Interface \cite{spnkr_stm_python_interface}, which is publicly available, and Intel’s Lava-Loihi \cite{lava2025} framework, access to which was granted by means of the Intel Neuromorphic Research Community. Access to the chips was provided by the respective manufacturers.

\section{Matching the single-neuron input–output curves of the spiking and rate models}\label{sec:snn2rnn}
To match the dynamics of rate‑ and spike‑based neurons, we must ensure that the input–output (i.e., current–rate) relationship of the two models is comparable. An LIF neuron receiving a constant background input  \(I>\dfrac{v_{\text{thr}}-v_{\text{reset}}}{R}\)  possesses a maximally normalized equilibrium firing rate given by  

\begin{equation}\label{eq:lif_rate}
    r_s (I)= \nu/\nu_{\max} = \frac{1}{1 + \frac{\tau_m}{\tau_{\text{ref}}} \log\big(\frac{RI - (v_{\text{reset}}-E)}{RI - (v_{\text{thr}}-E)}\big)},
\end{equation}
where the parameters are those introduced in Section~\ref{sec:dyn}. This expression is the generic form of Eq.~\ref{eq:neuromorphic_emulation_voltage}.  Similarly, for a rate neuron with the activation function defined in Eq.~\ref{eq:af}, the (normalized) firing rate at equilibrium--i.e., when \(v(t)=I(t)R\)--is  
\begin{equation}
    r_{r}(I)=\frac{1}{1+\exp(-IR)}.
\end{equation}

\subsection{Global or local match}
Assuming a fixed membrane time constant and refractory period, matching the dynamics corresponds to finding a set of parameters \((E,\,v_{\text{reset}},\,v_{\text{thr}})\) that yields \(r_{s}\approx r_{r}\) for all values of \(I\), i.e.,
\begin{equation} \label{eq:snnrnn_opt}
    \min_{\{E, v_{\text{thr}}, v_{\text{reset}}\}} \mathcal L = ||r_r(I) - r_s(I) ||^2.
\end{equation}
Figure~\ref{fig:snnrnn} shows the firing‑rate curves of both models and their difference as a function of the input current. The minimization problem above yields a globally optimal solution. However, because the spiking neuron's firing rate converges logarithmically to the asymptotic line $r=1$, this global optimization places excessive weight on reducing the discrepancy at high‑$I$ values. In practice, $I$ is bounded, so the problem should be solved over the (application‑dependent) domain of $I$ rather than over the entire real line.
\begin{figure}[!h]
	\centering
	\includegraphics[width=.5\linewidth]{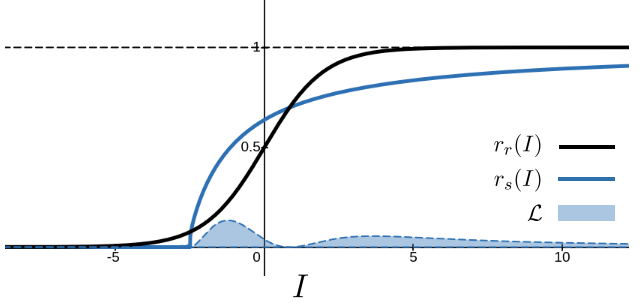}
    \caption{\textbf{Matching a spiking neuron to its rate‑based counterpart}. Normalized firing rates of the spiking neuron (blue) and the rate neuron (black) are shown together with their difference, $\mathcal{L}(I)$, as a function of the input current $I$. The LIF neuron parameters are set to $\tau_{\text{ref}}=2\,\text{ms}$, $\tau_{m}=10\,\text{ms}$, $R=1\,\Omega$, $v_{\text{thr}}-E=2.8\,\text{mV}$, and $v_{\text{reset}}-E=2.5\,\text{mV}$. $\mathcal{L}$ is multiplied by two in the plot to highlight the slow convergence at large $I$ values.}
	\label{fig:snnrnn}
\end{figure}

Note that if the membrane time constants are heterogeneous, solving the optimization problem \ref{eq:snnrnn_opt} yields neurons with heterogeneous thresholds, reversal potentials, and reset potentials.

Please note that, for homogeneous networks, it is also possible to match the dynamics of rate and spiking networks at the population as opposed to neural level \cite{chang2025}. However, since such a mapping is not guaranteed in the heterogeneous case, we adopted the simpler approach described above.

\subsection{Matching the mean activity}
One can aim to match the baseline resting activity of the two neurons. This imposes the condition $r_s(0)=0.5$, which, together with Eq.~\ref{eq:lif_rate}, yields the following relationship among the reset potential, the threshold, and the reversal potential:
\begin{equation} \label{eq:mid_activity}
    v_\text{reset} - E  = (v_\text{thr} - E) \exp(\tau_\text{ref}/\tau_m)
\end{equation}
This condition can be combined with the global or local optimization problem above to further constrain the other parameters.

\section{Distinction between task–state overlap and dimensionality}\label{sec:supp_toy}
Expanding dimensionality has long been known to improve performance on both classification and regression tasks \cite{maass2002, fusi2016, badre2021}, but performance improvements due to heterogeneity are not always accompanied by an expansion of dimensionality. To understand why, we must clarify what constitutes higher performance in temporal tasks.

Consider the following simple example: the activity of neurons in a network, $x_i(t)$, is linearly read out to reconstruct a periodic function $y(t)$ composed of two sinusoids (Fig.~\ref{fig:supp_toy}A). The most efficient way to represent such a task perfectly is to use two independent and specialized units, $x_1(t)$ and $x_2(t)$, that oscillate at exactly the same frequencies. If such units exist, then, with respect to $y(t)$, the activity of the remaining neurons is entirely irrelevant. Consequently, a two‑dimensional system that generates activity patterns $x_1(t)$ and $x_2(t)$ performs as well as a high‑dimensional system that also contains other independent activity patterns (Fig.~\ref{fig:supp_toy}B).  Conversely, a network’s activity may be high‑dimensional yet insufficient for this simple task if none of its neurons’ states overlap with the sinusoids in $y(t)$ (Fig.~\ref{fig:supp_toy}C). This counterexample shows that higher dimensionality is beneficial only when the extra dimensions span, overlap, or align with the temporal modes required by the tasks.

Simply put, an ideal network is one whose neuronal spectra have peaks that align well with those of the tasks.

\begin{figure}[!h]
	\centering
	\includegraphics[width=\linewidth]{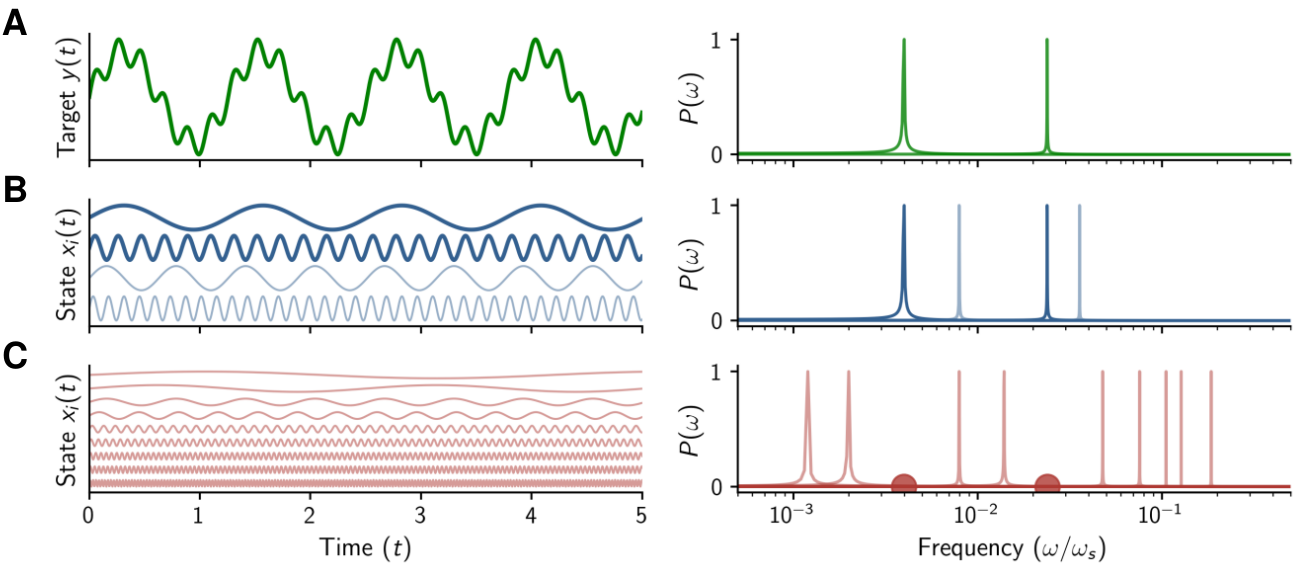}
	\caption{\textbf{High performance requires better state–task alignment}.  
    \textbf{(A)} Temporal (right) and frequency (left) representations of a simple target function composed of two sinusoids. This target is to be reconstructed linearly by two candidate networks.  
    \textbf{(B)} The first network consists of four independent neurons, two of which oscillate at frequencies that match those of the target function (bold lines). Consequently, a linear combination of the network’s state can perfectly reconstruct the target.  
    \textbf{(C)} The second network contains many more neurons but lacks units with the frequencies required by the target function (marked with red circles). Despite its higher dimensionality than the first network, it fails to reconstruct the simple task.}
	\label{fig:supp_toy}
\end{figure}

\clearpage

\section{Extended robustness figures}\label{sec:supp_robustness_figs}
\begin{figure}[!h]
	\centering
	\includegraphics[width=\linewidth]{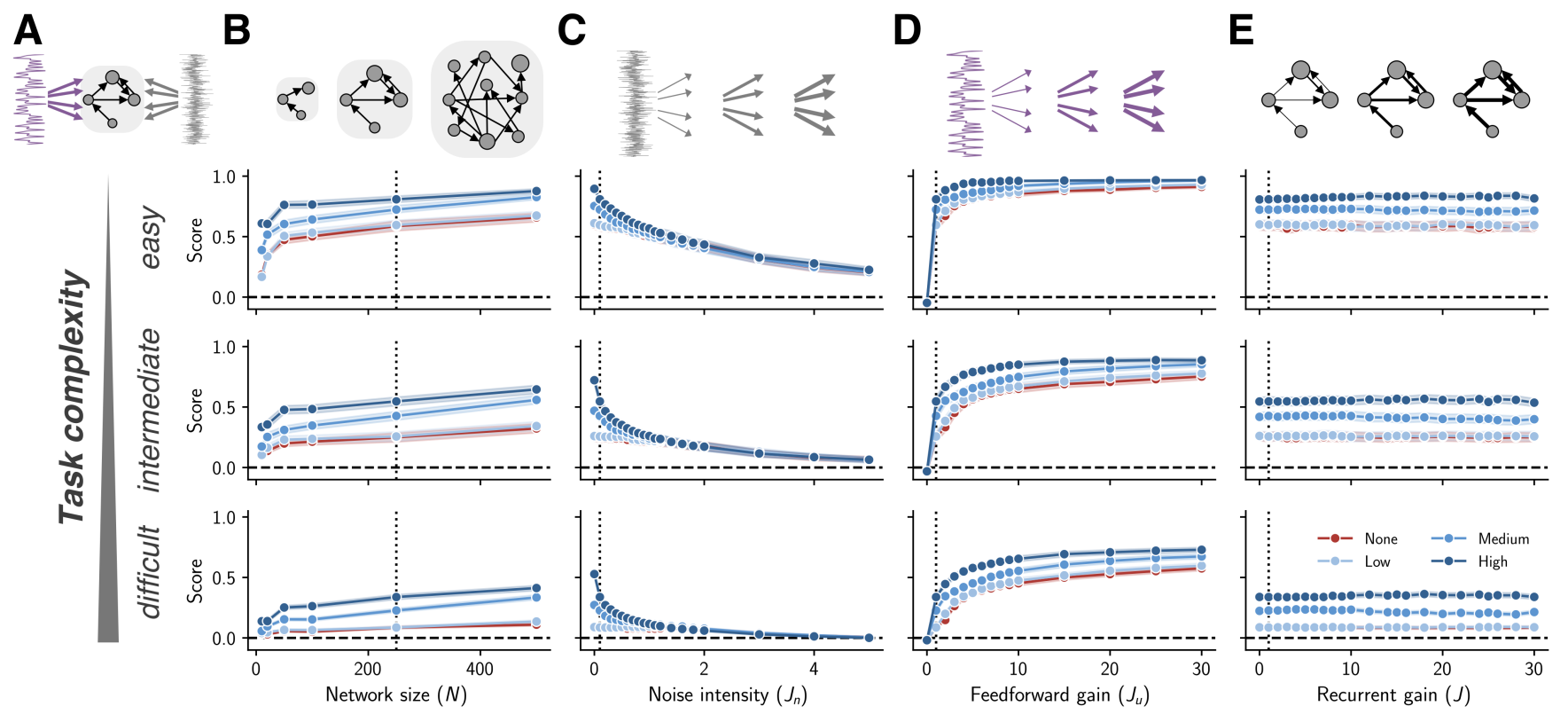}
    \caption{\textbf{Neuronal heterogeneity renders networks robust.} As in Fig.~\ref{fig:robustness_li}, but for networks of LIF neurons. \textbf{(A)} Schematic of the network and the deterministic and noise \textit{(top)} and the task complexity \textit{(bottom)}. Panels depict the influence of \textbf{(B)} network size, \textbf{(C)} noise intensity,  \textbf{(D)} feedforward gain, and \textbf{(E)} recurrent gain. The shaded area indicates the 95\% confidence interval across tasks within the corresponding complexity tier. Unlike rate‑based networks, the scores do not drop to the chance‑level value in panel (E) because LIF networks remain in a non‑chaotic regime, operating at firing rates far below their maximum. See Fig.~\ref{fig:supp_energy_bd_lif}G.}
	\label{fig:supp_robustness_lif}
\end{figure}

\clearpage

\begin{figure}[!h]
	\centering
	\includegraphics[width=\linewidth]{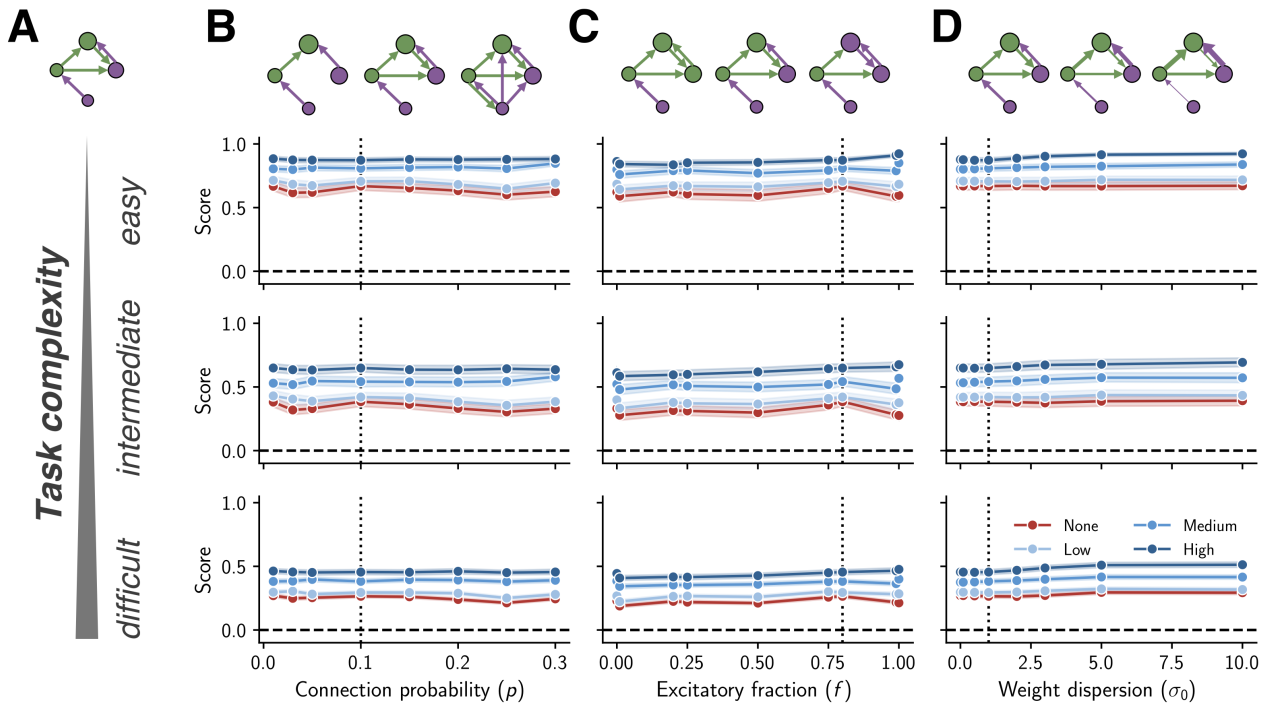}
    \caption{
        \textbf{Effect of population size and synaptic weights.}  
        \textbf{(A)} \textit{(Top)} Schematic of the recurrent connections in the networks. \textit{(Bottom)} Task‑complexity tiers.  
        \textbf{(B–D)} \textit{(Top)} Schematic of the varied hyperparameter: connection probability (B), fraction of excitatory neurons (C), and weight dispersion (D), which controls the degree to which Dale’s law is respected. \textit{(Bottom)} When excitatory–inhibitory balance is preserved, none of these changes affect the performance of rate‑based networks at any level of heterogeneity. The shaded areas in all panels indicate the 95\% confidence interval across tasks within the corresponding complexity tier.
    }
	\label{fig:supp_pfsig_li}
\end{figure}

\begin{figure}[!h]
	\centering
	\includegraphics[width=\linewidth]{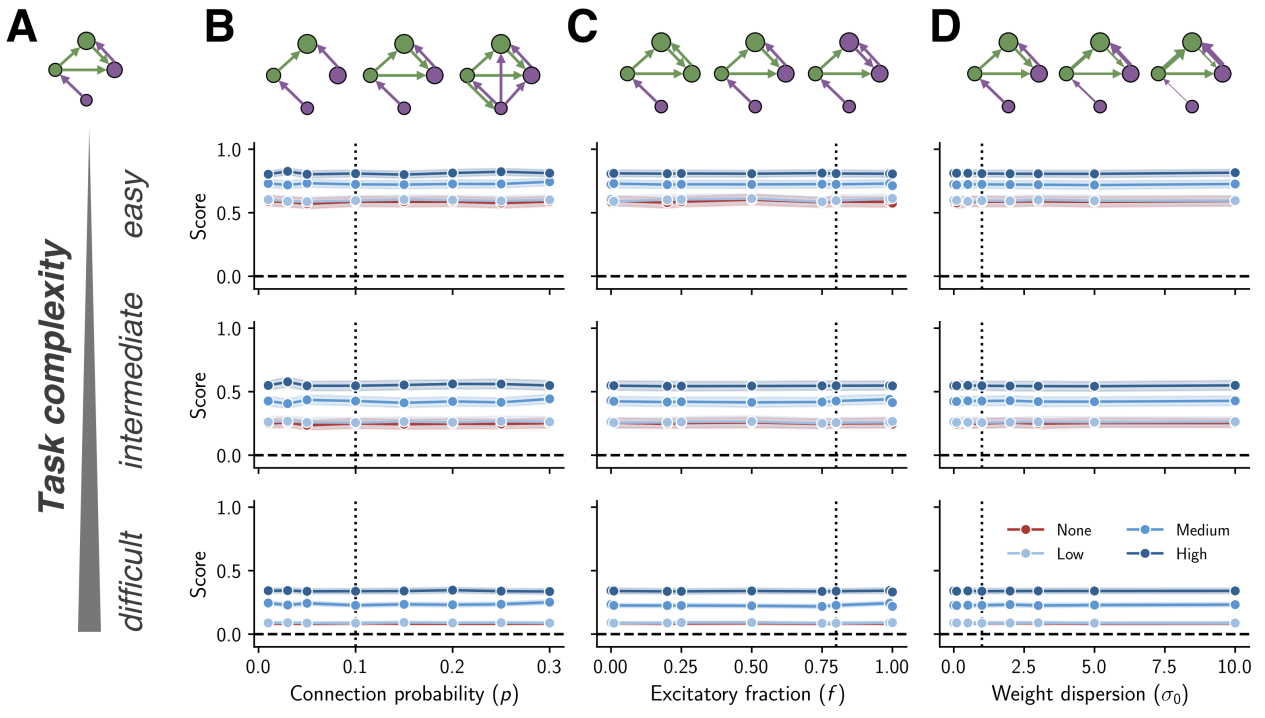}
    \caption{\textbf{Effect of population and weights}. Same as Fig.~\ref{fig:supp_pfsig_li}, but for spiking networks. \textbf{(A)} Schematic overview of the network and the task complexity. The impact on performance due to variation of \textbf{(B)} recurrent sparsity, \textbf{(C)} E-I ratio, and \textbf{(D)} recurrent weight dispersion.}
	\label{fig:supp_pfsig_lif}
\end{figure}

\clearpage

\section{Score distribution}\label{sec:supp_box_figs}

\begin{figure}[!ht]
  \centering
  \includegraphics[width=\linewidth]{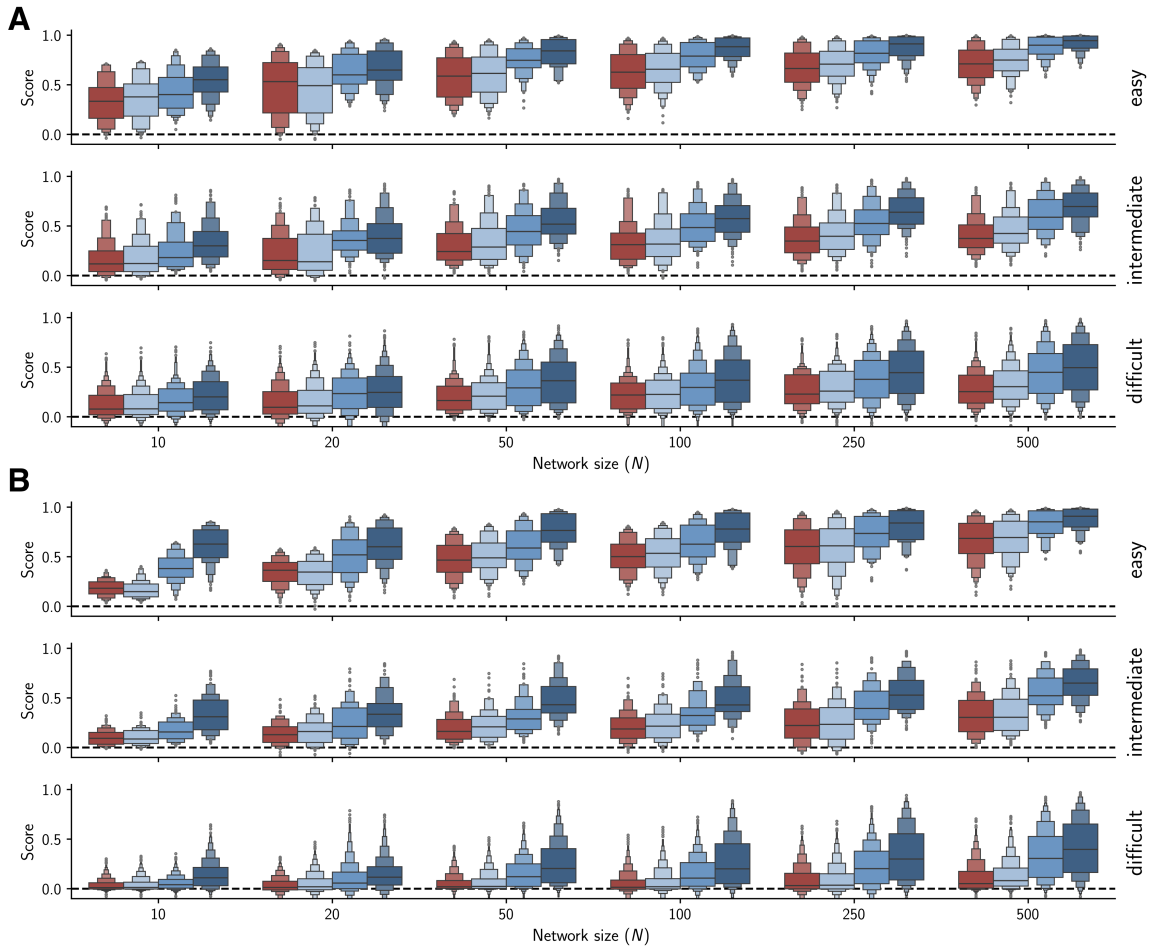}
  \caption{\textbf{Variation of score distribution with network size.} Performance distributions for \textbf{(A)} rate‑based and \textbf{(B)} spiking networks across \textbf{(rows)} tasks of different complexities. The mean of each distribution corresponds to the levels shown in Figures \ref{fig:robustness_li}B and \ref{fig:supp_robustness_lif}B, respectively. The color code is the same as in the main text.}
  \label{fig:supp_box_N}
\end{figure}

\clearpage

\begin{figure}[!ht]
  \centering
  \includegraphics[width=\linewidth]{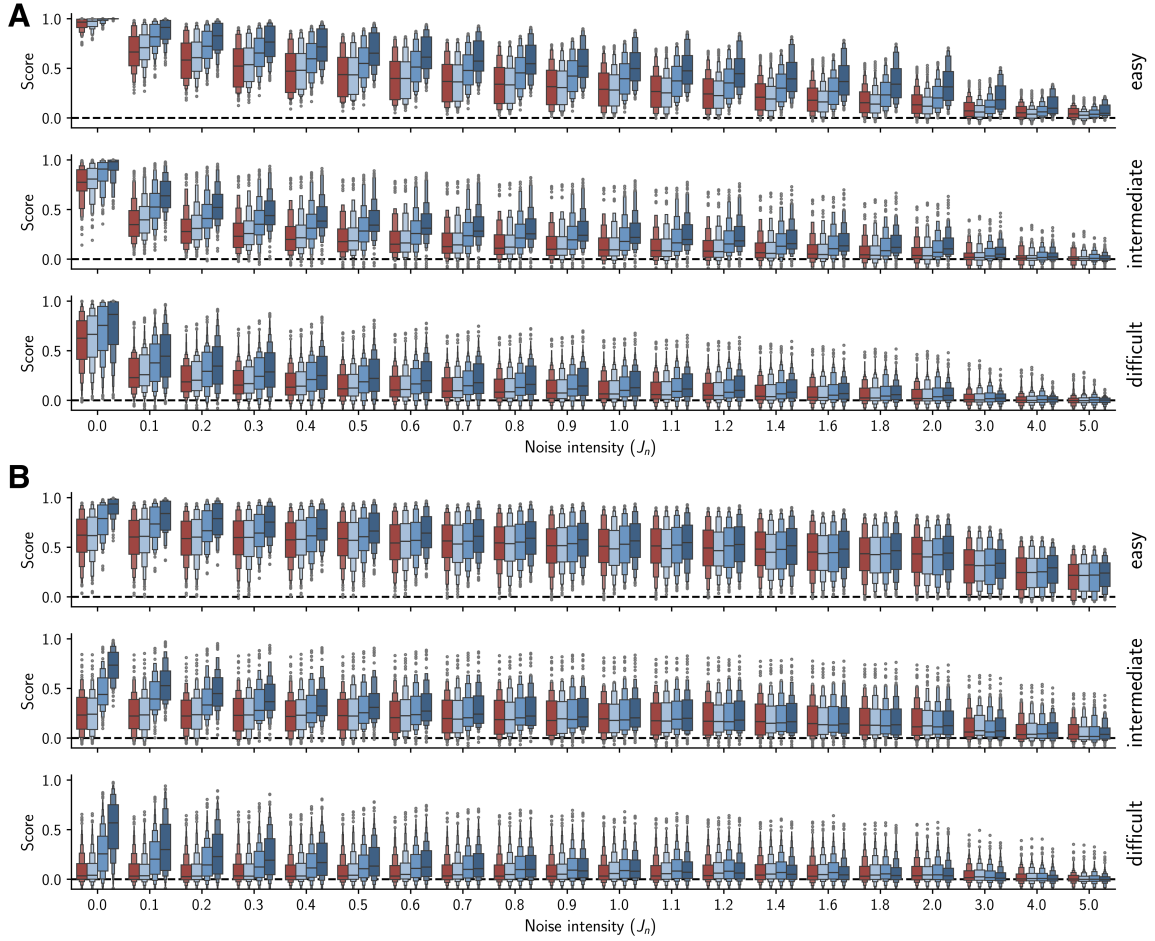}
  \caption{\textbf{Variation of score distribution versus noise intensity.} Performance distributions for \textbf{(A)} rate‑based networks and \textbf{(B)} spiking networks across \textbf{(rows)} tasks of different complexities. The mean of each distribution corresponds to the levels shown in Figures \ref{fig:robustness_li}C and \ref{fig:supp_robustness_lif}C, respectively. The color code is the same as in the main text.}
\end{figure}

\clearpage

\begin{figure}[!ht]
  \centering
  \includegraphics[width=\linewidth]{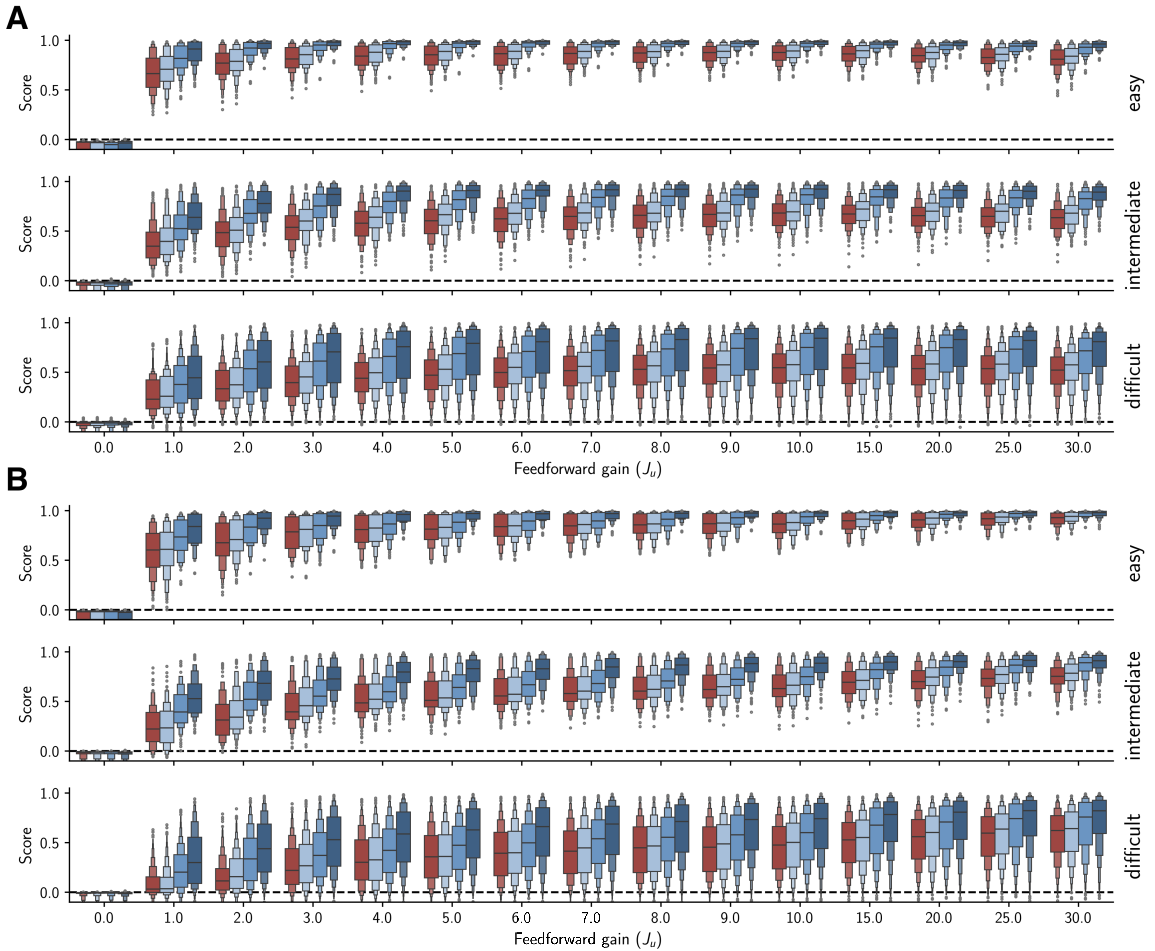}
  \caption{\textbf{Variation of score distribution versus input gain.} Performance distributions for \textbf{(A)} rate‑based networks and \textbf{(B)} spiking networks across \textbf{(rows)} tasks of different complexities. The average of each distribution corresponds to the levels shown in Figures \ref{fig:robustness_li}D and \ref{fig:supp_robustness_lif}D, respectively. The color code is the same as in the main text.}
\end{figure}

\clearpage

\begin{figure}[!ht]
  \centering
  \includegraphics[width=\linewidth]{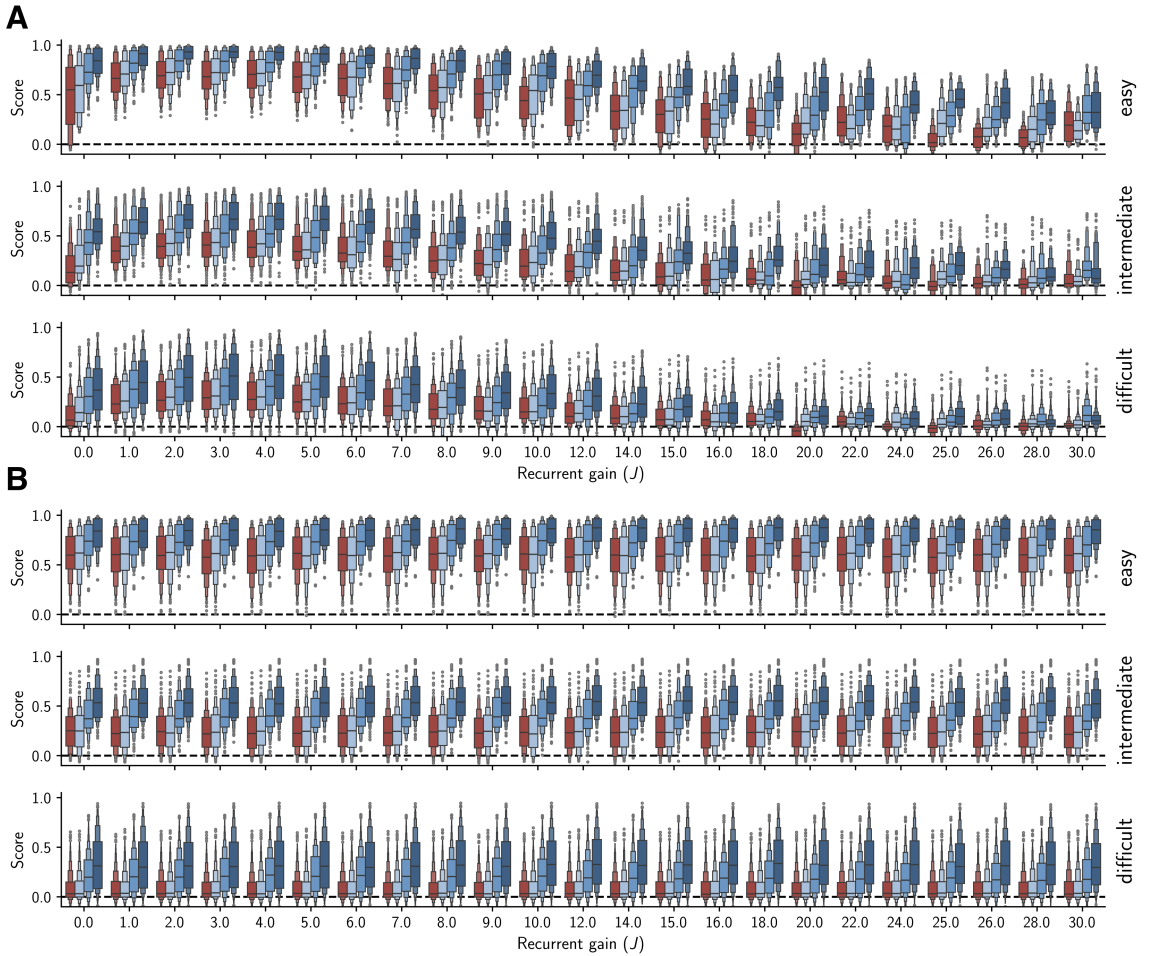}
  \caption{\textbf{Variation of score distribution versus recurrent gain.} Performance distributions for \textbf{(A)} rate‑based networks and \textbf{(B)} spiking networks across \textbf{(rows)} tasks of different complexities. The mean of each distribution corresponds to the levels shown in Figures \ref{fig:robustness_li}E and \ref{fig:supp_robustness_lif}E, respectively. The color code is the same as in the main text.}
\end{figure}

\clearpage

\begin{figure}[!ht]
  \centering
  \includegraphics[width=\linewidth]{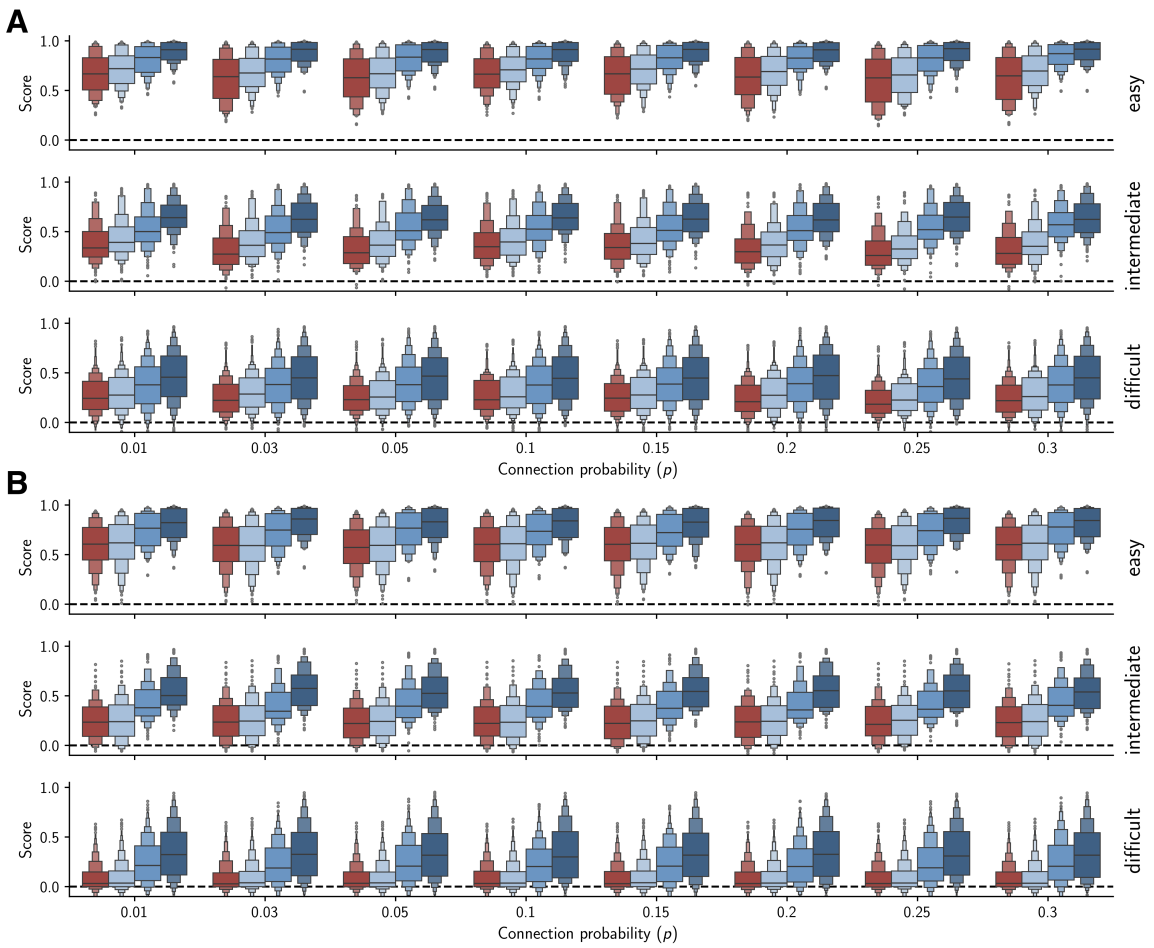}
  \caption{\textbf{Variation of score distribution versus connection probability.} Performance distributions for \textbf{(A)} rate‑based networks and \textbf{(B)} spiking networks across \textbf{(rows)} tasks of different complexities. The mean of each distribution corresponds to the levels shown in Figures \ref{fig:supp_pfsig_li}B and \ref{fig:supp_pfsig_lif}B, respectively. The color code is the same as in the main text.}
\end{figure}

\clearpage

\begin{figure}[!ht]
  \centering
  \includegraphics[width=\linewidth]{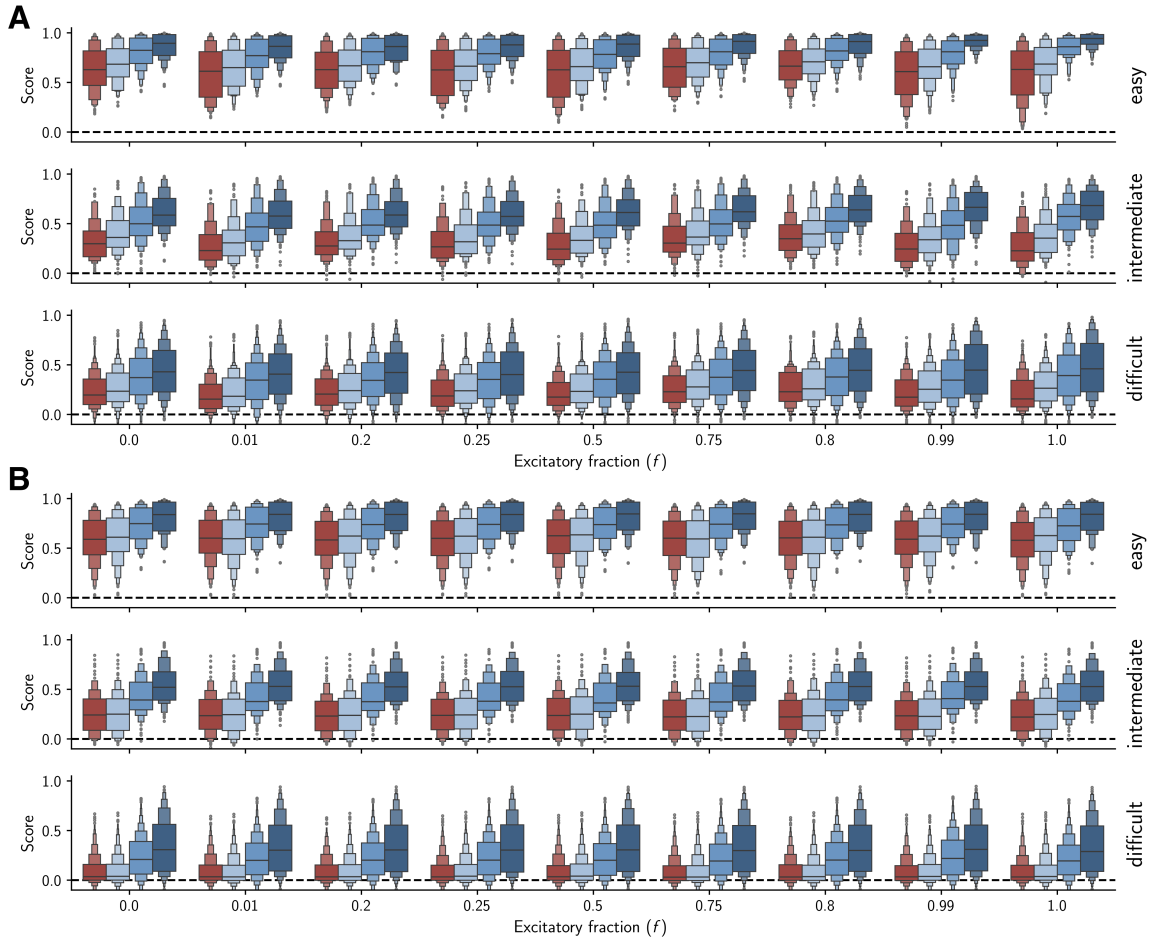}
  \caption{\textbf{Variation of score distribution versus excitation-inhibition ratio.} Performance distributions for \textbf{(A)} rate‑based networks and \textbf{(B)} spiking networks across \textbf{(rows)} tasks of different complexities. The mean of each distribution corresponds to the levels shown in Figures \ref{fig:supp_pfsig_li}C and \ref{fig:supp_pfsig_lif}C, respectively. The color code is the same as in the main text.}
\end{figure}

\clearpage

\begin{figure}[!ht]
  \centering
  \includegraphics[width=\linewidth]{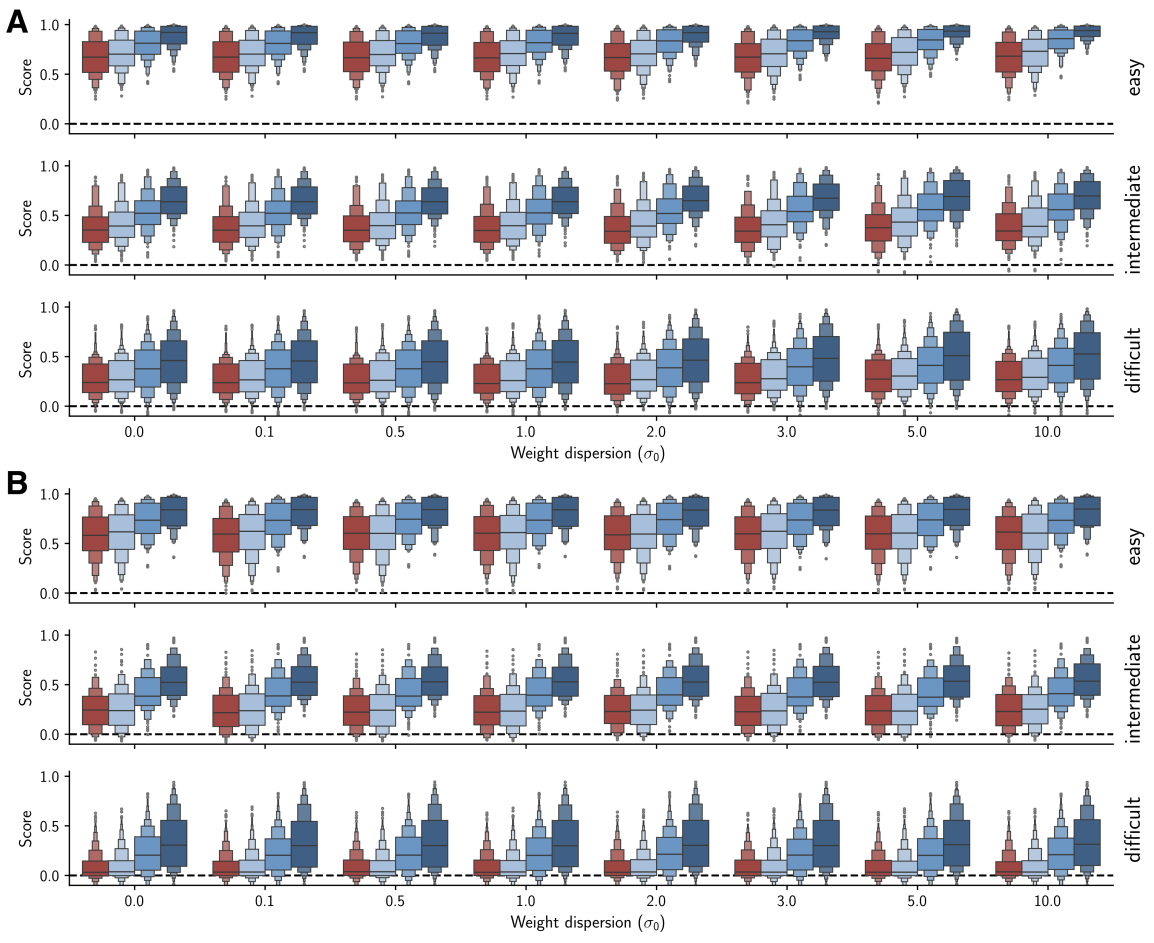}
  \caption{\textbf{Variation of score distribution versus recurrent weight dispersion.} Performance distributions for \textbf{(A)} rate‑based networks and \textbf{(B)} spiking networks across \textbf{(rows)} tasks of different complexities. The mean of each distribution corresponds to the levels shown in Figures \ref{fig:supp_pfsig_li}D and \ref{fig:supp_pfsig_lif}D, respectively. The color code is the same as in the main text.}
  \label{fig:supp_box_sigma}
\end{figure}

\clearpage

\section{Performance uncertainty quantification}\label{sec:supp_std_figs}

\begin{figure}[!ht]
	\centering
	\includegraphics[width=\linewidth]{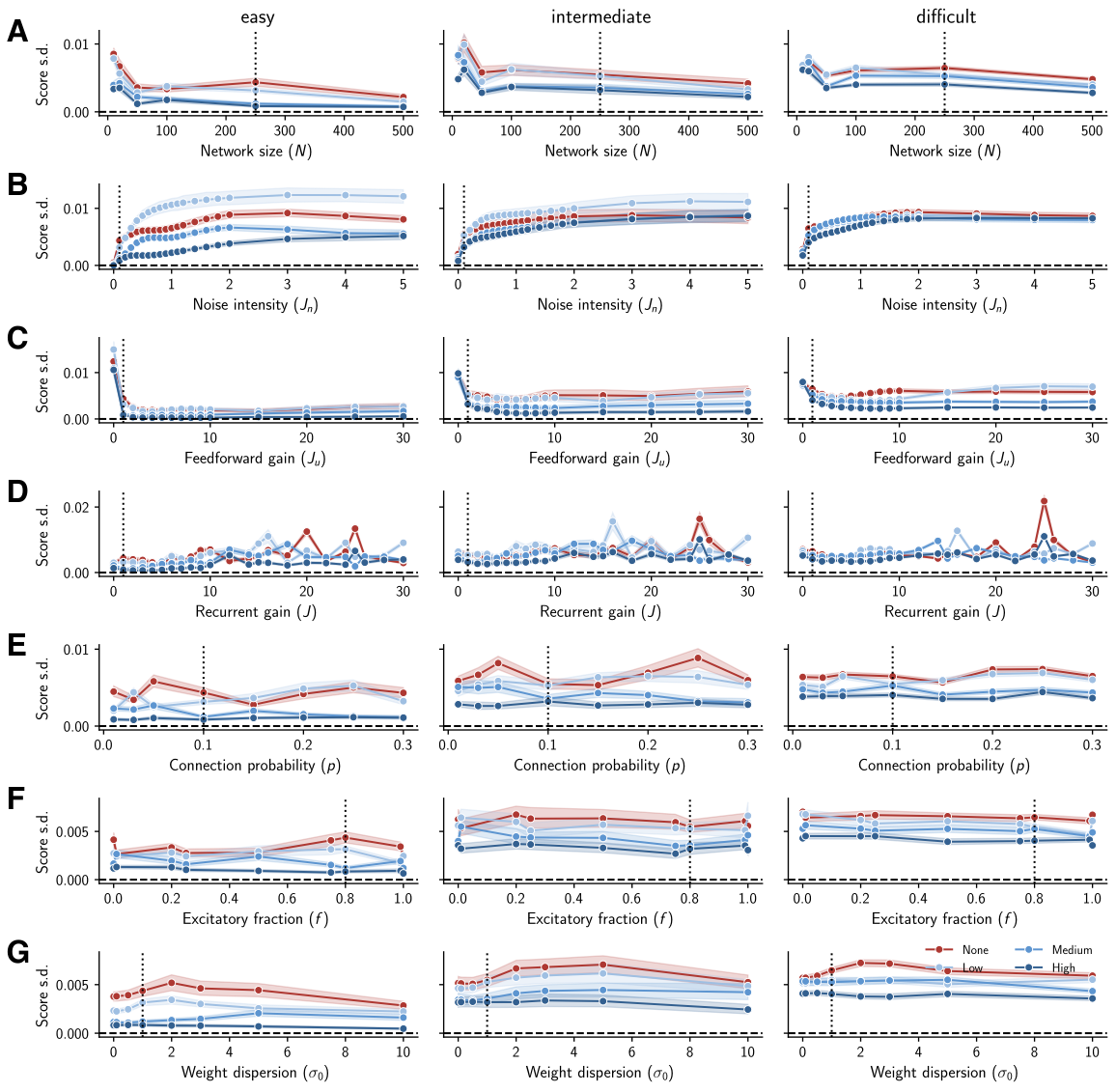}
	\caption{\textbf{Performance uncertainty in rate networks.} Performance standard deviation of rate networks, computed from three readouts obtained with fully independent samples. The data are broken down by task‑complexity tier (\textbf{columns}) and by the hyperparameter values shown: \textbf{(A)} network size, \textbf{(B)} noise intensity, \textbf{(C)} feedforward gain, \textbf{(D)} recurrent gain, \textbf{(E)} connection probability, \textbf{(F)} fraction of excitatory neurons, and \textbf{(G)} weight dispersion. The narrow range of standard deviations across all panels indicates that the error in the average performance is small, and three independent trials are sufficient. Note also the lower error observed in heterogeneous networks. The shaded areas indicates the 95\% confidence interval across tasks within the corresponding complexity tier.}
    \label{fig:supp_std_li}
\end{figure}

\begin{figure}[!ht]
	\centering
	\includegraphics[width=\linewidth]{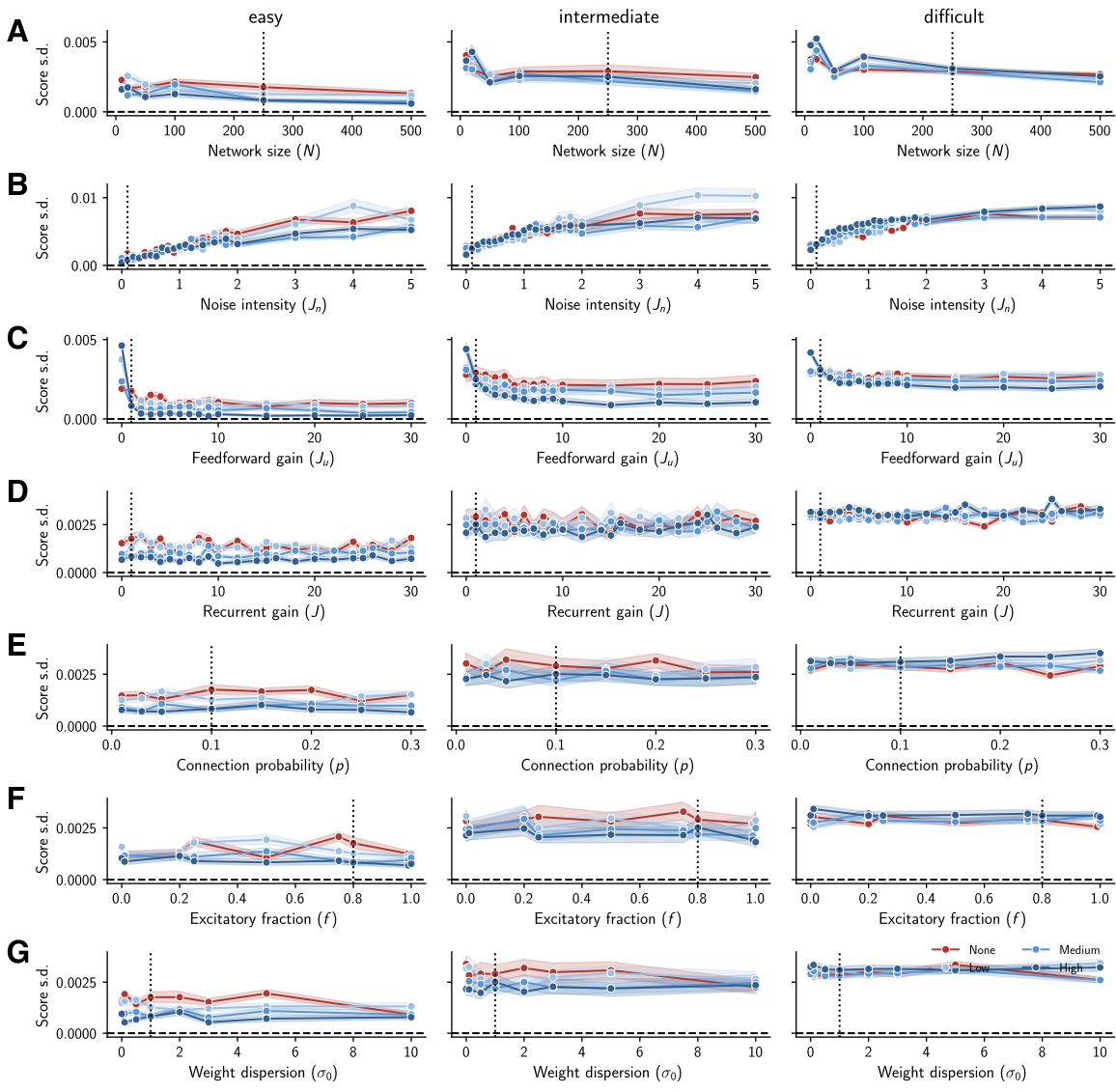}
	\caption{
    \textbf{Performance error in spiking networks.} As in Fig.~\ref{fig:supp_std_li}, but for spiking networks. The error bars are computed from three independent readouts. As before, heterogeneous networks exhibit lower variability. Panels indicate the influence on standard deviation as a function of \textbf{(A)} network size, \textbf{(B)} noise intensity, \textbf{(C)} feedforward gain, \textbf{(D)} recurrent gain, \textbf{(E)} recurrent sparsity, \textbf{(F)} E-I ratio, and \textbf{(G)} weight dispersion.
    }\label{fig:supp_std_lif}
\end{figure}

\begin{figure}[!ht]
	\centering
	\includegraphics[width=\linewidth]{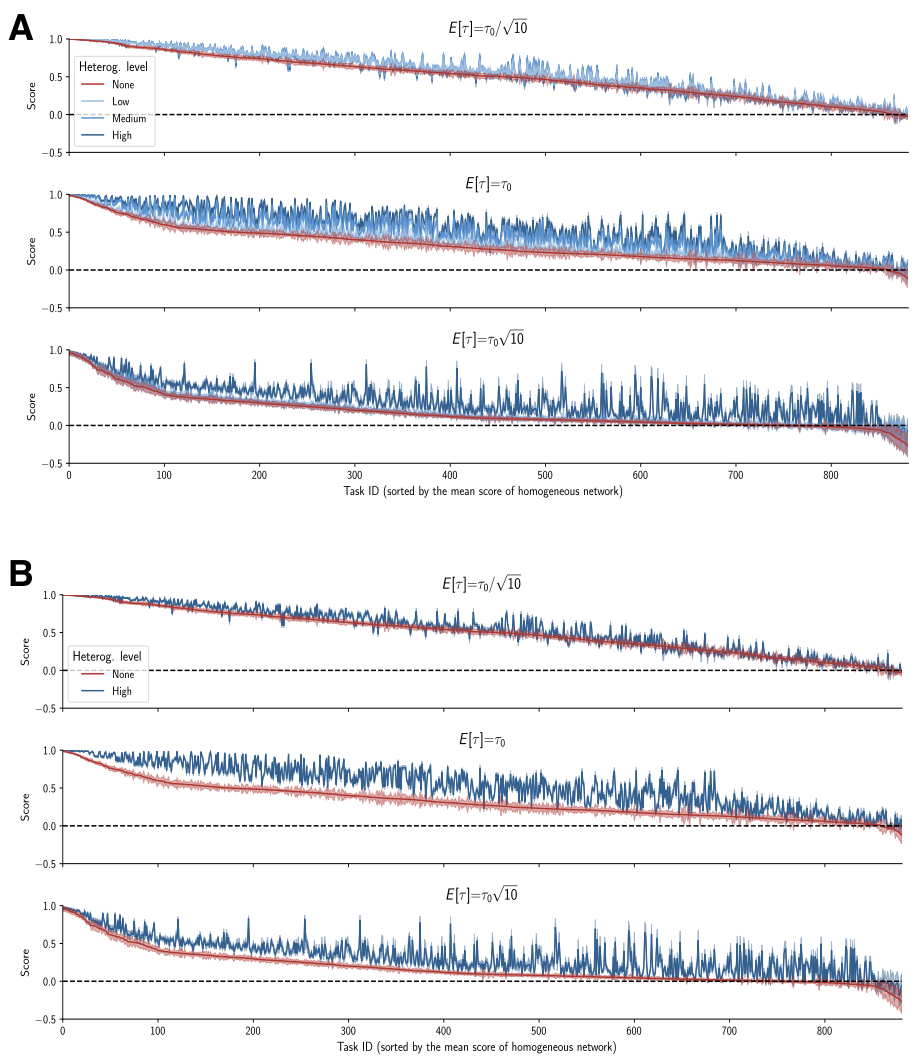}
	\caption{
    \textbf{Performance variation for all tasks in large LI network} \textbf{(A)} Tasks (x-axis) are sorted based on their mean score on the homogeneous network. The darker lines and the shaded area respectively indicate the average and standard deviation obtained over 10 different network realizations. \textit{(Top to bottom)} Different average time constants. 
    \textbf{(B)} Same as (A) but only for the homogeneous and highly heterogeneous network, shown for better visibility. Networks size is 250.
    }\label{fig:supp_UQ_LI250}
\end{figure}

\begin{figure}[!ht]
	\centering
	\includegraphics[width=\linewidth]{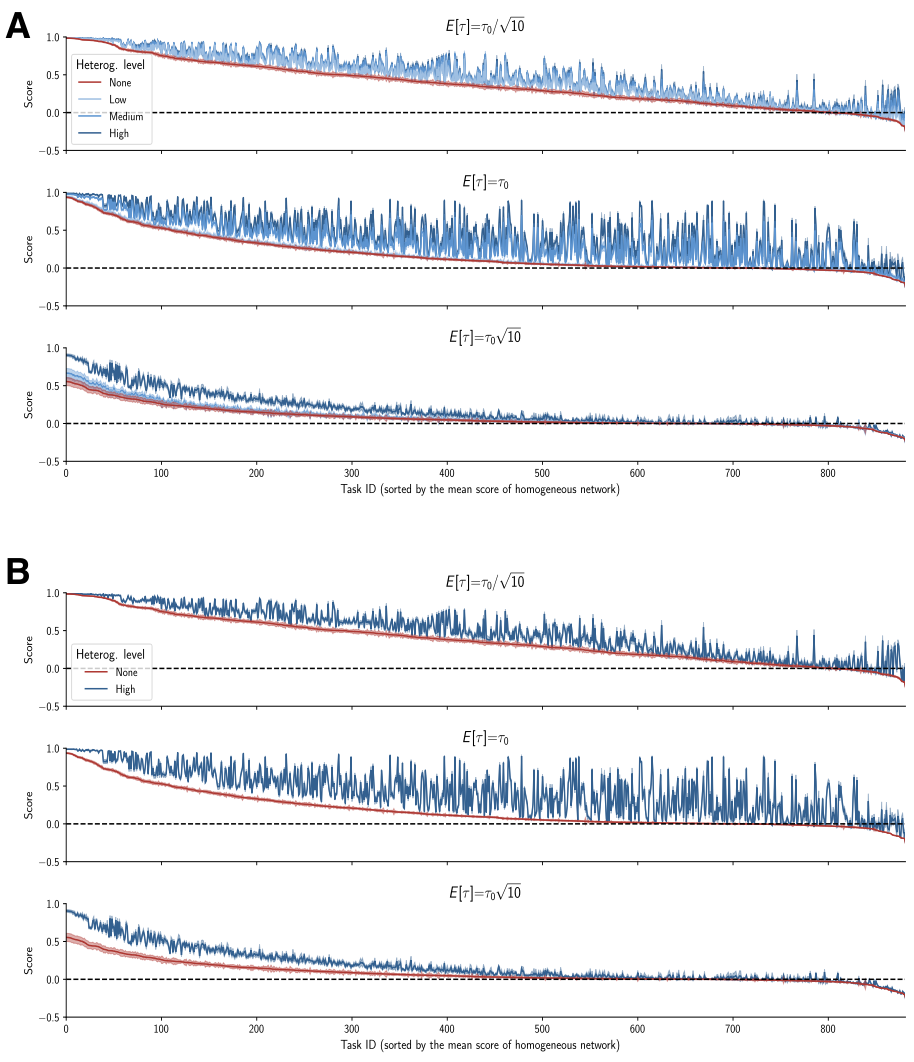}
	\caption{
    \textbf{Performance variation for all tasks in large LIF network} As Fig.~\ref{fig:supp_UQ_LI250} but for spiking neurons. \textbf{(A)} All heterogeneity levels; \textbf{(B)} Only minimal and maximal heterogeneity levels for better visibility.
    }\label{fig:supp_UQ_LIF250}
\end{figure}

\begin{figure}[!ht]
	\centering
	\includegraphics[width=\linewidth]{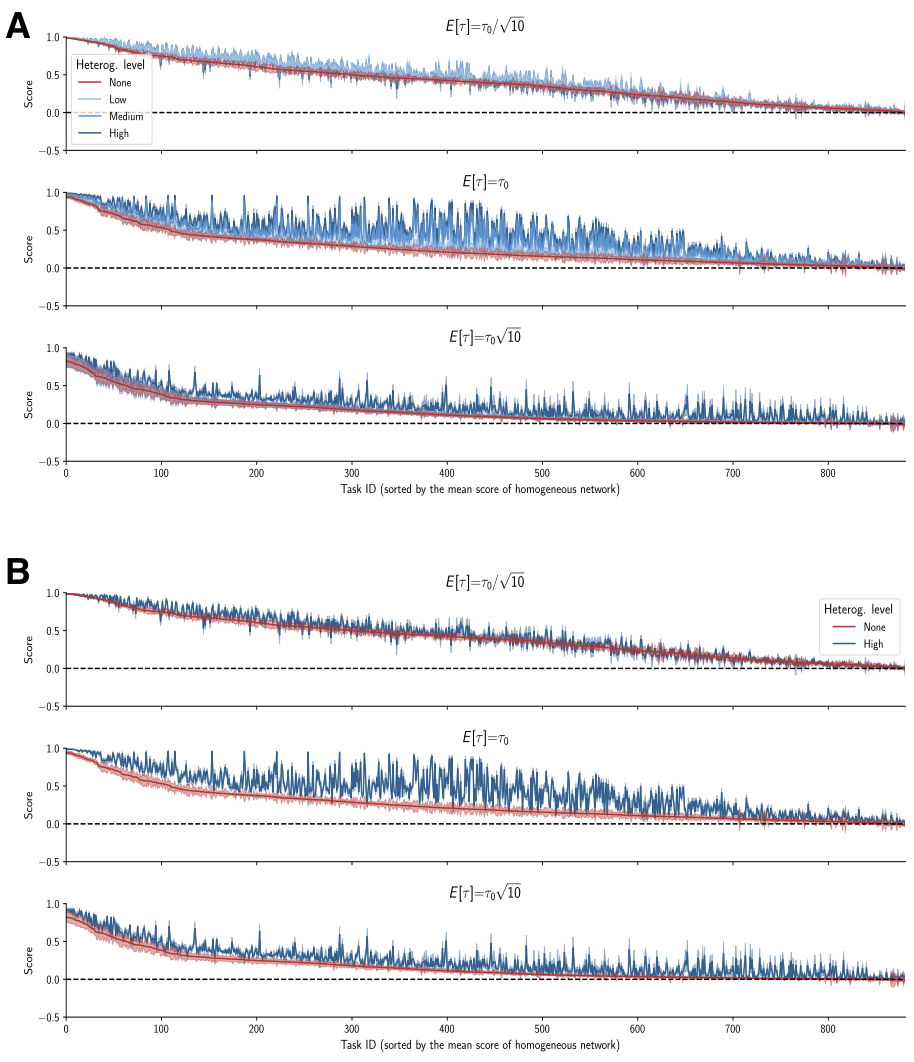}
	\caption{
    \textbf{Performance variation for all tasks in small LI network} As Fig.~\ref{fig:supp_UQ_LI250} but for network size of 50. \textbf{(A)} All heterogeneity levels; \textbf{(B)} Only minimal and maximal heterogeneity levels for better visibility.
    }
    
    \label{fig:supp_UQ_LI50}
\end{figure}

\begin{figure}[!ht]
	\centering
	\includegraphics[width=\linewidth]{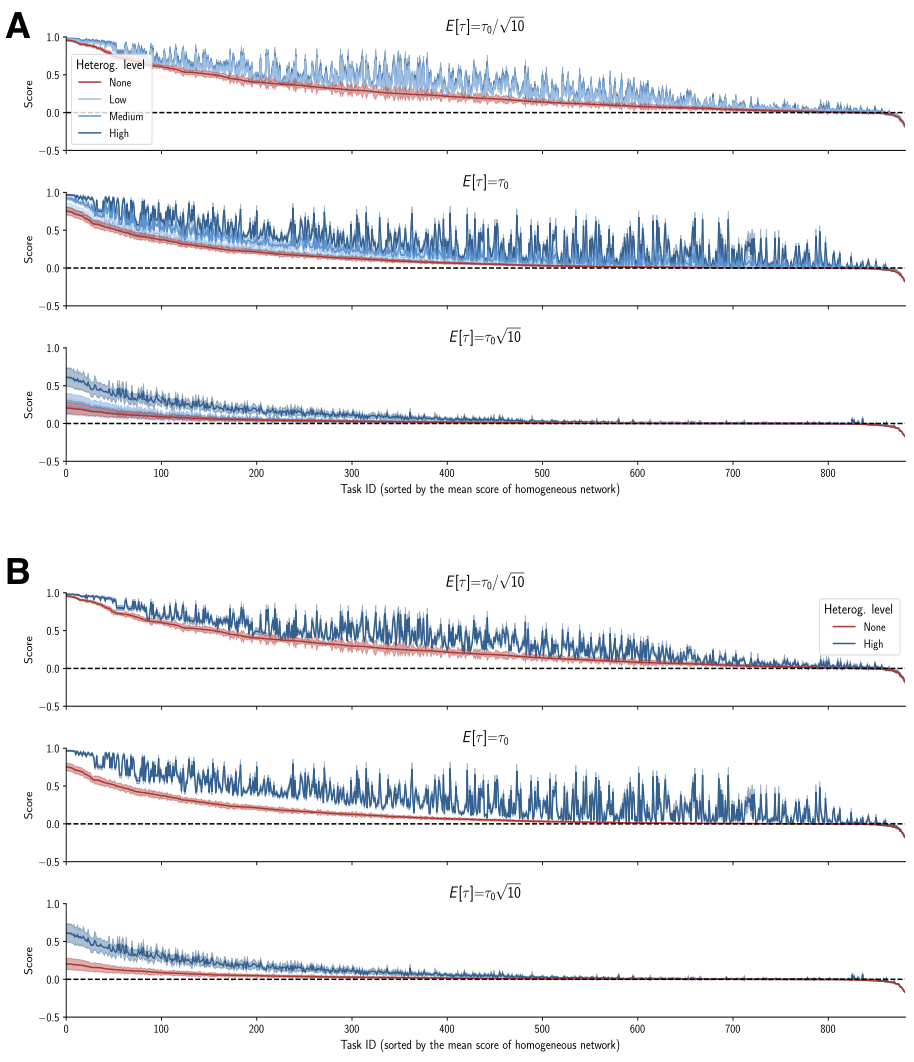}
	\caption{
    \textbf{Performance variation for all tasks in small LI network} As Fig.~\ref{fig:supp_UQ_LIF250} but for network size of 50. \textbf{(A)} All heterogeneity levels; \textbf{(B)} Only minimal and maximal heterogeneity levels for better visibility.
    }
    \label{fig:supp_UQ_LIF50}
\end{figure}

\clearpage

\section{Score and efficiency breakdown}

\begin{figure}[!ht]
	\centering
	\includegraphics[width=\linewidth]{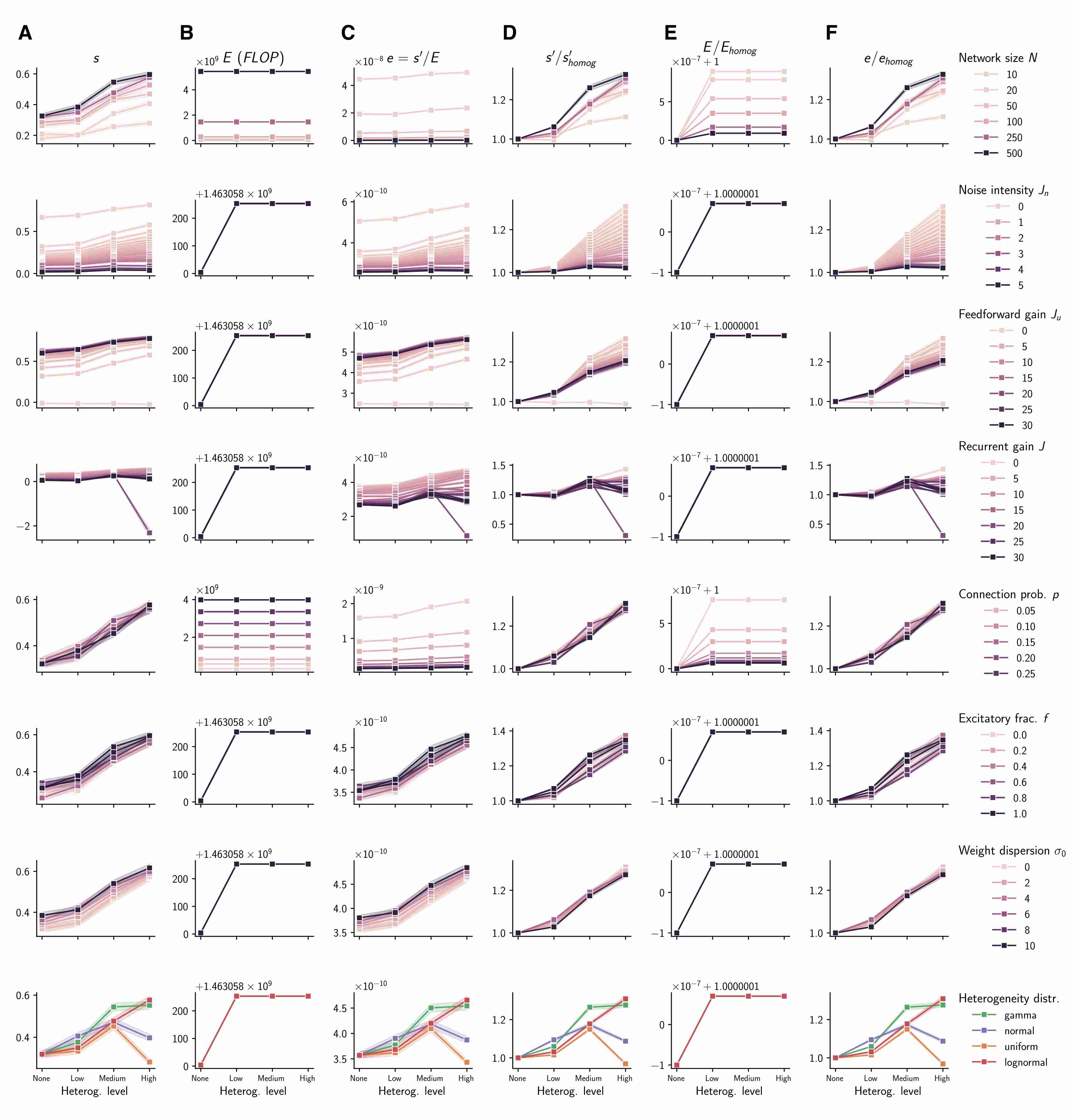}
	\caption{\textbf{Energy and performance variation of rate networks with heterogeneity.} Task‑averaged \textbf{(A)} performance $s$, \textbf{(B)} energy $E$, and \textbf{(C)} efficiency $e$ across different heterogeneity levels, together with \textbf{(D–F)} their ratios relative to the corresponding values in a homogeneous network. \textbf{(Rows)} each varied hyperparameter. Note that, because the score $s\in(-\infty,1]$, we first map it monotonically to the interval $(0,1)$ via $s'=\exp(s-1)$ before computing efficiency and the ratio metrics, thereby avoiding negative values or division by zero.}
    \label{fig:supp_energy_bd_li}
\end{figure}

\begin{figure}[!ht]
	\centering
	\includegraphics[width=\linewidth]{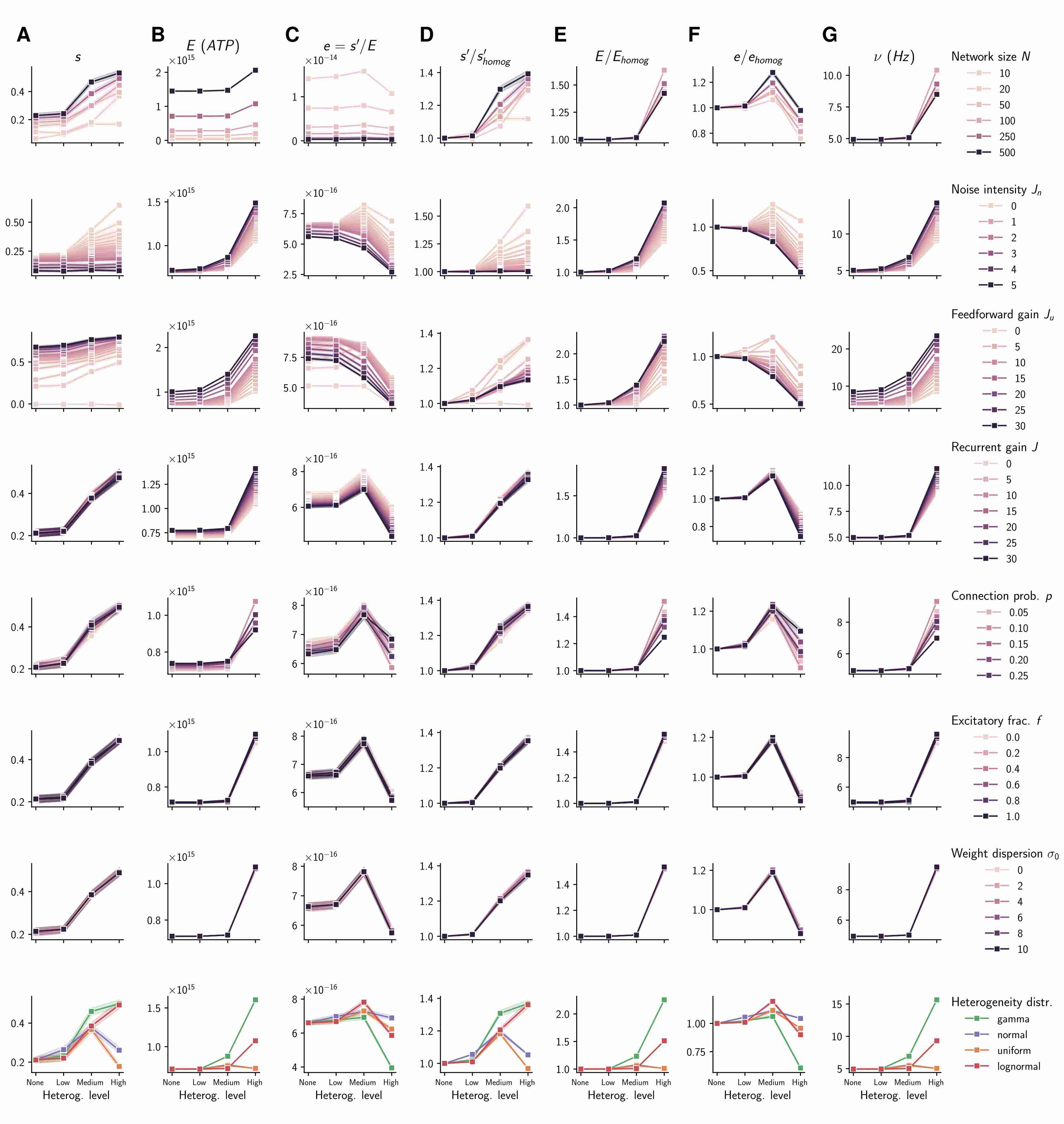}
    \caption{\textbf{Energy and performance variation of spiking networks with heterogeneity}. As in Fig.~\ref{fig:supp_energy_bd_li}, but for spiking networks. \textbf{(G)} The additional column shows the average firing rate $\nu$. Its correlation with $E$ indicates that the energy increase in heterogeneous networks is primarily due to higher neuronal activity. Note that in all scenarios, the networks fire with a rate far below the maximum possible firing rate of $\nu_{\max}=1/\tau_{\text{ref}}=500$ Hz. Consequently, even at the maximal recurrence level simulated ($J=30$), spiking networks stay effectively input‑driven, which also explains the lack of performance variation with $J$ in Fig.~\ref{fig:supp_robustness_lif}E.
    }
    \label{fig:supp_energy_bd_lif}
\end{figure}

\clearpage

\section{Details of the real-world benchmarks}\label{sec:prerprocess}

\subsection{Speech commands}\label{sec:sc-prerprocess}

The SC dataset introduced in \cite{warden2018} consists of thousands of utterances of 35 words, which appear with varying frequency in the dataset. Each sample is a 1‑second (or shorter) wave file sampled at 16 kHz, resulting in roughly 16000 data points. Because the utterances’ semantics are encoded in the relative acoustic features across different frequency bands rather than in the raw waveform, it is common to transform the input into the time‑frequency domain. Common approaches include the short‑time Fourier transform, Mel transform, wavelet transform, or similar. These transomed inputs are then used as inputs to the RNN.

We downloaded the dataset using the \texttt{torchaudio} package, together with an RNN bundle (also provided in \texttt{torchaudio}) to conveniently transform the audio files into the time‑frequency domain. Bundles are pipeline stacks consisting of three blocks: feature extractor, inference, and post‑processing\footnote{Please see the pipelines documentation here \url{https://docs.pytorch.org/audio/main/pipelines.html}}. The feature extractor block consists of a set of fixed transformations (e.g., pooling, padding, transposition, logarithm, etc.) and is devoid of learnable parameters. The other two blocks, however, contain pre‑trained weights. We fetched \texttt{EMFORMER\_RNNT\_BASE\_LIBRISPEECH}, a bundle wrapping an EMformer‑based architecture \cite{shi2020} trained on the \texttt{LibriSpeech} dataset \cite{panayotov2015}, and used its feature extractor; the remaining blocks of the bundle were not used. Direct inspection of the feature extractor reveals that it is composed of the following transformation layers:
\begin{enumerate}
    \item \texttt{MelSpectrogram}
    \item \texttt{Transpose}
    \item \texttt{Log}
    \item \texttt{GlobalStatsNormalization}
    \item \texttt{Pad}
\end{enumerate}
which convert the audio files into log‑Mel filter banks. For a typical 1‑second input, the resulting features are of size $80 \times 105$, where 80 is the number of Mel components (frequency bands) and 105 is the number of timesteps; the columns may be fewer for shorter trials. These features follow the same trial‑normalization and weight scaling described in Sec.~\ref{sec:method_demo}. A “fixation” channel (always zero except at the beginning of each trial) is prepended as the first channel. A few exemplary trial streams are shown in Fig.~\ref{fig:speech_uvx}.

To minimize the computational load, we only classified the ten most frequent labels. The associated samples were collected by fetching 300 batches of size 32 and retaining only those samples that belong to the top‑10 most frequent labels.

\subsection{Permuted sequential MNIST}\label{sec:psmnist-prerprocess}
This task was first introduced in \cite{goodfellow2015} and highlights the ability of a network to integrate and combine long‑term dependencies in the input. In psMNIST, all 784 pixels of the images in the MNIST dataset \cite{lecun1998} are first shuffled using a random (but fixed) permutation and then flattened into a 1D vector. The pixels within each vector are then sequentially fed to the RNN. The goal is to predict the label of the image after all pixels have been observed.

The MNIST dataset was obtained from the \texttt{torchvision} library. Samples were shuffled, first at the image level, and then (using a fixed permutation) at the pixel level, reshaped, and transformed to \texttt{numpy} arrays before being fed to the RNN. All 784 pixels were first mapped to the (0, 1) range. Before concatenation, each pixel train was prepended with a pixel of value –1, acting as a fixation cue , used to reset the network state at the beginning of each trial. To reduce the computational load, as before, we fetched only 3200 training and 3200 test samples from the MNIST dataset, in 100 batches of size 32.

\subsection{Contextual decision making}\label{sec:cdm-prerprocess}

The task setup is similar to that introduced in \cite{mante2013}. An agent receives a contextual cue together with two spatial stimulus modalities (e.g., color and motion). The coherence level of each modality is randomly selected from 5\%, 15\%, or 50\%. The agent must choose the option associated with the maximal coherence of the modality signaled by the contextual cue. A white noise with intensity $\sigma=1$ is added to each stimulus channel. Stimuli are presented for a fixed (or stochastic) duration, followed by a fixed (or stochastic) delay period. Only after the delay and the onset of a go cue may the agent make its decision.

We used the \texttt{neurogym} library to set up the task environment. However, due to the presence of noise in the stimuli, especially for inputs with low coherence values or short presentation times, integrating the empirically observed stimuli may yield a different correct choice than the intended target. Therefore, we decided to correct the labels by prioritizing for the empirical observation rather than the pre‑specified ground truths. In addition, to unify our setup with the other two benchmarks, we removed the explicitly presented go‑cue channel. Furthermore, since the contextual input is binary (either color or motion), one context channel is sufficient; thus, we dropped one of the context channels generated by the \texttt{neurogym} package. The simulation timestep is set to $\Delta t = 1$. For both train and test sets, sequences of length 200k timesteps were generated, and partial trials were discarded. On average, this sequence length results in roughly 1200 trials in each set.

The original setup in \cite{mante2013} presents agents with only two choices (right and left), conforming to the two‑armed bandit setup commonly used in psychological and cognitive experiments. However, we found this task too easy, as both homogeneous and heterogeneous networks could solve it with nearly perfect performance (cf. Fig.~\ref{fig:easy_mante}). Thus, we increased the complexity by introducing six possible choices (\texttt{dim\_ring=6} in the package) rather than two. This setup can be understood as six choices arranged on a ring, 60 degrees apart, and with six different colors. The stimulus modalities are applied according to a circular coherence profile, maximizing at one of the possible choices.

We considered two conditions: a stochastic stimulus period or a stochastic delay period. For this, we adjusted the stimulus \texttt{timing} argument to generate random period lengths for one interval and fixed lengths for the other. The random interval lengths are drawn from a truncated exponential distribution with a mean of 50 timesteps and a minimum and maximum of 10 and 250 timesteps, respectively. The fixed interval always has a length of 100. Exemplary trial streams and their effects on voltage and neural activations are shown in Figs.~\ref{fig:mante_delay_stoch} and \ref{fig:mante_stim_stoch}. 

\clearpage

\section{Exemplary trials}

\begin{figure}[!ht]
	\centering
	\includegraphics[width=\linewidth]{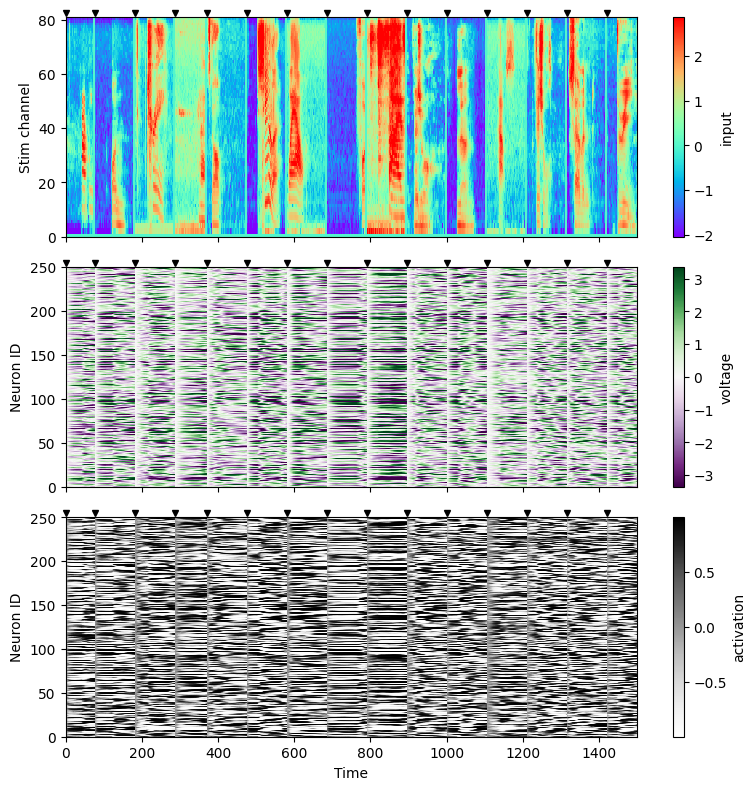}
    \caption{
    \textbf{The speech command setup}. \textbf{(Top)} Input stream to the agent. Except for the first channel (starting from the bottom), which encodes the “fixation” cue, the other 80 channels contain the log‑Mel content of the audio files and can be regarded as frequency bands. The triangles at the top indicate the start of a trial. \textbf{(Middle)} Time‑aligned neuronal voltages; note the resetting effect at the start of each trial. \textbf{(Bottom)} Activation map of neurons over time.
    }
    \label{fig:speech_uvx}
\end{figure}

\clearpage

\begin{figure}[!ht]
	\centering
	\includegraphics[width=\linewidth]{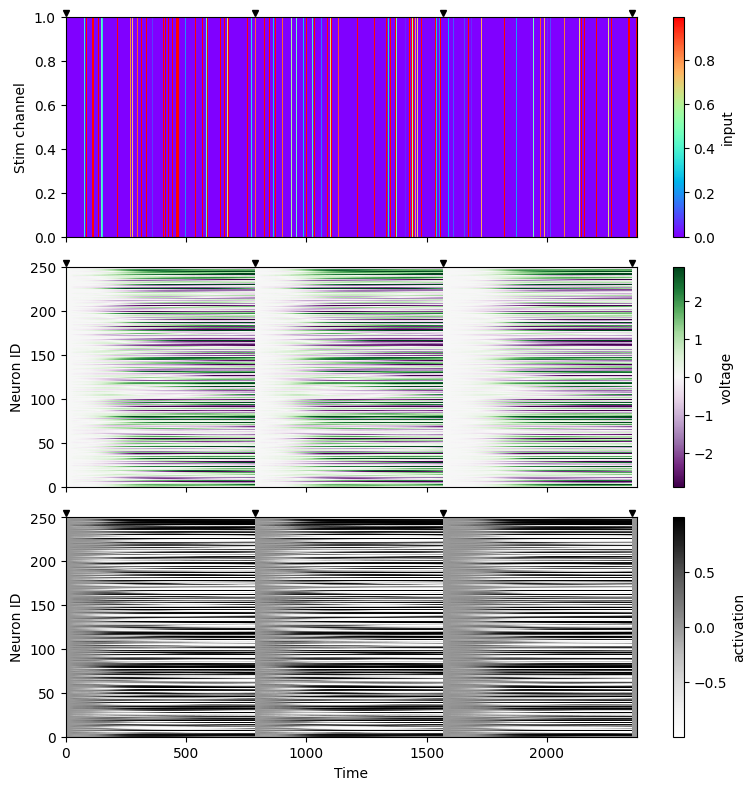}
    \caption{
    \textbf{The permuted sequential MNIST setup}. Same as Fig.~\ref{fig:speech_uvx} but for the psMNIST task with only one input channel. Note that here, as opposed to the other tasks, the fixation signal is provided as within the sole signal channel (as a negative value).
    Panels indicate the temporal variation of \textbf{(top)} the stimulus, \textbf{(middle)} neurons' voltage, and \textbf{(bottom)} their activation.
    }
    \label{fig:mnist_uvx}
\end{figure}

\clearpage

\begin{figure}[!ht]
	\centering
	\includegraphics[width=\linewidth]{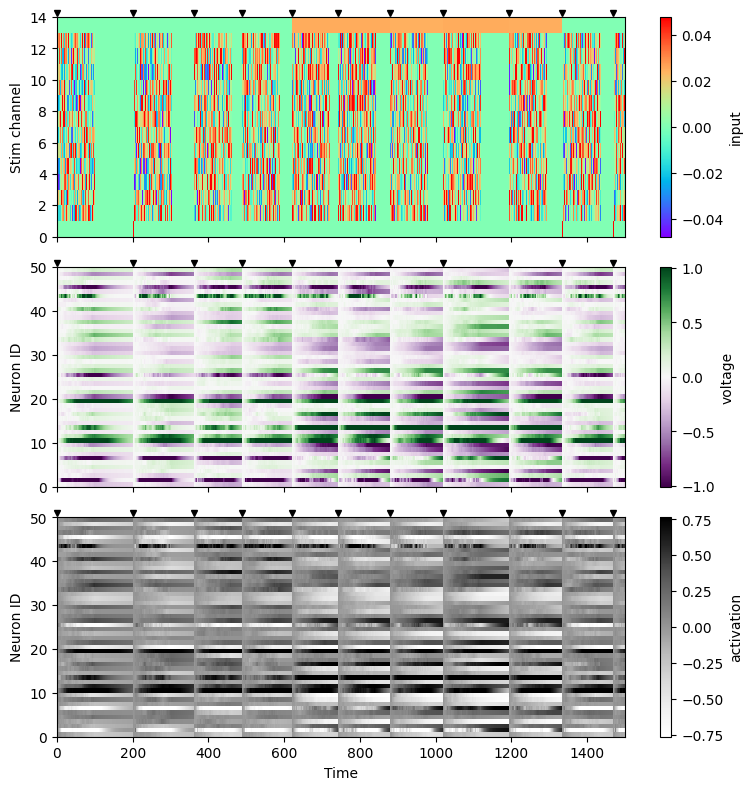}
    \caption{
    \textbf{The CDM setup with stochastic delay period}. Same as Fig.~\ref{fig:speech_uvx} but for CDM task. The first channel of the input stream (starting from the bottom) encodes the "fixation" cue as in SC task. The next 6 channels indicate the observed stimulus of the first modality (e.g., color) at distinct angular location (60 degrees apart), followed by analogous signal for modality 2 (motion) in the subsequent 6 channels. The last channel indicate the context (motion or color). Panels indicate the temporal variation of \textbf{(top)} the stimulus, \textbf{(middle)} neurons' voltage, and \textbf{(bottom)} their activation.
    }
    \label{fig:mante_delay_stoch}
\end{figure}

\clearpage

\begin{figure}[!ht]
	\centering
	\includegraphics[width=\linewidth]{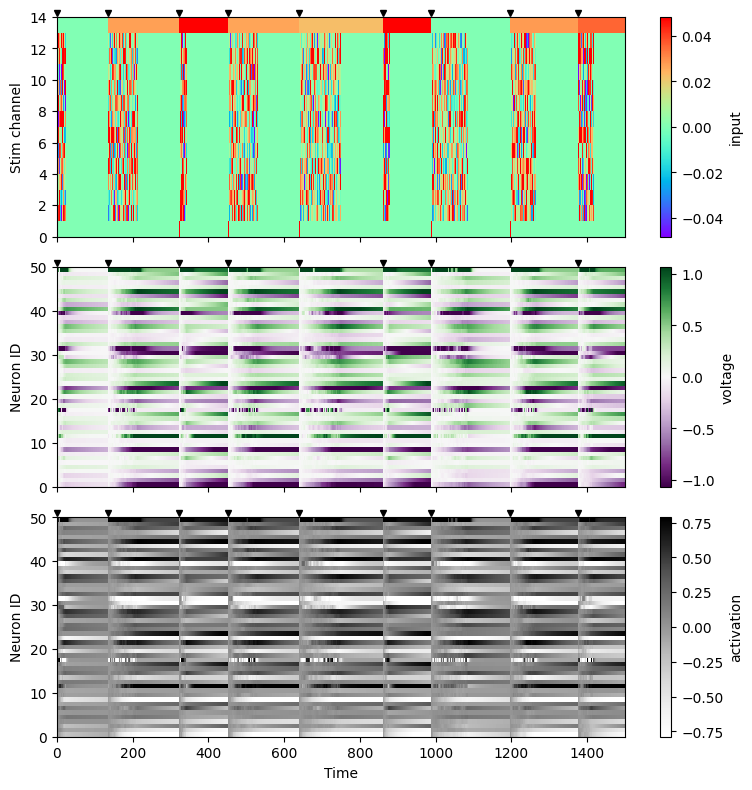}
    \caption{
    \textbf{The CDM setup with stochastic stimulus period}. Same as Fig.~\ref{fig:speech_uvx} but for fixed delay period and stochastic stimulus period. Panels indicate the temporal variation of \textbf{(top)} the stimulus, \textbf{(middle)} neurons' voltage, and \textbf{(bottom)} their activation.}
    \label{fig:mante_stim_stoch}
\end{figure}

\clearpage

\section{Dependence of the performance on temporal feature pooling and regularization}\label{sec:pool-creg}

\begin{figure}[!ht]
	\centering
	\includegraphics[width=\linewidth]{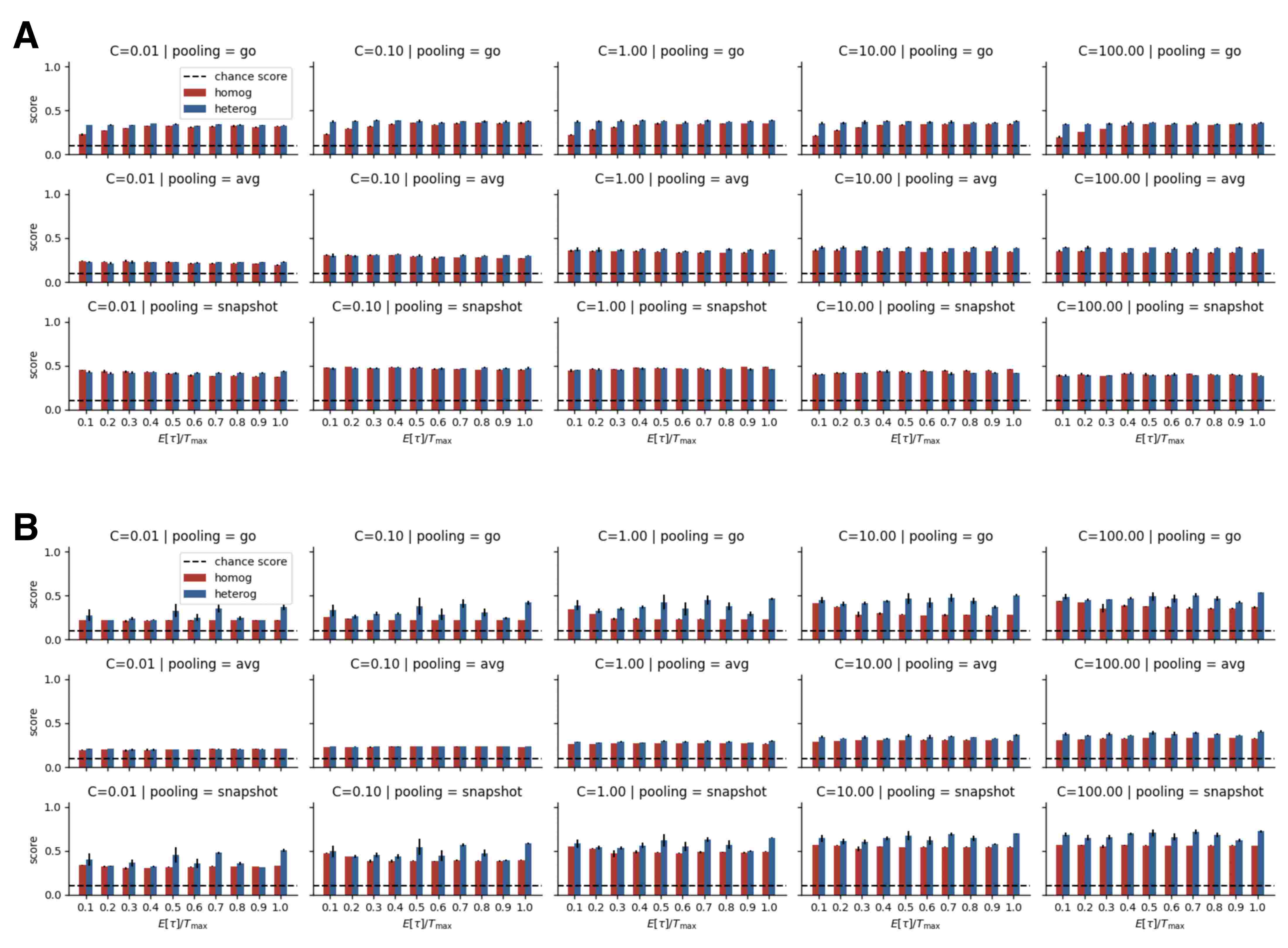}
    \caption{
    \textbf{(A)} Speech command task; \textbf{(B)} Permuted sequential MNIST task. A higher $C$ value leads to weaker regularization. For the SC and psMNIST tasks, the top row ("Go" pooling) with regularization values of, respectively, $C=10.0$ and $C=100$, are shown in the main text. Network setup and the number of realizations are identical to the ones described in the main text.
    }
    \label{fig:creg_pool-sc-mnist}
\end{figure}

\begin{figure}[!ht]
	\centering
	\includegraphics[width=\linewidth]{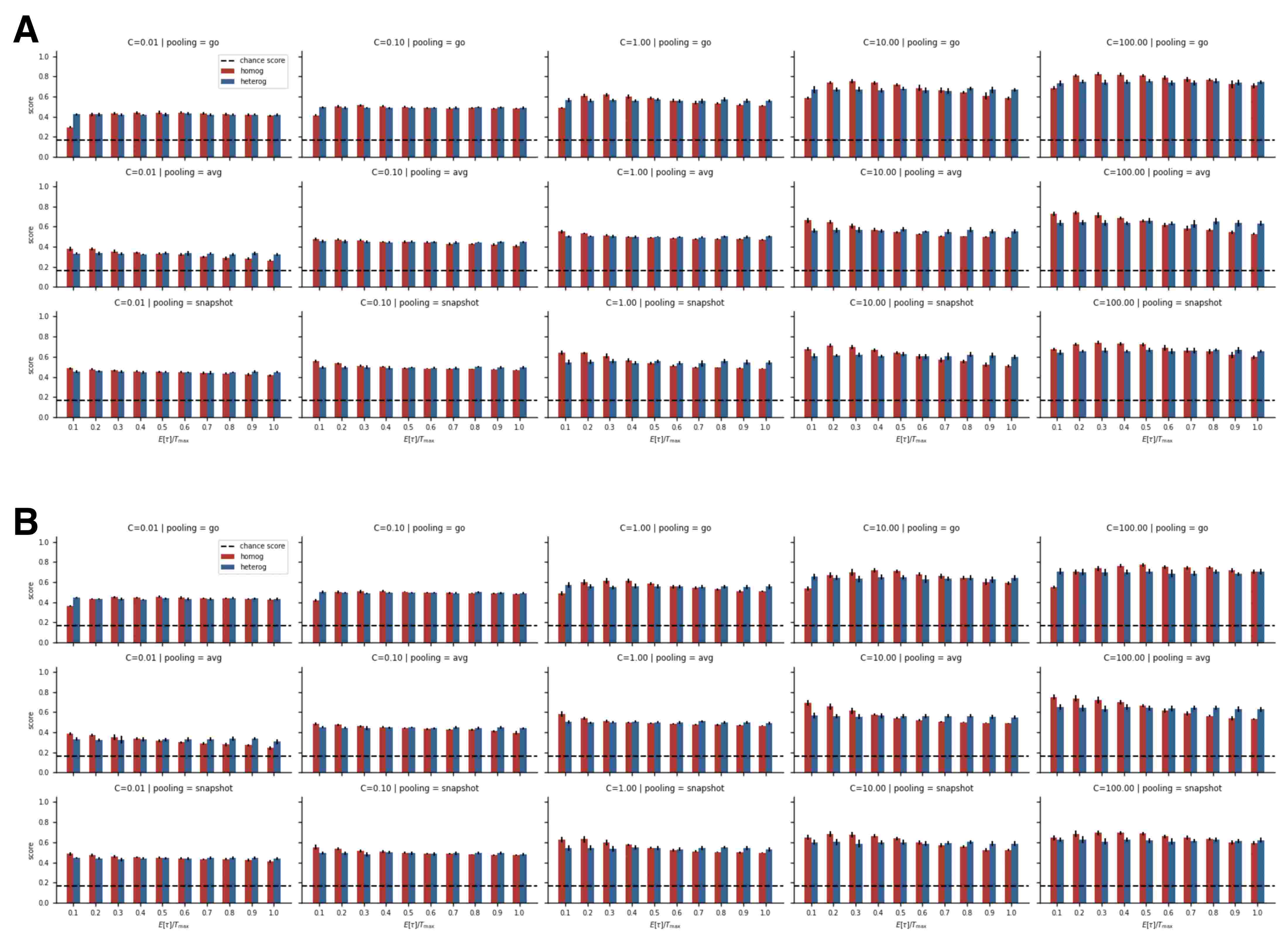}
    \caption{
    \textbf{(A)} CDM task with stochastic stimulation period and fixed delay; \textbf{(B)} CDM task with stochastic delay period and fixed stimulation time.
    }
    \label{fig:creg_pool-mante}
\end{figure}

\begin{figure}[!ht]
	\centering
	\includegraphics[width=\linewidth]{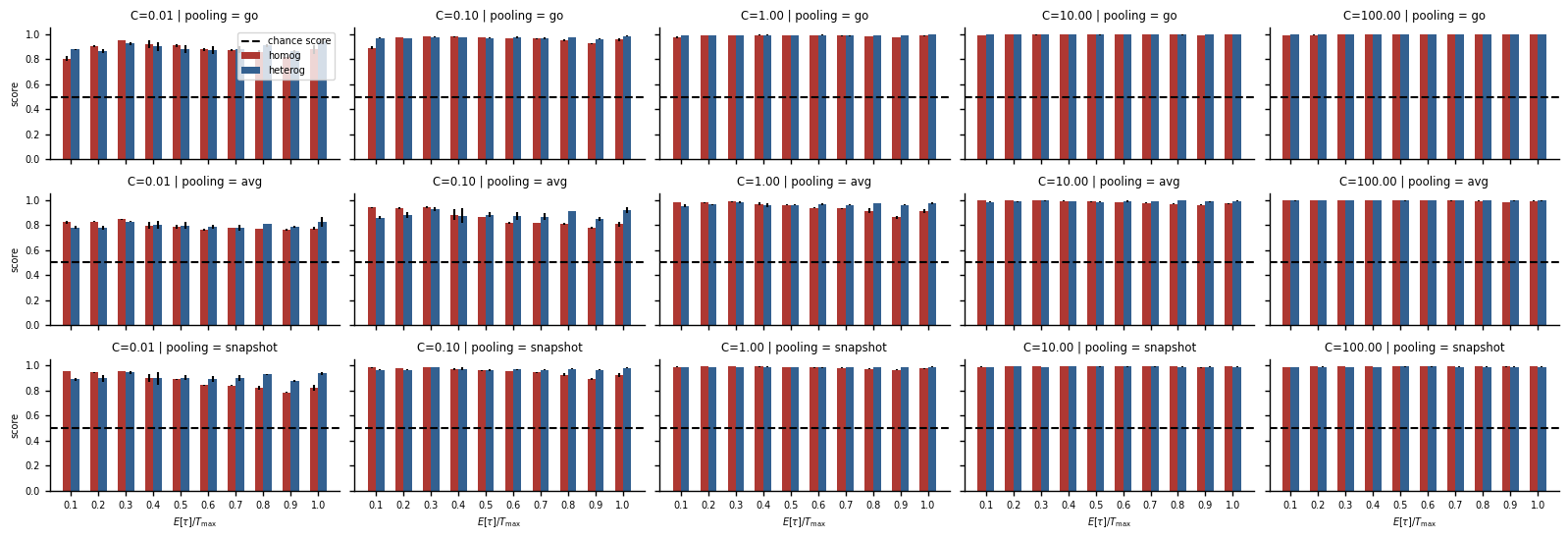}
    \caption{
    \textbf{CDM task with two options.} The performance of both networks is almost perfect for all pooling-regularization combinations. This is due to the simplicity of the task. Note the 50\% chance level as opposed to $1/6\approx 16\%$ in Fig.~\ref{fig:creg_pool-mante} for a setup with 6 choices.
    }
    \label{fig:easy_mante}
\end{figure}

\clearpage

\clearpage

 \subsection{Energy estimate on various computing platforms}\label{sec:loihi}

 \begin{figure}[!ht]
 	\centering
 	\includegraphics[width=\linewidth]{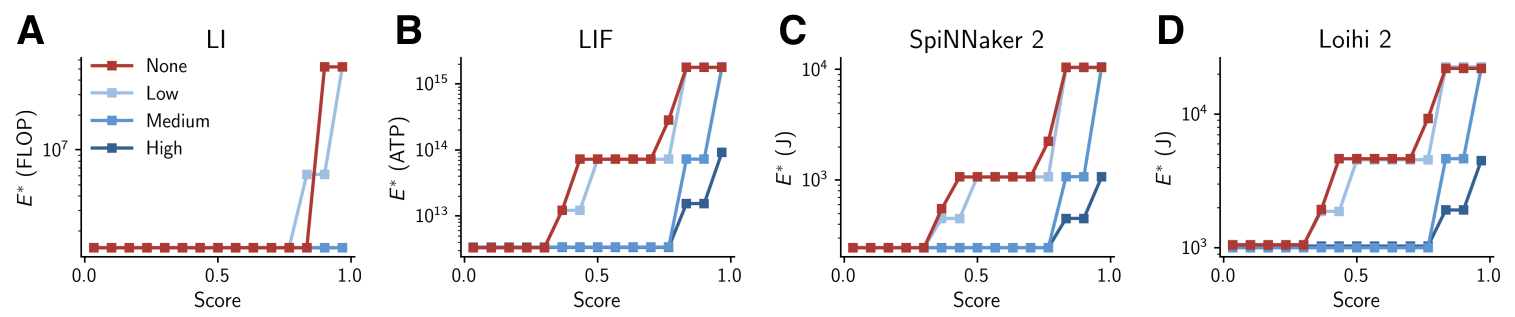}
     \caption{
         \textbf{Energy-score without Dale's law}. Similar to Figure~\ref{fig:energy}, but size-dependent simulation horizon and weights randomly ascribed to different synapses, thereby violating Dale's law. The traces associated with higher heterogeneity levels are partially occluded by those of lower heterogeneity levels. 
         }
     \label{fig:energy_nodale}
 \end{figure}

\subsection{Energy-score signatures}\label{sec:set_signiture}

\begin{figure}[!ht]
	\centering
	\includegraphics[width=.7\linewidth]{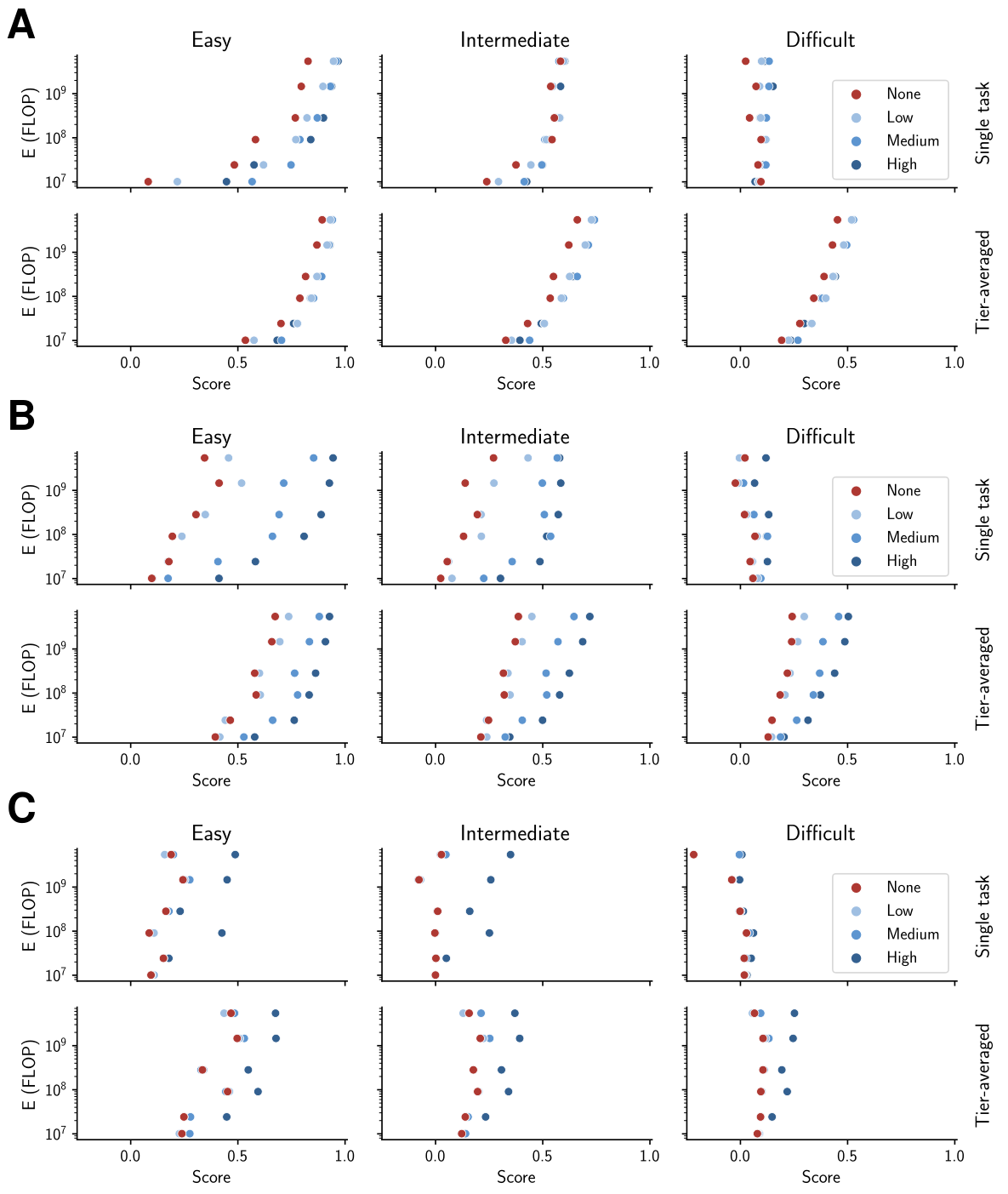}
    \caption{
        \textbf{Score-energy signatures in rate networks}. Seventy-two different state conditions (3 levels of $\E[\tau]$, 4 heterogeneity levels, and 6 network sizes) yield a distinct score for each task. Each subpanel shows the (score-energy) pair for either a single task \textit{(top row)} across different complexity tiers \textit{(columns)} or the average (score-energy) pair across all tasks within each tier \textit{(bottom row)}. Different panels display results for different $\E[\tau]$ values: \textbf{(A)} $1/\sqrt{10}$, \textbf{(B)} $1$, and \textbf{(C)} $\sqrt{10}$.
        }
    \label{fig:set_li}
\end{figure}

\begin{figure}[!ht]
	\centering
	\includegraphics[width=.7\linewidth]{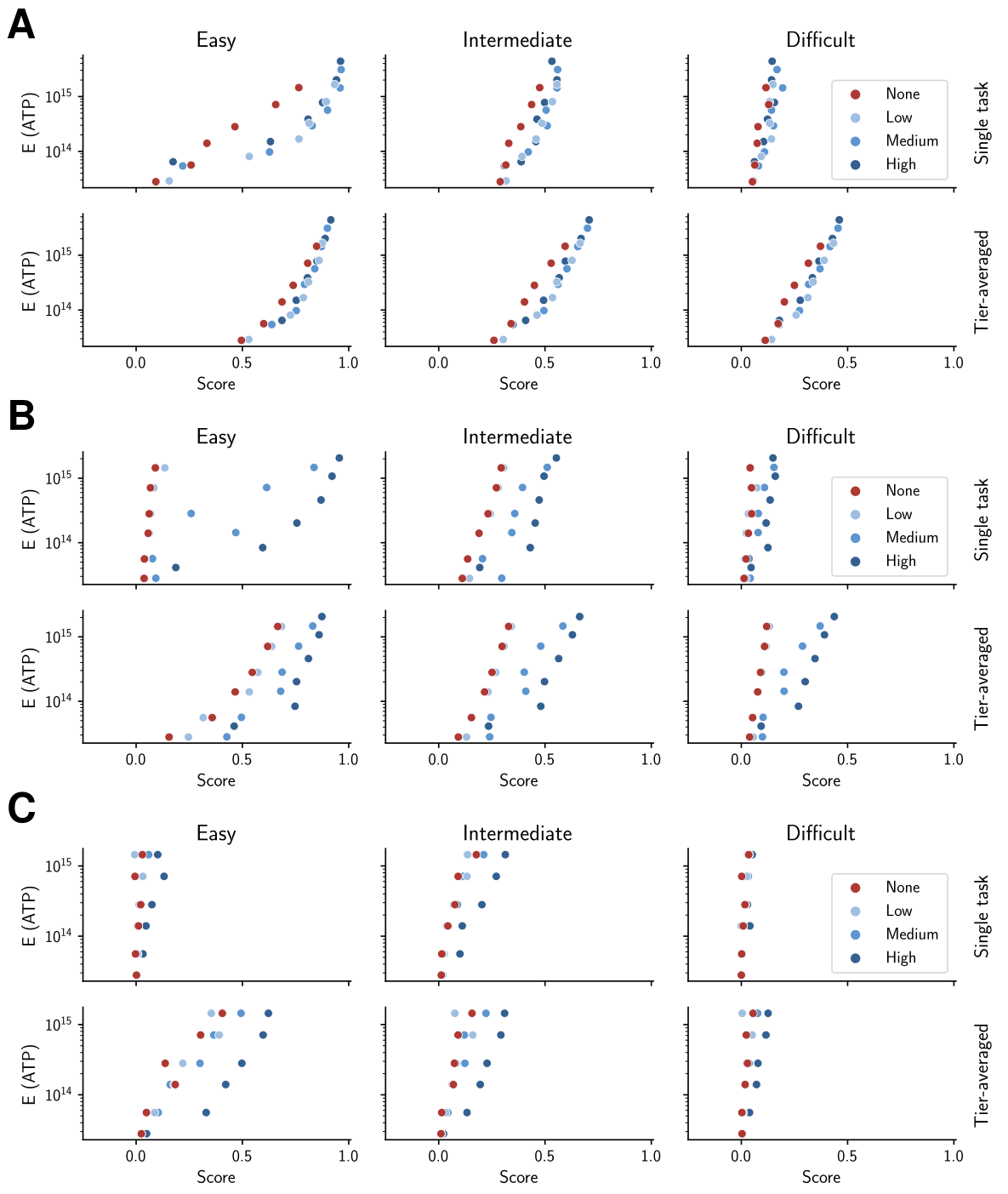}
    \caption{\textbf{Score-energy signatures in spiking networks}. As in Fig.~\ref{fig:set_li}, but for spiking networks. Note the different slopes for different heterogeneity levels.}
    \label{fig:set_lif}
\end{figure}

\printbibliography[title={References}]
\end{supplementary}
\end{refsection}

\end{document}